\PassOptionsToPackage{pagebackref,breaklinks,bookmarks=true}{hyperref}

\documentclass[sigconf, nonacm]{template/acmart}

\makeatletter
\def\@secfont{\sffamily\bfseries\Large\section@raggedright}
\def\@subsecfont{\sffamily\bfseries\large\section@raggedright}
\def\@subsubsecfont{\sffamily\itshape}
\def\@parfont{\sffamily\itshape}
\makeatother

\usepackage[utf8]{inputenc}   %
\usepackage[T1]{fontenc}      %
\usepackage[scaled]{helvet}   %

\usepackage{amsmath}
\usepackage{amssymb}
\usepackage{amsfonts}         %
\usepackage{bm}               %
\usepackage{nicefrac}         %
\usepackage[makeroom]{cancel} %

\usepackage{graphicx}
\usepackage{epsfig}           %
\usepackage[table]{xcolor}
\usepackage{tikz}
\usetikzlibrary{arrows.meta, backgrounds, calc, decorations.pathreplacing, fit, patterns}
\definecolor{ourportwine}{RGB}{128,25,55}
\usepackage{adjustbox}        %
\usepackage[most]{tcolorbox}  %

\usepackage{booktabs}         %
\usepackage{tabularx}

\usepackage{microtype}        %
\usepackage{url}              %
\usepackage{multicol}
\usepackage{dirtytalk}        %
\usepackage{soul}             %
\usepackage{epigraph}
\usepackage{comment}
\usepackage{xspace}
\usepackage{algorithm}
\usepackage{lipsum}

\usepackage[shortlabels]{enumitem}

\usepackage[font=small,labelfont=bf,textfont=normalfont]{caption}

\definecolor{ourblue}{rgb}{0.368,0.507,0.71}    %
\definecolor{ourdarkblue}{HTML}{354d72}         %
\definecolor{ourorange}{rgb}{0.881,0.611,0.142} %
\definecolor{ourgreen}{rgb}{0.56,0.692,0.195}   %
\definecolor{ourred}{rgb}{0.923,0.386,0.209}    %
\definecolor{ourviolet}{rgb}{0.528,0.471,0.701} %
\definecolor{ourbrown}{rgb}{0.772,0.432,0.102}  %
\definecolor{ourazure}{rgb}{0.364,0.619,0.782}  %
\definecolor{ourolive}{rgb}{0.572,0.586,0.}     %
\definecolor{ourgray}{RGB}{102,88,84}           %

\definecolor{ourblue2}{RGB}{9,134,223} %
\definecolor{ourdarkblue2}{RGB}{5,97,164} %
\definecolor{ourlightblue2}{RGB}{132,201,250} %

\definecolor{ourorange2}{RGB}{224,90,18} %
\definecolor{ourdarkorange2}{RGB}{160,63,9} %
\definecolor{ourlightorange2}{RGB}{246,175,137} %

\definecolor{ouryellow2}{RGB}{227,213,25} %
\definecolor{ourdarkyellow2}{RGB}{177,166,17} %
\definecolor{ourlightyellow2}{RGB}{242,235,140} %

\definecolor{ourpink2}{RGB}{247,24,139} %
\definecolor{ourdarkpink2}{RGB}{164,4,86} %
\definecolor{ourlightpink2}{RGB}{250,163,207} %

\definecolor{ourgreen2}{RGB}{159,198,52} %
\definecolor{ourdarkgreen2}{RGB}{109,138,30} %
\definecolor{ourlightgreen2}{RGB}{209,228,154} %

\definecolor{ourgray2}{RGB}{124,124,115} %
\definecolor{ourdarkgray2}{RGB}{87,87,81} %
\definecolor{ourlightgray2}{RGB}{194,194,189} %

\colorlet{primalconecolor}{ourblue2}        %
\colorlet{dualconecolor}{ourpink2}          %
\colorlet{polarconecolor}{ourgray2}         %
\colorlet{forcecolor}{ourportwine}          %
\colorlet{velocitycolor}{ourdarkgreen2}     %
\colorlet{correctioncolor}{ourorange}       %
\colorlet{stabilizationcolor}{ourviolet}    %
\colorlet{coordinatecolor}{ourdarkblue2}    %
\colorlet{torquecolor}{ourdarkorange2}      %
\colorlet{geometrycolor}{black}             %
\colorlet{bodycolor}{ourlightblue2}         %
\colorlet{fixedbodycolor}{ourlightgray2}    %
\colorlet{edgecolor}{ourdarkgray2}          %
\colorlet{annotationcolor}{ourdarkgray2}    %
\colorlet{procolor}{ourdarkgreen2}          %
\colorlet{concolor}{ourportwine}            %
\colorlet{relaxationcolor}{ourportwine}     %
\colorlet{highlightcolor}{ourblue2}         %
\colorlet{panelcolor}{ourblue!7}            %
\colorlet{mutedtextcolor}{ourgray2}         %
\colorlet{hyperlinkcolor}{ourblue}          %
\colorlet{biblinkcolor}{ourdarkblue}        %
\colorlet{bulletcolor}{ourblue}             %
\colorlet{boxcolor}{black!60}               %

\newtcolorbox{equationbox}[1]{
    enhanced,
    sharp corners,
    colback=white,
    colframe=boxcolor,
    boxrule=0.8pt,
    top=0.6em, %
    attach boxed title to top left={
        xshift=-1mm,
        yshift=-5mm,
        yshifttext=-1mm,
    },
    boxed title style={
        sharp corners,
        size=small,
        colback=boxcolor,
        colframe=boxcolor,
    },
    fonttitle=\bfseries,
    title=#1
}

\makeatletter
\newcommand*{\addFileDependency}[1]{%
  \typeout{(#1)}
  \@addtofilelist{#1}
  \IfFileExists{#1}{}{\typeout{No file #1.}}
}
\makeatother

\newcommand{\captitle}[1]{\textbf{#1}}

\makeatletter
\DeclareRobustCommand\onedot{\futurelet\@let@token\@onedot}
\def\@onedot{\ifx\@let@token.\else.\null\fi\xspace}
\makeatother

\newcommand{\trsp}{{\scriptscriptstyle\top}}
\newcommand{\Id}{\mathbf{I}}  %

\DeclareMathOperator*{\argmin}{arg\,min}
\DeclareMathOperator{\interior}{int}
\DeclareMathOperator{\boundary}{bd}

\input{contact_notation.tex}

\usepackage{algorithm}
\usepackage{algpseudocode}
\makeatletter
\renewcommand{\theHALG@line}{\thealgorithm.\arabic{ALG@line}}
\makeatother
\newcolumntype{P}[1]{>{\raggedright\arraybackslash}p{#1}} %

\usepackage{etoolbox}
\newenvironment{mysubequations}
  {\begin{subequations}%
   }
  {\end{subequations}}

\AtEndPreamble{\hypersetup{colorlinks=true,allcolors=hyperlinkcolor}}
\usepackage{xr-hyper}  %
\usepackage[capitalise]{cleveref} %

\AtBeginDocument{%
  }

\newcommand{\marc}[1]{}

\begin{document}

\title{A Tutorial on Complementarity-Based \\ Contact Models for Robotics}

\makeatletter
\def\@mkauthors{%
  \global\setbox\mktitle@bx=\vbox{\vskip-21.5pt\noindent\unvbox\mktitle@bx\par\medskip\vskip6.95pt\leavevmode
    \lineskip=1pc\relax\centering\hspace*{-1em}%
    \vtop{\hsize=\dimexpr\textwidth-2\author@bx@sep\relax\normalbaselines\centering\normalfont\normalsize
      Onur Beker\textsuperscript{1}, A. Ren\'e Geist\textsuperscript{1}, Michael Posa\textsuperscript{2}, Marc Toussaint\textsuperscript{3}, Justin Carpentier\textsuperscript{4}, Georg Martius\textsuperscript{1}\par
      \small
      \textsuperscript{1}University of T\"ubingen\quad
      \textsuperscript{2}University of Pennsylvania\quad
      \textsuperscript{3}Technical University Berlin\quad
      \textsuperscript{4}Inria \& DI-ENS\par}%
    \par\bigskip\vskip8pt}}
\makeatother

\begin{abstract}
Simulating rigid-body dynamics with contact is fundamental to robotics, from motion synthesis and control to mechanism design. Many widely used simulators rely on cone complementarity formulations, whose physical assumptions and numerical solution require substantial background. While excellent surveys, monographs, and course notes exist, they mostly address computational mechanics or computer graphics, or document a single simulator. To the best of our knowledge, a self-contained pedagogical treatment for robotics, bringing these formulations and their solvers together with complete derivations and a unified notation, does not yet exist. This tutorial addresses this gap by connecting rigid-body dynamics, conic optimization, and frictional contact.
In particular, we explain how complementarity conditions encode non-penetration and frictional dissipation, determining separation, sticking, and sliding without prescribed contact modes. We distinguish equivalent transcriptions from convex relaxations, showing how the latter can introduce artifacts such as lift-off and creep. We examine existence and uniqueness of motion and contact forces, including the solver's role in selecting among multiple solutions. We then compare first-order and Newton-based methods, explain regularization and proximal strategies, and discuss trade-offs between computational cost, robustness, and physical fidelity. Connections to widely used simulators, optimal control, and differentiable simulation help readers interpret simulation results, choose suitable formulations and solvers, and develop their own contact dynamics routines.

\end{abstract}

\maketitle

\newcommand{\folionumber}{\raisebox{-20pt}[0pt][0pt]{\normalsize\thepage\ of \ref*{TotPages}}}                               
\fancypagestyle{myfolio}[standardpagestyle]{%
\fancyhead{}%
\fancyfoot[C]{\folionumber}}                                             
\fancypagestyle{myfoliofirst}[firstpagestyle]{%
\fancyfoot[C]{\folionumber}}                                 \pagestyle{myfolio}\thispagestyle{myfoliofirst}

\section{Introduction} 
\label{intro}

\textit{What exactly is a contact model? And what does it mean to formulate one using cone complementarity?} Answering these questions requires taking a step back and asking a more fundamental question: where do contact forces come from?
The normal contact force arises from the short-range repulsion between atoms that are pushed close together. The frictional contact force arises from the interactions at the microscopic asperities where two surfaces actually touch, which adhere, deform, and interlock as the surfaces slide over each other. Contact is therefore a complex, multiscale effect, and a \say{mechanistic} approach that tries to numerically simulate these processes directly is intractable at the scale of robots. 

Practical contact models used in robotics are therefore \say{phenomenological}: they reproduce the macroscopic phenomenon (\say{what contacts do}) without reproducing the underlying microscopic mechanism (\say{how contacts do it}). The first question any contact model must answer is thus: what do contacts do? 
They do two things: 
i)~prevent intersections between surfaces that try to pass through each other, and 
ii)~dissipate energy between surfaces that try to slide on top of each other. All contact models, at their core, are therefore mathematical abstractions designed to produce these two effects in a way that makes theoretical analysis and numerical simulation tractable. And cone complementarity formulations of contact dynamics are a family of particularly elegant ways of doing so. 

Aimed at doctoral students and early-career robotics researchers, this tutorial collates the most common mathematical formulations of contact dynamics based on cone complementarity conditions, along with the key techniques and insights used to solve the resulting problems numerically. 
The material draws primarily from~\cite{baraff1997, mujoco_docs, stewart2000rigid, acary2018solving, andrews2022contact, lelidec2023contacts, carpentier2024compliant, tsounis2025solving, castro_sim, modern_robotics, boyd2004convex}, supplemented by the authors' own derivations. Upon completing the tutorial, readers should be well prepared to explore the open-source implementations of widely used simulators (e.g., MuJoCo~\cite{todorov2012mujoco}, Bullet~\cite{coumans2015bullet}, Drake~\cite{drake}, Kamino~\cite{tsounis2026kamino}, Simple~\cite{carpentier2024compliant}, Newton~\cite{The_Newton_Contributors_Newton_GPU-accelerated_physics_2025}), or to extend their understanding beyond forward simulation by investigating the use of cone complementarity formulations of contact dynamics in optimal control and trajectory optimization~\cite{mordatch2012discovery, posa2014direct, manchester2019variational, toussaint2020describing, tassa2012synthesis, howell2022trajectory, pang2023global}.
To get the most out of the tutorial, readers will benefit from prior familiarity with rigid-body dynamics~\cite{modern_robotics, featherstone2014rigid, toussaint_robot_kinematics, ETHRobotDynamics2017, Kim2012LieGroup, tedrake2024underactuated}, convex analysis~\cite{boyd2004convex, Nesterov2018, Bertsekas2009, parikh2014proximal}, and numerical methods~\cite{NocedalWright2006, solomon2015numerical, Strang2007CSE}.

\subsection{Scope and Positioning}
Before delving into the subject at a mathematical level, let us briefly clarify the scope of the tutorial and point readers to complementary pedagogical resources on topics that are not covered.

\subsubsection{Relation to Existing Surveys and Tutorials}
\label{sec:positioning}

Several existing resources cover material that overlaps with the contents of this tutorial. The SIGGRAPH course of Andrews, Erleben, and Ferguson~\cite{andrews2022contact} is the closest in spirit. Written for computer graphics, it covers collision detection in detail, and focuses on linear complementarity formulations as well as penalty and barrier formulations used for deformable bodies; with a brief mention of cone complementarity formulations used in robotics. This tutorial is complementary: it is written for robotics, focuses on cone complementarity formulations, and provides very brief summaries of collision detection and linear-complementarity formulations (mostly derived from the SIGGRAPH course itself). Acary, Br\'emond, and Huber~\cite{acary2018solving} survey the exact formulations of the frictional contact problem for a computational mechanics audience, and compare the solvers of Siconos numerically on a large benchmark collection. The present tutorial also covers the exact formulations (borrowing several of their mathematical constructions and terminology), and additionally discusses its convex relaxations that are commonly used in robotics. The monograph of Acary and Brogliato~\cite{acary2008numerical} and the review of Stewart~\cite{stewart2000rigid} treat the mathematical theory of nonsmooth dynamics and its time discretization in a depth and generality that go well beyond the scope of this tutorial. Le Lidec et al.~\cite{lelidec2023contacts} provide a comparative analysis of the contact dynamics formulations and solution algorithms used by current robotics simulators, including numerical experiments and discussions on the physical artifacts of the convex relaxations. This tutorial instead provides the pedagogy around the same subject, with step-by-step mathematical derivations, as well as additional formulations and solution algorithms that are no longer commonly employed in the modern simulation landscape of today but do nevertheless provide important context. Finally, the MuJoCo documentation~\cite{mujoco_docs} and the paper introducing Drake's SAP solver~\cite{castro_sim} present the mathematical formulations of these two simulators in detail, each within its own framework. This tutorial places both within a common family of transcriptions (i.e., the nonlinear least-squares transcription from \cref{sec:formulation}), and presents self-contained mathematical derivations.

\subsubsection{Time-stepping Methods and Impulse-Based Theories of Contact Dynamics}

This tutorial focuses on what are commonly called the \say{discrete time-stepping} family of methods for simulating cone complementarity formulations of contact dynamics, which underpin most of the widely used simulators of today~\cite{todorov2012mujoco, coumans2015bullet, drake, carpentier2024compliant, tsounis2026kamino}. The theoretical foundations of these methods can be traced to Moreau’s seminal work in the 1980s on the \say{measure differential inclusion} framework~\cite{moreau1983liaisons, moreau1985standard, Moreau1988}, followed by the development of \say{impulse-based theories} of contact dynamics~\cite{brogliato1996nonsmooth, Glocker2001, stewart2011dynamics}. Their computational foundations in turn originate from the \say{nonsmooth contact dynamics} framework developed by Moreau and Jean in the 1990s~\cite{Moreau1988, moreau1994numerical, Jean1999}, and the subsequent advances that shaped modern time-stepping methods~\cite{stewart1996implicit, stewart1998convergence, DeSaxceFeng1991, DeSaxceFeng1998, GlockerStuder2005, anitescu2006optimization, AnitescuTasora2010, TasoraAnitescu2011, acary2011formulation, todorov2012mujoco, todorov2014convex, acary2018solving, castro_sim, carpentier2024compliant}.

\subsubsection{Event-Driven Methods and Continuous-Time Theories of Contact Dynamics} 

Before impulse-based theories and discrete time-stepping methods rose to prominence, the prevailing approach was to formulate cone complementarity models of contact dynamics in continuous time~\cite{fourier1798statique, ostrogradsky1838momens, jellett1872treatise, Painleve1895, mayer1899aufstellung, zermelo1899bewegung, Delassus1917, Signorini1959, Fichera1963, Moreau1966, lotstedt1982mechanical}. Computationally, these formulations were commonly paired with what are called the \say{event-driven} family of methods, perhaps most notably in the research program of Glocker and Pfeiffer~\cite{GlockerPfeiffer1993, GlockerPfeiffer1995, PfeifferGlocker1996}. This family of event-driven methods can provide greater accuracy by explicitly resolving the timing of contact events, but typically incur substantially higher computational costs, making them more prevalent in computational mechanics than in robotics, where time-stepping methods offer a more efficient alternative with accuracy often sufficient for simulation and control. The event-driven family is therefore beyond the scope of this tutorial, but we refer interested readers to the classical textbook by Glocker and Pfeiffer~\cite{PfeifferGlocker1996}.

\subsubsection{Contact Formulations Beyond Cone Complementarity}
Not all widely used simulators rely on cone complementarity formulations of contact dynamics. In particular, linear complementarity formulations~\cite{Baraff1994, stewart1996implicit}, position-based dynamics~\cite{muller2007position, macklin2014unified, macklin2016xpbd} as well as penalty and barrier-based formulations~\cite{li2020incremental, ferguson2021intersection} are widely used in computer graphics. As these approaches are already covered extensively in existing tutorials and surveys~\cite{MullerTenMinutePhysics, li2026physics, Ferguson2023, andrews2022contact, bender2017survey, muller2020detailed, baraff1997}, we refer the interested reader to this material. Similarly, compliant contact models such as the classical spring-damper based point-contact laws~\cite{Hertz1882, hunt1975coefficient} and the hydroelastic pressure-field model~\cite{hydroelastic}, together with recent convex formulations and error-controlled integrators for such models~\cite{castro2025irrotational, kurtz2025cenic} are out of scope for this tutorial.

\subsection{Tutorial Content}
\label{sec:outline}

The tutorial presents the material in pedagogical order instead of chronological. Rather than tracing the historical development of the theory, the exposition builds directly on its current state of the art, using modern terminology from convex optimization and numerical methods to explain the key results and their underlying principles.
Following background sections that introduce the necessary concepts from rigid-body dynamics (\cref{sec:background_rbd}) and convex optimization with cone constraints (\cref{sec:background_cone}), the core material is organized under three main sections: Problem Formulation (\cref{sec:formulation}), Numerical Methods (\cref{sec:solution}), and Comparative Analysis (\cref{sec:comparison}). The tutorial then concludes with a discussion of future directions for the reader (\cref{sec:future}), and a retrospective overview of the material. The former highlights possible paths for further exploration such as trajectory optimization and differentiable simulation, helping readers identify where to continue their learning journey now that they are equipped with a foundational understanding of cone complementarity formulations contact dynamics.

\section{Background: Rigid-Body Simulation}
\label{sec:background_rbd}

A typical simulation framework consists of three main routines:
i)~rigid-body dynamics (accounting for internal joint constraints of articulated robots, without contacts),
ii)~collision detection (identifying contact points between colliding geometries), and
iii)~contact dynamics (computing contact forces and their effects on motion). Although this tutorial focuses on contact dynamics, every transcription of the contact problem relies on quantities computed by the first two, therefore this section briefly reviews them to put the rest of the tutorial into context. It also introduces the notational conventions used in the rest of the tutorial.

\subsection{Rigid-Body Dynamics}
\label{sec:rigid_body}
\usetikzlibrary{arrows.meta,calc,patterns}
\begin{figure}[t]
\centering
\begin{tikzpicture}[
    scale=1.3, font=\footnotesize,
    >={Stealth[length=3.5pt]},
    linkout/.style={line width=7pt, line cap=round, draw=edgecolor},
    linkin/.style={line width=5.6pt, line cap=round, draw=fixedbodycolor},
    joint/.style={circle, draw=edgecolor, fill=white, line width=0.6pt, inner sep=0pt, minimum size=5pt},
    coord/.style={->, line width=0.6pt, draw=coordinatecolor},
    torque/.style={->, line width=0.7pt, draw=torquecolor},
    vel/.style={->, line width=0.8pt, draw=velocitycolor},
    force/.style={->, line width=1.1pt, draw=forcecolor},
    frame/.style={->, line width=0.5pt, draw=black},
    helper/.style={line width=0.4pt, draw=annotationcolor!75, densely dashed},
    com/.pic={%
      \fill[white] (0,0) circle (0.06);
      \fill (0,0) -- (0.06,0) arc (0:90:0.06) -- cycle;
      \fill (0,0) -- (-0.06,0) arc (180:270:0.06) -- cycle;
      \draw[line width=0.4pt] (0,0) circle (0.06);
    },
  ]
  \coordinate (J1) at (1.0,0.55);
  \coordinate (J2) at ($(J1)+(55:2.0)$);
  \coordinate (E)  at ($(J2)+(-20:1.9)$);   %
  \coordinate (C1) at ($(J1)!0.5!(J2)$);
  \coordinate (C2) at ($(J2)!0.5!(E)$);
  \coordinate (W)  at (5.75,0.2);           %

  \fill[pattern=north east lines, pattern color=annotationcolor!75] (0.55,-0.12) rectangle (1.45,0);
  \draw[line width=0.6pt, edgecolor] (0.55,0) -- (1.45,0);
  \filldraw[fill=fixedbodycolor, draw=edgecolor, line width=0.6pt]
    (0.78,0) -- (0.9,0.55) -- (1.1,0.55) -- (1.22,0) -- cycle;

  \foreach \a/\b in {J1/J2, J2/E} {
    \draw[linkout] (\a) -- (\b);
    \draw[linkin]  (\a) -- (\b);
  }
  \node[joint] at (J1) {};
  \node[joint] at (J2) {};

  \begin{scope}[shift={(E)}, rotate=12]
    \filldraw[fill=bodycolor!35, draw=edgecolor, line width=0.6pt] (0,-0.4) rectangle (1.5,0.7);
    \coordinate (C3) at (0.75,0.15);
    \draw[frame] (C3) -- ++(0.45,0);
    \draw[frame] (C3) -- ++(0,0.42);
    \node at (1.2,0.47) {$R_3$};
    \node at (0.38,0.47) {$m_3, I_3$};
    \draw[force] (0,0) -- (0.42,0) node[pos=0.5, below, inner sep=2pt, text=forcecolor] {$f$};
  \end{scope}
  \pic at (C3) {com};
  \fill[forcecolor] (E) circle (0.035);
  \node[anchor=north east, inner sep=2.5pt, text=forcecolor] at (E) {$J$};

  \draw[helper] (J1) -- ++(0.75,0);
  \draw[coord] ($(J1)+(0:0.5)$) arc (0:55:0.5);
  \node[text=coordinatecolor] at ($(J1)+(27:0.68)$) {$q_1$};
  \draw[helper] (J2) -- ++(55:0.7);
  \draw[coord] ($(J2)+(55:0.45)$) arc (55:-20:0.45);
  \node[text=coordinatecolor] at ($(J2)+(18:0.65)$) {$q_2$};
  \draw[coord, densely dashed, shorten >=2pt] (W) -- (C3) node[pos=0.4, left, inner sep=2pt, text=coordinatecolor] {$q_3$};

  \draw[torque] ($(J1)+(110:0.22)$) arc (110:250:0.22);
  \node[text=torquecolor] at ($(J1)+(180:0.42)$) {$u_1$};
  \draw[torque] ($(J2)+(80:0.22)$) arc (80:200:0.22);
  \node[text=torquecolor] at ($(J2)+(140:0.42)$) {$u_2$};

  \pic at (C1) {com};
  \node[left=3pt] at (C1) {$m_1, I_1$};
  \pic at (C2) {com};
  \draw[vel] (C2) -- ++(70:0.6) node[left, inner sep=1.5pt, text=velocitycolor] {$v_2$};
  \draw[vel] ($(C2)+(185:0.24)$) arc (185:335:0.24);
  \node[text=velocitycolor] at ($(C2)+(262:0.42)$) {$w_2$};
  \node[text=velocitycolor] at ($(C2)+(0,-0.72)$) {$(v_2, w_2) = J_2\dot q$};

  \draw[frame] (W) -- ++(0.5,0) node[right, inner sep=1.5pt] {$\hat{e}_x$};
  \draw[frame] (W) -- ++(0,0.5) node[above, inner sep=1.5pt] {$\hat{e}_z$};
  \draw[->, line width=0.8pt, draw=annotationcolor] (6.1,2.65) -- ++(0,-0.6) node[midway, right, inner sep=2pt] {$g\hat{e}_z$};
\end{tikzpicture}
\caption{\captitle{An example of rigid-body system and its quantities.} A planar two-link arm (bodies 1 and 2) on a fixed base pushes a free-floating box (body 3). The generalized coordinates $q = (q_1, q_2, q_3)$ are the two joint angles and the pose of the box, and the joint torques constitute the controls $u = (u_1, u_2, 0)$, as the box is not actuated. Each body $k$ has a mass $m_k$, an inertia tensor $I_k$, a rotation matrix $R_k$, and a spatial velocity $(v_k, w_k) = J_k \dot{q}$; to avoid clutter, each of these quantities is drawn on one body only. Where the arm touches the box, it exerts the force $f$ on the box at an application point with kinematic Jacobian $J$. Gravity acts along $g\hat{e}_z$.}
\label{fig:rigid_body}
\Description{A planar two-link arm on a fixed base, with a joint angle and a joint torque at each joint, pushes a free-floating rectangular box. Mass, inertia, rotation, spatial velocity and a contact force are each annotated on one body; gravity points down and a world frame is drawn on the right.}
\end{figure}
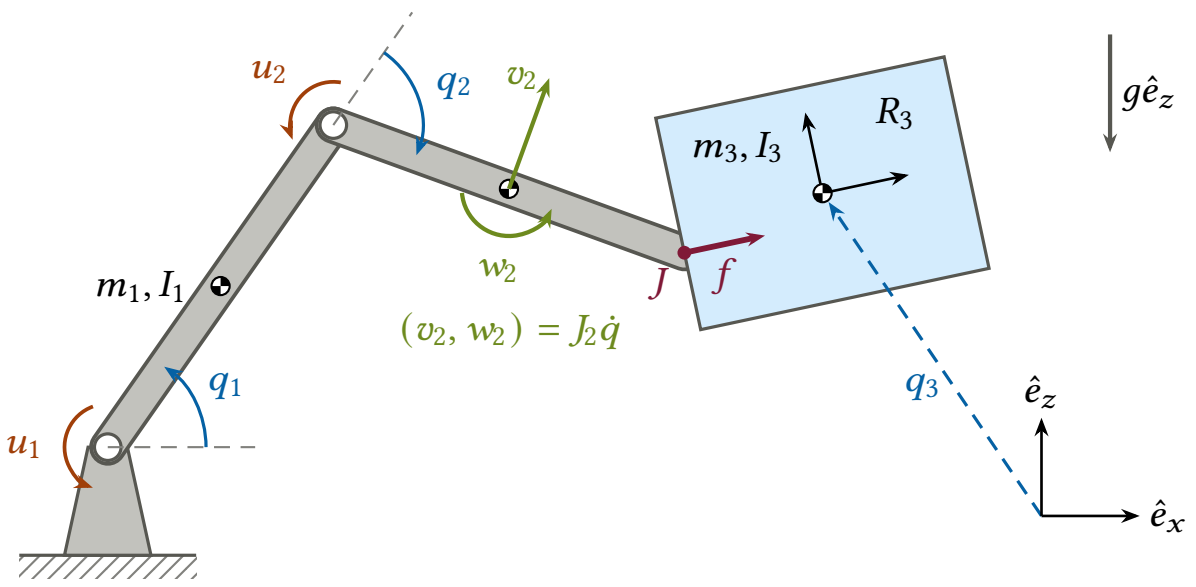

\subsubsection{Continuous Time} 
The material presented in this section is mostly derived from~\cite{toussaint_robot_kinematics}, to which we refer the interested reader. Alternatively, a more comprehensive, textbook-style coverage of rigid body dynamics can be found in~\cite{modern_robotics}.
Consider a rigid body system (\cref{fig:rigid_body}) with $n_v$ degrees of freedom and generalized position $q\in\mathbb{R}^{n_v}$, with local $se(3)$-Jacobians $J_k\in\mathbb{R}^{6\times n_v}$, mass $m_k \in \mathbb R$, and inertia tensor $I_k \in \mathbb{R}^{3 \times 3}$ for each body $k$, and joint torques $u\in\mathbb{R}^{n_v}$. The contact-free dynamics of the system are given by:
\begin{align}
    u &= \sum_{k} J_k^\trsp
    \left[ M_k J_k \ddot{q} + c_k + g_k \right], \\
    M_k &= \begin{pmatrix}
        m_k \Id_3 & 0 \\
        0 & \bar{I}_k
    \end{pmatrix}, \displaybreak[0] \\ %
    c_k &= M_k \dot{J}_k \dot{q}
    + \begin{pmatrix}
        0_3 \\
        w_k \times \bar{I}_k w_k
    \end{pmatrix}, \\
    g_k &= \begin{pmatrix}
        m_k g \hat{e}_z \\
        0_3
    \end{pmatrix}.
\end{align}
Here, $M_k \in \mathbb{R}^{6\times6}$ is the generalized inertia matrix of body $k$, \mbox{$\Id_3\in\mathbb{R}^{3\times3}$} denotes the identity matrix, $g \hat{e}_z$ is gravitational acceleration, $\bar{I}_i = R_i I_i R_i^\trsp$ is the rotated inertia tensor, $(v_k,w_k) = J_k \dot{q} \in \mathbb{R}^6$ is the spatial velocity of body $k$ expressed in its center-of-mass frame (making $M_k$ block-diagonal), and $R_k$ denotes the rotation matrix of body $k$. This \emph{inverse dynamics} equation is typically rewritten in a more compact form, known as the \emph{Newton-Euler equation}:
\begin{align} \label{eq:eom-q}
M(q)~ \ddot q + b(q, \dot q) = u,
\end{align}
with the generalized inertia matrix $M(q) = \sum_k J_k^\trsp M_k J_k \in \mathbb{R}^{n_v\times n_v}$ and bias force $b(q,\dot q) = \sum_k J_k^\trsp (c_k + g_k) \in \mathbb{R}^{n_v}$. In practice, $M(q)$ and $b(q, \dot q)$ can be computed efficiently using the composite rigid-body (CRB) and recursive Newton-Euler (RNE) algorithms respectively~\cite{featherstone2014rigid, carpentier2019pinocchio}. The \emph{forward dynamics} equation then becomes:
\begin{align}\label{eq:a_free}
\ddot q_f = M^{-1}(u - b),
\end{align}
where $\ddot q_f$ is called the \emph{free acceleration}. Consider a set of $N$ forces $f_i \in \mathbb{R}^3$ acting on the rigid-body system at application points with kinematic Jacobians $J_i \in \mathbb{R}^{3 \times n_v}$, concatenated into $f \in \mathbb{R}^{3N}$ and $J \in \mathbb{R}^{3N \times n_v}$. They enter the Newton-Euler equation as~\cite{modern_robotics}:
\begin{align} \label{eq:dynamics_with_forces}
M(q)~ \ddot q + b(q, \dot q) = u + J^\trsp f.
\end{align}

\subsubsection{Discrete Time}
Most simulation frameworks operate via a symplectic Euler discretization.
For an ordinary differential equation (ODE) of the form $\ddot{q}(q, \dot{q}, u)$, the corresponding equations are:
\begin{align}
    \dot{q}^+ &= \dot{q} + \ddot{q}(q, \dot{q}, u) \Delta t \label{eq:Euler1},\\
    q^+ &= q + \dot{q}^+ \Delta t,\label{eq:Euler2}
\end{align}
where $\Delta t$ is the step size, and $(q^+, \dot{q}^+)$ denote the next state of the system. For the dynamics equation \eqref{eq:dynamics_with_forces}, the corresponding ODE form, after substituting the free acceleration \eqref{eq:a_free}, is $\ddot{q} = \ddot{q}_f + (M^{-1}J^\trsp) f$. 
Applying the symplectic Euler discretization then defines the \emph{end-of-step generalized velocity} $\dot{q}^+$ as:
\begin{align}
\dot{q}^+ &= \dot{q} + \ddot{q}_f\Delta t + (M^{-1}J^\trsp) f \Delta t \\
&= \dot{q}_f + (M^{-1}J^\trsp) f \Delta t,   
\end{align}
where $\dot{q}_f = \dot{q} + \ddot{q}_f\Delta t$ is called the \say{free velocity}.
The concatenated free velocities $v_f \in \mathbb{R}^{3N}$ and end-of-step velocities $v^+ \in \mathbb{R}^{3N}$ at the application points of the $N$ forces can then be computed as:
\begin{align}
    v_f &= J\dot{q}_f, \\
    v^+ &= J\dot{q}^+ \label{eq:vplus}\\
    &= J\dot{q}_f + (JM^{-1}J^\trsp) f \Delta t\\
    &= v_f + G f\Delta t,
\end{align}
where \mbox{$G = JM^{-1}J^\trsp \in \mathbb{R}^{3N \times 3N}$} is the \emph{Delassus matrix}~\cite{Delassus1917}, the inverse generalized inertia projected into constraint space through the Jacobian $J$. It maps the applied impulse $f\Delta t$ to the velocity change \mbox{$v^+ - v_f$} at the force application points.

\subsection{Collision Detection}\label{sec:collision}
\usetikzlibrary{arrows.meta,patterns}
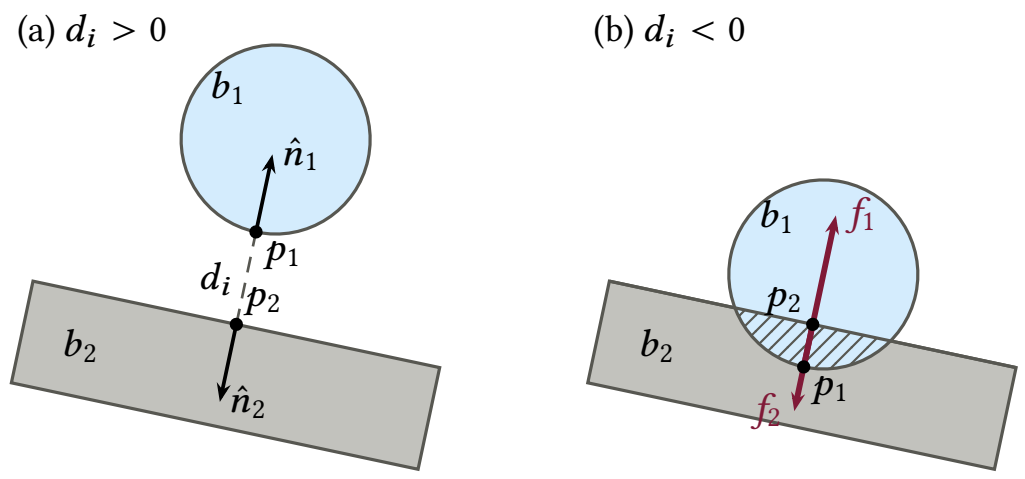
\begin{figure}[bht]
\centering
\begin{tikzpicture}[
    scale=1.25, font=\footnotesize,
    >={Stealth[length=3.5pt]},
    body1/.style={fill=bodycolor!35, draw=edgecolor, line width=0.6pt},
    body2/.style={fill=fixedbodycolor, draw=edgecolor, line width=0.6pt},
    normal/.style={->, line width=0.7pt, draw=geometrycolor},
    force/.style={->, line width=1.1pt, draw=forcecolor},
    helper/.style={line width=0.5pt, draw=annotationcolor, densely dashed},
    wpoint/.style={fill=geometrycolor},
  ]
  \begin{scope}[rotate=-12]
    \filldraw[body2] (-1.1,-0.55) rectangle (1.1,0);
    \filldraw[body1] (0,1.0) circle (0.5);
    \node at (-0.78,-0.28) {$b_2$};
    \node at (-0.3,1.22) {$b_1$};
    \coordinate (P1) at (0,0.5);
    \coordinate (P2) at (0,0);
    \draw[helper] (P2) -- (P1) node[pos=0.5, left, inner sep=2pt] {$d_i$};
    \draw[normal] (P1) -- ++(0,0.42) node[right, inner sep=1.5pt] {$\hat{n}_1$};
    \draw[normal] (P2) -- ++(0,-0.42) node[right, inner sep=1.5pt] {$\hat{n}_2$};
    \fill[wpoint] (P1) circle (0.035) node[below right, inner sep=1.5pt] {$p_1$};
    \fill[wpoint] (P2) circle (0.035) node[above right, inner sep=1.5pt] {$p_2$};
  \end{scope}
  \node[anchor=west] at (-1.25,1.55) {(a) $d_i > 0$};

  \begin{scope}[shift={(3.05,0)}, rotate=-12]
    \filldraw[body2] (-1.1,-0.55) rectangle (1.1,0);
    \filldraw[body1] (0,0.27) circle (0.5);
    \begin{scope}
      \clip (-1.1,-0.55) rectangle (1.1,0);
      \fill[pattern=north east lines, pattern color=edgecolor] (0,0.27) circle (0.5);
    \end{scope}
    \draw[edgecolor, line width=0.6pt] (-1.1,0) -- (1.1,0);
    \node at (-0.78,-0.28) {$b_2$};
    \node at (-0.3,0.52) {$b_1$};
    \coordinate (P1) at (0,-0.23);
    \coordinate (P2) at (0,0);
    \draw[force] (P1) -- ++(0,0.82) node[right, inner sep=1.5pt, text=forcecolor] {$f_1$};
    \draw[force] (P2) -- ++(0,-0.47) node[left, inner sep=1.5pt, text=forcecolor] {$f_2$};
    \fill[wpoint] (P1) circle (0.035) node[below right, inner sep=1.5pt] {$p_1$};
    \fill[wpoint] (P2) circle (0.035) node[above left, inner sep=1.5pt] {$p_2$};
  \end{scope}
  \node[anchor=west] at (1.8,1.55) {(b) $d_i < 0$};
\end{tikzpicture}
\caption{\captitle{Collision detection and its quantities.} A ball $b_1$ and a box $b_2$, (a) separated and (b) penetrating. The witness points $p_1$ and $p_2$ lie on the two surfaces, and the normals satisfy $\hat{n}_1 = -\hat{n}_2$, with $\hat{n}_i = \hat{n}_1$ pointing from $b_2$ toward $b_1$. The signed distance is $d_i = \hat{n}_i^\trsp (p_1 - p_2)$, and a repulsive force pair $f_1 = -f_2$ satisfies $\hat{n}_i^\trsp f_1 \geq 0$.}
\label{fig:collision}
\Description{Two panels with a ball above a box. In the left panel the bodies are separated: a witness point lies on each surface, the two normals point away from each other, and the positive signed distance spans the gap. In the right panel the ball penetrates the box, and opposite forces act on the two bodies along the normal.}
\end{figure}

The role of collision detection (\cref{fig:collision}) is to identify intersecting or nearby collision geometries and characterize their relative geometric configuration. For each potential contact $i$, the following information is used to formulate the contact dynamics problem:
i)~the pair of bodies $(b_1, b_2)_i$ involved in the collision,
ii)~a signed-distance $d_i$,
iii)~a pair of witness points $(p_1, p_2)_i$ for contact forces to act upon, one on each body, with corresponding contact normals $\left(\hat{n}_1, \hat{n}_2\right)_i$, satisfying $\hat{n}_1 = -\hat{n}_2$, and the witness-point kinematic Jacobians $(J_1, J_2)$~\cite{van2003collision, gregorius2015robust}. The \emph{contact Jacobian} is then defined as $J_i = J_1 - J_2$, and the \emph{contact velocity} as $v_i = J_i \dot{q}$.

The definition of the contact Jacobian mainly constitutes a notational convenience. As the two bodies experience equal and opposite forces $f_1 = -f_2$ along $\left(\hat{n}_1, \hat{n}_2\right)$, the contact Jacobian allows the pair to enter the dynamics equation \eqref{eq:dynamics_with_forces} as a single term $J_i^\trsp f_i$ with
$f_i := f_1$, rather than two terms as $J_1^\trsp f_1 + J_2^\trsp f_2$.
For a model with $N$ active contact pairs, stacking their velocities, Jacobians, and forces gives $v \in \mathbb{R}^{3N}$, $J \in \mathbb{R}^{3N \times n_v}$, and $f \in \mathbb{R}^{3N}$. The dynamics then read:
\begin{equation}\label{eq:dyn_full}
M \ddot{q} + b = u + J^\trsp f.
\end{equation}

\smallbreak
\textbf{Computing Contact Geometry.} For convex shapes, the Gilbert--Johnson--Keerthi (GJK) algorithm~\cite{gilbert1988fast} and its accelerated variants GJK++~\cite{montaut2024gjk} use support mappings to test whether there is an intersection,\footnote{For a convex mesh, the support map is the function that returns the vertex $p$ that lies furthest along any input direction $d$ (i.e. that maximizes $p^\trsp d$).} and compute separation distances and witness points. For intersecting pairs, the Expanding Polytope Algorithm (EPA) estimates the penetration depth and associated normal~\cite{van2003collision}. 
For nonconvex geometries, collision queries can be performed using several approaches: approximate convex decomposition methods, such as CoACD~\cite{wei2022coacd}, followed by pairwise queries between the resulting convex components; triangle-mesh queries accelerated by bounding-volume hierarchies (BVHs), such as OBBTree~\cite{gottschalk1996obbtree}; or signed distance fields (SDFs), whose values and gradients provide point-to-surface distances and surface normals, respectively, wherever the field is differentiable~\cite{guendelman2003nonconvex}.
These queries provide the geometric inputs for the subsequent contact dynamics computation.

The quality of the contact points and normals generated by the collision detection routine plays a crucial role in determining the resulting dynamics, particularly near sharp edges of mesh-based shapes. Erleben~\cite{erleben2018methodology} provides a taxonomy of contact point generation methods and a benchmark methodology for assessing their geometric quality. His comparative study shows that none of the eight local methods considered consistently produces correct contact information in all test cases, highlighting limitations that cannot be addressed by the contact solver alone.

\smallbreak
\textbf{Kinematic Jacobians and Geometric Derivatives.} An important point of caution is that the kinematic Jacobians $(J_1, J_2)$ are defined such that $J_1 \dot{q}$ is the velocity that an observer would measure if they are positioned at $p_1$ but move \emph{attached to the body} $b_1$, rather than the velocity of the witness point $p_1$ itself (i.e., $\dot{p}_1 \neq J_1 \dot{q}$). The difference is that the witness points are defined strictly as the endpoints of the maximal penetration distance between two colliding surfaces, which means $\dot{p}_1$ is naturally a function of the curvature of both surfaces at $p_1$ and $p_2$, and the kinematic Jacobian $J_1$ does not in any way capture this dependency. The distinction becomes particularly relevant when transitioning from forward simulation to a trajectory optimization setting where witness points are treated as decision variables and having access to the correct $\frac{\partial}{\partial q}p_1$ is essential~\cite{toussaint2020describing}.

\subsection{Further Notational Conventions}
Scalars, vectors, and matrices are not distinguished typographically; their dimensions are stated where they are introduced, and calligraphic letters denote sets and cones. For a set $\mathcal{C}$, $\interior\mathcal{C}$ denotes its interior and $\boundary\mathcal{C} = \mathcal{C} - \interior\mathcal{C}$ its boundary. All sets considered in this tutorial are closed. The generalized velocity and acceleration are written $\dot q$ and $\ddot q$, while $v$ always denotes contact velocities.\footnote{For rotational degrees of freedom, this tutorial assumes the use of intrinsic coordinates on the $SO(3)$ manifold, namely, exponential coordinates derived from Rodrigues' formula~\cite{modern_robotics}. Consequently, no additional kinematic map is required~\cite{Bender2014Interactive, andrews2022contact}.} Quantities stacked over all contacts carry no contact index, as in $f \in \mathbb{R}^{3N}$ with blocks $f_i \in \mathbb{R}^3$, and passages that treat a single contact drop the index as well.
For any 3D vector $x \in \mathbb{R}^3$ and a normal direction $\hat{n}$ (with unit norm), the normal component of $x$ is defined as the scalar $x_n = \hat{n}^\trsp x$ and the tangential component as the vector $x_t = x - x_n \hat{n} = \left(\Id - \hat{n} \hat{n}^\trsp\right) x$, with the tangential direction $\hat{t}$ in turn defined through $x_t = \|x_t\| \hat{t}$. When convenient, the exposition also adopts a normal-aligned coordinate frame and uses the shorthand $x = [x_n, x_t]^\trsp$ with $x_n \in \mathbb R$ and $x_t \in \mathbb R^2$, overloading the notation of the tangential component.
The complementarity notation $0 \leq a \perp b \geq 0$ is shorthand for the three conditions $a \geq 0$, $b \geq 0$, $a^\trsp b = 0$ (where the inequalities are defined element-wise). Similarly, $\mathcal S_1 \ni a \perp b \in \mathcal S_2$ is shorthand for $a \in \mathcal S_1$, $b \in \mathcal S_2$, $a^\trsp b = 0$.

\section{Background on Conic Optimization}
\label{sec:background_cone}

The mathematical structure of contact models follows naturally from their two phenomenological duties: preventing interpenetration and dissipating energy. Both involve opposing relative motion: normal forces prevent bodies from moving into one another, while friction resists tangential sliding. This motivates an \emph{optimization} viewpoint, combining constraints on admissible motion and forces with dissipation or energy-minimization principles (\cref{sec:kinmin}).

\smallskip
Crucially, contact forces act selectively. In ideal unilateral contact, separated bodies exert no force, while a nonzero normal force requires closed contact. Likewise, sliding requires the friction bound to be saturated, whereas a friction force strictly below that bound implies sticking. \emph{Complementarity constraints} express these mutually exclusive possibilities algebraically, allowing contact modes to emerge from the solution rather than being prescribed in advance.

\smallskip
The admissible forces themselves have a \emph{conic} structure: normal forces are nonnegative, and Coulomb friction bounds the tangential force in proportion to the normal force. These constraints define convex cones, making \emph{conic optimization} a natural framework. Convexity is valuable computationally: every local minimum is global, and strict convexity ensures uniqueness when a minimizer exists. Complementarity also appears in the optimality conditions of cone-constrained optimization, connecting these mathematical tools to the physical contact laws. Although the full frictional contact problem is generally nonconvex, this connection yields convex relaxations and convex subproblems that underpin many practical solvers. This section develops the essential geometry and optimality conditions needed to understand these formulations.

\subsection{Convex Cones}\label{sec:convex_cones}
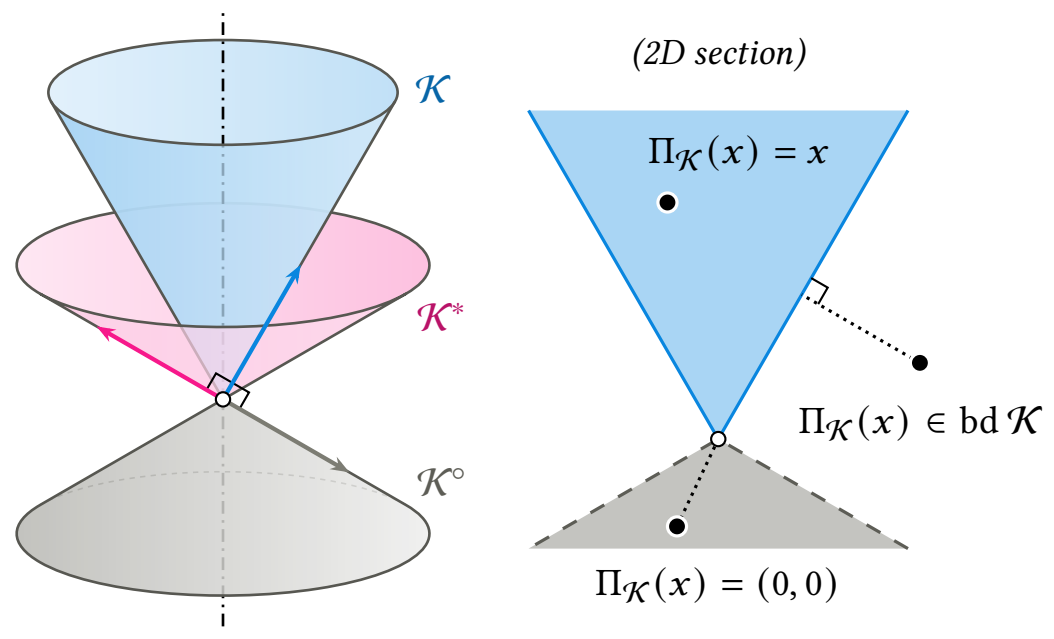
\begin{figure}[th]
    \centering
    \usetikzlibrary{arrows.meta,calc}
\begin{tikzpicture}[
    scale=1.05, font=\footnotesize,
    >={Stealth[length=3.5pt]},
    outline/.style={draw=edgecolor, line width=0.5pt, line join=round},
    hidden/.style={draw=edgecolor, line width=0.3pt, dash pattern=on 1pt off 1pt},
    axis/.style={draw=geometrycolor, line width=0.5pt, dash pattern=on 2.4pt off 1.2pt on 0.5pt off 1.2pt},
    bdray/.style={->, line width=0.8pt},
    rangle/.style={draw=geometrycolor, line width=0.4pt},
    apex/.style={fill=white, draw=geometrycolor, line width=0.4pt},
    point/.style={fill=geometrycolor, draw=white, line width=0.6pt},
    proj/.style={draw=geometrycolor, line width=0.6pt, densely dotted},
  ]
  \pgfmathsetmacro\aK{1.10}
  \pgfmathsetmacro\bK{0.3*\aK}
  \pgfmathsetmacro\hK{sqrt(3*\aK*\aK + \bK*\bK)}
  \pgfmathsetmacro\tK{asin(\bK/\hK)}
  \pgfmathsetmacro\aD{1.30}
  \pgfmathsetmacro\bD{0.3*\aD}
  \pgfmathsetmacro\hD{sqrt(\aD*\aD/3 + \bD*\bD)}
  \pgfmathsetmacro\tD{asin(\bD/\hD)}
  \coordinate (A) at (0,0);
  \coordinate (KL) at ({-\aK*cos(\tK)},{\hK-\bK*sin(\tK)});   %
  \coordinate (KR) at ({\aK*cos(\tK)},{\hK-\bK*sin(\tK)});
  \coordinate (DL) at ({-\aD*cos(\tD)},{\hD-\bD*sin(\tD)});   %
  \coordinate (DR) at ({\aD*cos(\tD)},{\hD-\bD*sin(\tD)});
  \coordinate (PL) at ({-\aD*cos(\tD)},{-\hD+\bD*sin(\tD)});  %
  \coordinate (PR) at ({\aD*cos(\tD)},{-\hD+\bD*sin(\tD)});

  \shade[left color=dualconecolor!10, right color=dualconecolor!28]
    (A) -- (DR) arc[start angle=-\tD, end angle=180+\tD, x radius=\aD, y radius=\bD] -- cycle;
  \draw[outline] (DR) arc[start angle=-\tD, end angle=180+\tD, x radius=\aD, y radius=\bD];
  \shade[left color=primalconecolor!14, right color=primalconecolor!32, fill opacity=0.8]
    (A) -- (KR) arc[start angle=-\tK, end angle=180+\tK, x radius=\aK, y radius=\bK] -- cycle;
  \draw[hidden] (PR) arc[start angle=\tD, end angle=180-\tD, x radius=\aD, y radius=\bD];
  \draw[axis] (0,-1.43) -- (0,2.44);
  \shade[left color=polarconecolor!55, right color=polarconecolor!10, fill opacity=0.85]
    (A) -- (PR) arc[start angle=\tD, end angle=-180-\tD, x radius=\aD, y radius=\bD] -- cycle;
  \draw[outline] (PL) -- (A) -- (PR) arc[start angle=\tD, end angle=-180-\tD, x radius=\aD, y radius=\bD];
  \shade[left color=primalconecolor!42, right color=primalconecolor!16, fill opacity=0.75]
    (A) -- (KL) arc[start angle=-180+\tK, end angle=-\tK, x radius=\aK, y radius=\bK] -- cycle;
  \draw[outline] (KL) -- (A) -- (KR);
  \draw[outline] (0,\hK) ellipse[x radius=\aK, y radius=\bK];
  \shade[left color=dualconecolor!24, right color=dualconecolor!13, fill opacity=0.7]
    (A) -- (DL) arc[start angle=-180+\tD, end angle=-\tD, x radius=\aD, y radius=\bD] -- cycle;
  \draw[outline] (DL) -- (A) -- (DR) arc[start angle=-\tD, end angle=-180+\tD, x radius=\aD, y radius=\bD];

  \draw[rangle] (60:0.12) -- ++(150:0.12) -- ++(240:0.12);
  \draw[rangle] (-30:0.12) -- ++(60:0.12) -- ++(150:0.12);
  \draw[bdray, draw=primalconecolor] (A) -- (60:0.98);
  \draw[bdray, draw=dualconecolor] (A) -- (150:0.92);
  \draw[bdray, draw=polarconecolor] (A) -- (-30:0.92);
  \filldraw[apex] (A) circle (0.045);
  \node[anchor=west, inner sep=1pt, text=primalconecolor!75!black] at (1.16,1.96) {$\mathcal{K}$};
  \node[anchor=west, inner sep=1pt, text=dualconecolor!75!black] at (1.17,0.52) {$\mathcal{K}^*$};
  \node[anchor=west, inner sep=1pt, text=polarconecolor!75!black] at (1.17,-0.52) {$\mathcal{K}^\circ$};

  \begin{scope}[shift={(3.12,-0.25)}]
    \coordinate (R) at (0,0);
    \coordinate (Xin) at (-0.32,1.49);         %
    \coordinate (Xout) at (1.27,0.48);         %
    \coordinate (F) at ($(R)!(Xout)!(60:1)$);  %
    \coordinate (Xpol) at (-0.26,-0.55);       %
    \node at (0,2.42) {\textit{(2D section)}};
    \fill[primalconecolor!35] (R) -- (60:2.39) -- (120:2.39) -- cycle;
    \draw[primalconecolor, line width=0.6pt] (60:2.39) -- (R) -- (120:2.39);
    \fill[polarconecolor!45] (R) -- (-30:1.38) -- (-150:1.38) -- cycle;
    \draw[polarconecolor!75!black, line width=0.6pt, dashed] (-30:1.38) -- (R) -- (-150:1.38);
    \draw[proj] (Xout) -- (F);
    \draw[rangle] ($(F)+(60:0.12)$) -- ++(-30:0.12) -- ++(240:0.12);
    \draw[proj] (Xpol) -- (R);
    \filldraw[point] (Xin) circle (0.06);
    \filldraw[point] (Xout) circle (0.06);
    \filldraw[point] (Xpol) circle (0.06);
    \filldraw[apex] (R) circle (0.045);
    \node at (0.13,1.80) {$\Pi_{\mathcal{K}}(x) = x$};
    \node[anchor=west] at (0.40,0.10) {$\Pi_{\mathcal{K}}(x) \in \boundary\mathcal{K}$};
    \node at (0,-0.92) {$\Pi_{\mathcal{K}}(x) = (0,0)$};
  \end{scope}
\end{tikzpicture}
    \caption{\captitle{Circular cones and the projection operator.} \textit{Left:} A 3D circular cone $\mathcal{K}$, together with its dual cone $\mathcal{K}^*$ and polar cone $\mathcal{K}^\circ$. \textit{Right:} The action of the projection operator $\Pi_{\mathcal{K}}$ for the same cone, on three different parts of the domain: (i)~$x \in \mathcal{K}$, (ii)~$x \in \mathcal{K}^\circ$, and (iii)~$x \notin \mathcal{K} \cup \mathcal{K}^\circ$.}
    \label{fig:cone_vis}
    \Description{Left: a 3D circular cone opening upward, a wider dual cone around it, and a polar cone opening downward, all sharing the same apex. Right: a 2D section of the cone with three points and their projections: a point inside the cone projects onto itself, a point in the polar cone projects onto the apex, and a point outside both projects orthogonally onto the cone boundary.}
\end{figure}
A non-empty set $\mathcal{K}$ is called a \emph{convex cone} if it is closed under addition and $x \in \mathcal{K}$ implies $\alpha  x \in \mathcal{K}$ for all $\alpha  \geq 0$. For a convex cone $\mathcal{K}$, its \emph{dual cone} is defined as the set:
\begin{equation}
    \mathcal{K}^* = \left\{s: s^\trsp x \geq 0 \:\: \text{for all} \:\: x \in \mathcal{K}\right\}. \label{eq:dualcone}
\end{equation}
The \emph{polar cone} $\mathcal{K}^\circ = \left\{s: s^\trsp x \leq 0 \:\: \text{for all} \:\: x \in \mathcal{K}\right\}$ is the negative of the dual cone such that $\mathcal{K}^\circ = -\mathcal{K}^*$ holds. The \emph{projection operator} onto a convex set $\mathcal{C}$ is defined as:
\begin{equation*}
    \Pi_{\mathcal C}(x) = \argmin_{z \in \mathcal C} \| x - z\|.
\end{equation*}
By definition of $\mathcal{K}^*$ and $\mathcal{K}^\circ$, the relation $\Pi_{\mathcal{K}^\circ}(x) = - \Pi_{\mathcal{K}^*}(- x)$ holds. Another useful identity is Moreau's decomposition theorem, which states that $x = \Pi_{\mathcal{K}}(x) + \Pi_{\mathcal{K}^\circ}(x)$ holds.\footnote{For the special case of circular cones, the proof simply follows from the geometric observation that the boundaries of $\mathcal{K}$ and $\mathcal{K}^\circ$ are orthogonal, as can be seen in \cref{fig:cone_vis}.} This in turn implies that $\left\|\Pi_{\mathcal{K}}(x)\right\|^2 = \min_{w \in \mathcal{K}^*} \| x + w\|^2$ holds, that is, $\left\|\Pi_{\mathcal{K}}(x)\right\|_2$ equals the L2 distance of $x$ to the polar cone $\mathcal{K}^\circ$. This can be proved as:
\begin{align}
    &\left\| \Pi_{\mathcal{K}}(x) \right\|^2 \nonumber \\
    &= \left\| x - \Pi_{\mathcal{K}^\circ}(x) \right\|^2 \nonumber \\
    &= \left\| x + \Pi_{\mathcal{K}^*}(-x) \right\|^2 \nonumber \\
    & = \Big\| x + \big[\arg\min_{w \in \mathcal{K}^*} \| - x - w \|^2 \big] \Big\|^2 \nonumber \\
    & = \Big\| x + \big[\arg\min_{w \in \mathcal{K}^*} \| x + w \|^2 \big] \Big\|^2 \nonumber \\
    & = \min_{w \in \mathcal{K}^*} \| x + w\|^2. \nonumber
\end{align}

\begin{figure}[h]
    \centering
    \usetikzlibrary{arrows.meta,calc}%
\begin{tikzpicture}[
    font=\footnotesize,
    >={Stealth[length=3.5pt]},
    set/.style={fill=primalconecolor!35, draw=primalconecolor, line width=0.6pt},
    cone/.style={fill=primalconecolor!35},
    ncone/.style={fill=polarconecolor!45},
    lbl/.style={inner sep=0pt},
    normal/.style={->, line width=0.7pt, draw=polarconecolor!75!black},
    pointer/.style={line width=0.5pt, draw=geometrycolor},
    rangle/.style={line width=0.4pt, draw=geometrycolor},
    point/.style={fill=geometrycolor, draw=white, line width=0.6pt},
    ppoint/.style={fill=polarconecolor!75!black, draw=white, line width=0.5pt},
  ]
  \begin{scope}[shift={(-3.51,0.70)}]
    \coordinate (T) at (-0.94,0.73);
    \coordinate (X) at (0,0);
    \coordinate (B) at (-0.56,-1.31);
    \coordinate (L) at (-1.74,-1.09);
    \coordinate (E) at ($(X)!0.58!(B)$);
    \coordinate (NU) at ($(X)!1.14cm!-90:(T)$);   %
    \coordinate (NL) at ($(X)!1.17cm!90:(B)$);    %
    \coordinate (NE) at ($(E)!1.17cm!90:(B)$);    %
    \begin{scope}
      \clip (0,-1.2) rectangle (1.36,1.2);
      \fill[ncone, overlay] (X) -- ($(X)!3!(NU)$) -- ($(X)!3!(NU) + 3*(NL) - 3*(X)$) -- ($(X)!3!(NL)$) -- cycle;
    \end{scope}
    \filldraw[set] (T) -- (X) -- (B) -- (L) -- cycle;
    \node at (-1.38,-0.88) {$\mathcal{C}$};
    \draw[rangle] ($(X)!0.12cm!(T)$) -- ($(X)!0.12cm!(T)!0.12cm!90:(X)$) -- ($(X)!0.12cm!-90:(T)$);
    \draw[rangle] ($(X)!0.12cm!(B)$) -- ($(X)!0.12cm!(B)!0.12cm!-90:(X)$) -- ($(X)!0.12cm!90:(B)$);
    \draw[rangle] ($(E)!0.12cm!(X)$) -- ($(E)!0.12cm!(X)!0.12cm!90:(E)$) -- ($(E)!0.12cm!-90:(X)$);
    \draw[normal] (X) -- (NU);
    \draw[normal] (X) -- (NL);
    \draw[normal] (E) -- (NE);
    \filldraw[point] (X) circle (0.06);
    \filldraw[point] (E) circle (0.06);
    \node at ($(X)+(194:0.36)$) {$x_2$};
    \node at ($(E)+(157:0.33)$) {$x_1$};
    \node at (-0.26,0.86) {$\mathcal{N}_{\mathcal{C}}(x_2)$};
    \node at (0.10,-1.35) {$\mathcal{N}_{\mathcal{C}}(x_1)$};
    \draw[pointer] (0.18,0.72) to[bend left=35] (0.40,0.18);
    \draw[pointer] (-0.10,-0.95) to[bend right=40] (-0.06,-1.16);
  \end{scope}

  \draw[edgecolor, line width=0.4pt] (-1.78,-0.78) -- (-1.78,1.84);

  \coordinate (A) at (0,0);
  \coordinate (Xin) at (-0.18,1.00);          %
  \coordinate (P) at (60:1.35);               %
  \coordinate (Xout) at ($(P)+(-30:0.72)$);   %
  \coordinate (Xpol) at (0.32,-0.74);         %
  \fill[cone] (A) -- (60:1.80) -- (120:1.80) -- cycle;
  \fill[ncone] (A) -- (-30:1.20) -- (-150:1.20) -- cycle;
  \draw[polarconecolor, line width=0.5pt, densely dashed] (-30:1.42) -- (-30:1.74);
  \draw[polarconecolor, line width=0.5pt, densely dashed] (-150:1.42) -- (-150:1.74);
  \draw[->, line width=0.7pt, draw=primalconecolor] (A) -- (60:2.10);
  \draw[->, line width=0.7pt, draw=primalconecolor] (A) -- (120:2.10);
  \draw[normal] (A) -- (-30:1.42);
  \draw[normal] (A) -- (-150:1.42);
  \node at (0.17,0.62) {$\mathcal{K}$};
  \draw[line width=0.6pt, draw=geometrycolor, dashed] (Xpol) -- (A);
  \draw[rangle, draw=primalconecolor] (P) ++(60:0.12) -- ++(-30:0.12) -- ++(240:0.12);
  \draw[normal] (P) -- ++(-30:1.25);
  \filldraw[point] (Xin) circle (0.06);
  \filldraw[point] (Xout) circle (0.06);
  \filldraw[point] (Xpol) circle (0.06);
  \filldraw[ppoint] (P) circle (0.035);
  \filldraw[fill=white, draw=primalconecolor, line width=0.5pt] (A) circle (0.065);
  \fill[polarconecolor!75!black] (A) circle (0.032);
  \node[lbl, anchor=base] at (0.10,1.62) {$\mathcal{N}_{\mathcal{K}}(\Pi_{\mathcal{K}}(x))$};
  \node[lbl, anchor=base] at (0.10,1.33) {$= \{0\}$};
  \node[lbl, anchor=base west] at (-1.67,0.20) {$\mathcal{N}_{\mathcal{K}}(\Pi_{\mathcal{K}}(x))$};
  \node[lbl, anchor=base west] at (-1.67,-0.09) {$= \mathcal{K}^\circ$};
  \node[lbl, anchor=base east] at (1.85,0.20) {$\mathcal{N}_{\mathcal{K}}(\Pi_{\mathcal{K}}(x))$};
  \node[lbl, anchor=base east] at (1.85,-0.09) {$= \{c\,\hat{n} : c \geq 0\}$};
  \draw[pointer] (0.02,1.22) to[bend left=35] (-0.12,1.04);
  \draw[pointer] (-1.00,-0.13) to[bend right=35] (-0.55,-0.42);
  \draw[pointer] (0.98,0.48) to[bend left=35] (1.00,0.86);
\end{tikzpicture}%
    \caption{\captitle{Normal cones.}
    \textit{(Left)} The normal cone $\mathcal{N}_\mathcal{C}(x)$ for a 2D convex set $\mathcal{C}$. At the edge point $x_1$, $\mathcal{N}_\mathcal{C}(x_1)$ is the 1D positive span of the edge normal. At the corner $x_2$, $\mathcal{N}_\mathcal{C}(x_2)$ is the 2D positive span of the normals of the two edges that intersect at the corner. \textit{(Right)} Normal cones $\mathcal{N}_\mathcal{K}\left(\Pi_\mathcal{K}(x)\right)$ at three different projection points $\Pi_\mathcal{K}(x)$ on a 2D cone $\mathcal{K}$. When $x \in \interior \mathcal{K}$,  $\Pi_\mathcal{K}(x) = x$ and the normal cone only contains $0$. When $x \in \mathcal{K}^\circ$, $\Pi_\mathcal{K}(x) = 0$ and the normal cone is $\mathcal{K}^\circ$ itself. When $x \notin \mathcal{K}$ and $x \notin \mathcal{K}^\circ$, $\Pi_\mathcal{K}(x) \in \boundary \mathcal{K}$ and the normal cone is the 1D positive span of the unique vector $\hat{n}$ that satisfies $\hat{n}^\trsp \Pi_\mathcal{K}(x) = 0$ and points out of $\mathcal{K}$.}
    \label{fig:normal_cone_vis}
    \Description{Left: a 2D convex polygon with the normal cone at a point on an edge, a single ray along the edge normal, and at a corner, a shaded wedge between the normals of the two adjacent edges. Right: a 2D cone with three projection points: inside the cone the normal cone is only the zero vector, at the apex it is the polar cone, and on the boundary it is the ray along the outward normal.}
\end{figure}
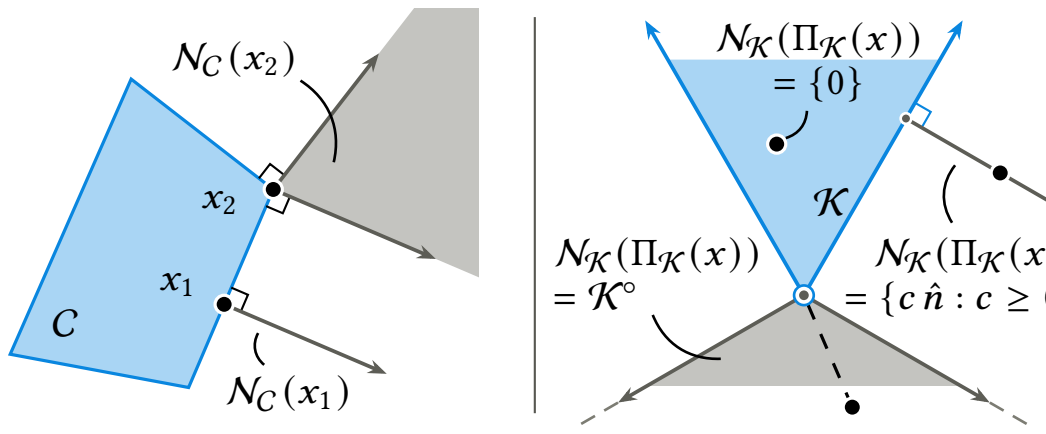
As shown in \cref{fig:normal_cone_vis}, the \emph{normal cone} of a convex set $\mathcal{C}$ at a point $x$ is defined as:
\begin{equation}
    \mathcal{N}_{\mathcal{C}}(x) = 
    \begin{cases}
        \left\{ n: n^\trsp (p - x) \leq 0 \:\: \text{for all} \:\: p \in \mathcal{C}\right\} & \text{if } x \in \boundary\mathcal{C}, \\
        \{0\} & \text{if } x \in \interior\mathcal{C}, \\
        \emptyset & \text{if } x \notin \mathcal{C}
    \end{cases},
\end{equation}
which is the set of all $n$ such that $x = \Pi_{\mathcal{C}}(x + n)$ holds for $x \in \mathcal C$. This implies the relation $x - \Pi_{\mathcal{C}}(x) \in \mathcal N_{\mathcal{C}}\left(\Pi_{\mathcal{C}}(x)\right)$ for an arbitrary point $x$ (not necessarily a member of $\mathcal{C}$), or equivalently stated as:
\begin{equation}
    \left\langle \Pi_{\mathcal{C}}(x) - x, p - \Pi_{\mathcal{C}}(x) \right\rangle \geq 0  \:\: \text{for all} \:\: p \in \mathcal{C}, \label{eq:proj_intuition}
\end{equation}
since the existence of $p \in \mathcal{C}$ with $\left\langle \Pi_{\mathcal{C}}(x) - x, p - \Pi_C(x) \right\rangle < 0$ implies the existence of a point on the line connecting $\Pi_{\mathcal{C}}(x)$ and $p$ that is closer to $x$ than $\Pi_{\mathcal{C}}(x)$, hence creating a contradiction.

For the L2 distance $d_{\mathcal{C}}(x) = \left\| x -  \Pi_{\mathcal{C}}(x)\right\|_2$, the relation \mbox{$\nabla \tfrac12 d_{\mathcal{C}}(x)^2 = x -  \Pi_{\mathcal{C}}(x)$} holds.\footnote{The proof is out of scope for this tutorial, but can be found in~\cite{kinderlehrer2000introduction}.}
Particularly for convex cones, combining this observation with the Moreau decomposition theorem gives another useful identity:
\begin{align}
    \nabla \tfrac{1}{2}\left\| \Pi_{\mathcal{K}}(x) \right\|^2
    &= \nabla \tfrac{1}{2}\left\| x -  \Pi_{\mathcal{K}^\circ}(x) \right\|^2 \nonumber \\
    &= x -  \Pi_{\mathcal{K}^\circ}(x) \nonumber \\
    &= \Pi_{\mathcal{K}}(x). \nonumber
\end{align}

\subsection{Circular Cones and Isotropic Friction}
\label{sec:circular_cone}

For a given normal direction $\hat n$, let us notate the normal and tangential components of a vector $x = [x_n, x_t]^\trsp$ as $x_n \in \mathbb R$ and $x_t \in \mathbb R^2$. The cone $\mathcal{K} = \left\{x \in \mathbb{R}^3:\left\|x_t\right\| \leq \mu x_n\right\}$, an ice cream cone with a circular base whose boundary consists of vectors with a constant slope $\norm{x_t}/x_n = \mu$, is called a \emph{circular cone}, with its dual again a circular cone with slope $1/\mu$. Circular cones are particularly relevant for this tutorial because they describe the admissible contact forces under \emph{isotropic Coulomb friction}.
Let us define the notation $\hat{x}_\perp \in \mathbb R^2$ to denote either of the two unit vectors that lie on the tangential plane of contact and satisfy $x_t^\trsp \hat{x}_\perp = 0$. For a circular cone $\mathcal K$, the normal cone $\mathcal{N_K}(x)$, the projection operator $\Pi_{\mathcal{K}}$, and its Jacobian $\partial{\Pi}_{\mathcal K}$ can all be evaluated in closed-form:\footnote{In general, the projection operator $\Pi_{\mathcal{K}}$ is only piece-wise differentiable and the Jacobian $\partial{\Pi}_{\mathcal K}$ is therefore a set-valued function. The closed-form expression for $\partial{\Pi}_{\mathcal K}$ presented here denotes the most common choice in practical implementations.}
\[
\hat t = \begin{cases}
x_t / \|x_t\| & \text{if } \|x_t\| \neq 0,\\ 
0 & \text{if } \|x_t\| = 0\\
\end{cases},
\]
\[
\mathcal N_{\mathcal K}(x) \;=\;
\begin{cases}
\mathcal K^\circ & \text{if } x = 0,\\ 
\left\{ \lambda\, [-\mu,\ \hat{t}] : \lambda \ge 0 \right\} & \text{if } x \in \boundary \mathcal K - \{0\}, \\
\{0\} & \text{if } x \in \interior\mathcal K, \\
\emptyset & \text{if } x \notin \mathcal K
\end{cases},
\]
\[
\hat g = \frac{[1,\ \mu \hat{t}]}{\sqrt{1+\mu^2}}, \qquad
\hat e_c = [0,\ \hat{x}_\perp], \qquad 
\hat{t}^\trsp \hat{x}_\perp = 0,
\]
\[
\Pi_{\mathcal{K}}(x) \;=\;
\begin{cases}
x & \text{if } x \in \mathcal K, \\
0 & \text{if } x \in \mathcal K^\circ, \\
\left(\hat{g}^\trsp x\right)\hat{g} & \text{otherwise}
\end{cases},
\]
\[
\partial{\Pi}_{\mathcal K}(x) \;=\;
\begin{cases}
\Id & \text{if } x \in \interior\mathcal K,\\[2pt]
0 & \text{if } x \in \interior\mathcal K^\circ \cup \{0\},\\[2pt]
\hat g\, \hat g^\trsp + \frac{\mu\,(x_n + \mu \|x_t\|)}{(1+\mu^2)\,\|x_t\|} \hat e_c \hat e_c^\trsp & \text{otherwise}
\end{cases},
\]
where all expressions are written in a normal-aligned coordinate frame where the $x$ axis points along the cone normal $\hat{n}$.

\subsection{Optimization Problems with Cone Constraints}
\label{sec:cone-constraints}
Let us consider a general cone-constrained convex program:
\begin{align}
    \min_{x} \:&f(x), \label{eq:cone_opt}\\
    \text{s.t.} \: &g(x) \in \mathcal{K}. \nonumber
\end{align}
It can be equivalently expressed in its Lagrangian form as:
\begin{align}
      \min_{x}\max_{s \in \mathcal{K}^*} \:&f(x) - s^\trsp g(x). \label{eq:alt_cone}
\end{align}
This is because whenever $g(x) \in \mathcal{K}$ holds, the definition of $ \mathcal{K}^*$ in \eqref{eq:dualcone} combined with the inner maximization implies that $s^\trsp g(x) = 0$ has to hold, and whenever $g(x) \not\in \mathcal{K}$, one can choose $s \in \mathcal{K}^*$ such that $- s^\trsp g(x) \to \infty$ holds in the limit.
Strong-duality holds for the cone-constrained convex programs considered in this tutorial~\cite{boyd2004convex}, meaning we can change the order of minimization and maximization in \eqref{eq:alt_cone} to obtain:
\begin{equation}
\max_{s\in\mathcal{K}^*} \min_{x} \: f(x) - s^\trsp g(x). \label{eq:alt_cone_dual}
\end{equation}
Converting the internal minimization with respect to $x$ in \eqref{eq:alt_cone_dual} into a stationarity condition by taking the gradient and setting it to zero, and combining it with the remaining conditions \mbox{$g(x) \in \mathcal{K}$} (primal-feasibility), $s \in \mathcal{K}^*$ (dual-feasibility), and $s^\trsp g(x) = 0$ (complementarity), we obtain the following optimality conditions:
\begin{mysubequations} \label{eq:cone_kkt}
\begin{align}
    \text{Find} \:\: &x, s: \nonumber \\ 
    &\nabla f(x) - J_g^\trsp(x)s = 0, \label{eq:cone_kkt_1} \\
    &g(x) \in \mathcal{K}, \label{eq:cone_kkt_2} \\
    &s\in\mathcal{K}^*, \label{eq:cone_kkt_3} \\
    &s^\trsp g(x) = 0. \label{eq:cone_kkt_4}
\end{align}
\end{mysubequations}

\subsection{Cone Complementarity Problems}
\label{sec:cone-complementarity}
For any nonlinear function $F(x)$, the relation $\mathcal{K} \ni  x \perp F(x)\in\mathcal{K}^*$ is called a \emph{nonlinear cone complementarity} condition. The special class with $F(x) = A x + b$ is called a \emph{linear cone complementarity} condition. The optimality conditions \crefrange{eq:cone_kkt_1}{eq:cone_kkt_4} for a cone-constrained convex program can be equivalently rewritten using a cone complementarity condition as
\begin{mysubequations} \label{eq:cone_kkt_short}
\begin{align}
    \text{Find} \:\: &x, s: \nonumber \\ 
    &\nabla f(x) - J_g^\trsp(x)s = 0, \label{eq:cone_kkt_1_short} \\
    &\mathcal{K} \ni  g(x) \perp s\in\mathcal{K}^*. \label{eq:cone_kkt_2_short}
\end{align}
\end{mysubequations}
When $g(x) = x$, this reduces to
\begin{align}
    \text{Find} \:\: &x: \nonumber \\ 
    &\mathcal{K} \ni  x \perp \nabla f(x)\in\mathcal{K}^*, \label{eq:cone_kkt_gI}
\end{align}
a nonlinear cone complementarity problem. 

\smallbreak
\textbf{Complementarity Functions.} Complementarity conditions can be reformulated as equality conditions using so-called \emph{complementarity functions}. This reformulation allows standard Newton-type methods~\cite{boyd2004convex} to be applied to the numerical solution of cone complementarity problems~\cite{fischer1992special}. In the context of this tutorial, complementarity functions are particularly relevant to the \say{Root Finding} transcription of contact dynamics (\cref{sec:root_finding}) and its associated solution routines (\cref{alg:nse-newton}).\footnote{Throughout this tutorial, the term \say{transcription} is used to refer to a particular choice of mathematical expressions used to represent a set of conditions. The same conditions may admit multiple, mathematically equivalent transcriptions.}

For a scalar complementarity condition $0 \leq x \perp y \geq 0$ on the non-negative cone $\mathbb R^+$, a complementarity function is any function $\psi(x, y): \mathbb{R}^2 \to \mathbb R$ satisfying
\begin{equation*}
    \psi(x,y) = 0 \iff 0 \leq x \perp y \geq 0.
\end{equation*} 
One example is the minimum function $\psi(x, y) = \min(x,y)$. Since it is non-smooth wherever $x = y$, it can be replaced by any of a large number of smooth approximations~\cite{smoothmin}. A more commonly adopted alternative is the \emph{Fischer--Burmeister (FB)} function~\cite{fischer1992special}
\begin{equation*}
    \psi(x, y) = x + y - \sqrt{x^2 + y^2},
\end{equation*}
which is smooth everywhere except at $(0,0)$. The same approach can be extended to a circular cone $\mathcal{K}$ by making use of the \emph{second-order-cone Fischer--Burmeister (SOCFB) } function~\cite{fukushima2002smoothing}. For the 3D case, the SOCFB function is defined for the Lorentz cone \mbox{$\mathcal{L} = \left\{x \in \mathbb{R}^3 : \|x_t\| \leq x_n\right\}$}, the circular cone with unit slope, which is self-dual ($\mathcal{L}^* = \mathcal{L}$). For a circular cone $\mathcal{K} = \left\{x \in \mathbb{R}^3:\left\|x_t\right\| \leq \mu x_n\right\}$ and any two vectors $v \in \mathcal{K}$ and $w \in \mathcal{K}^*$, the re-scaling is given by
\begin{equation}\label{eq:lorentz_map}
    x = (\mu v_{n},\; v_{t}), \qquad y = (w_{n},\; \mu w_{t}).
\end{equation}
This defines a one-to-one map $(x,y)=h(v, w)$ onto $\mathcal{L}\times \mathcal{L}$. A cone complementarity condition $\mathcal{K} \ni v \perp w \in \mathcal{K}^*$ is therefore equivalent to the condition $\mathcal{L} \ni x \perp y \in \mathcal{L}$. The Lorentz cone is equipped with the \emph{Jordan product} operator $x \circ y = (x^\trsp y,\; x_n y_t + y_n x_t)$, under which every $z \in \mathcal{L}$ has a unique square root in $\mathcal{L}$:
\begin{align}
    &\lambda_{1,2} = z_n \mp \|z_t\|\: , \: \hat{z}_t = z_t / \|z_t\|, \\
    &\sqrt{z} = \tfrac12\big(\sqrt{\lambda_2} + \sqrt{\lambda_1},\; \left(\sqrt{\lambda_2} - \sqrt{\lambda_1}\right)\,\hat{z}_t\big).
\end{align}
The SOCFB function is then defined as
\begin{equation}\label{eq:soc_fb}
    \psi(x, y) = x + y - \sqrt{x \circ x + y \circ y},
\end{equation}
which is smooth everywhere except when $x \circ x + y \circ y \in bd \mathcal{L}$ holds~\cite{fukushima2002smoothing}. Similar to its scalar counterpart, it satisfies:
\begin{equation*}
    \psi(x,y) = 0 \iff \mathcal{L} \ni x \perp y \in \mathcal{L}.
\end{equation*} 
If needed, smooth approximations to the FB and SOCFB functions can also be easily defined using a smoothing factor $\epsilon$ as:
\begin{align}
    \psi_\epsilon(x, y) &= x + y - \sqrt{x^2 + y^2 + \epsilon^2}, \\
    \psi_\epsilon(x, y) &= x + y - \sqrt{x \circ x + y \circ y + \epsilon^2[1, 0, 0]} \label{eq:socfb_smooth}.
\end{align}
Using the previously defined rescaling map \eqref{eq:lorentz_map}, the SOCFB function can be extended to arbitrary circular cones as: 
\begin{equation}\label{eq:socfb}
    \psi(h(v,w)) = 0 \iff \mathcal{K} \ni v \perp w \in \mathcal{K}^*.
\end{equation} 
\usetikzlibrary{arrows.meta,backgrounds,fit}
\providecolor{ourportwine}{RGB}{128,40,60}
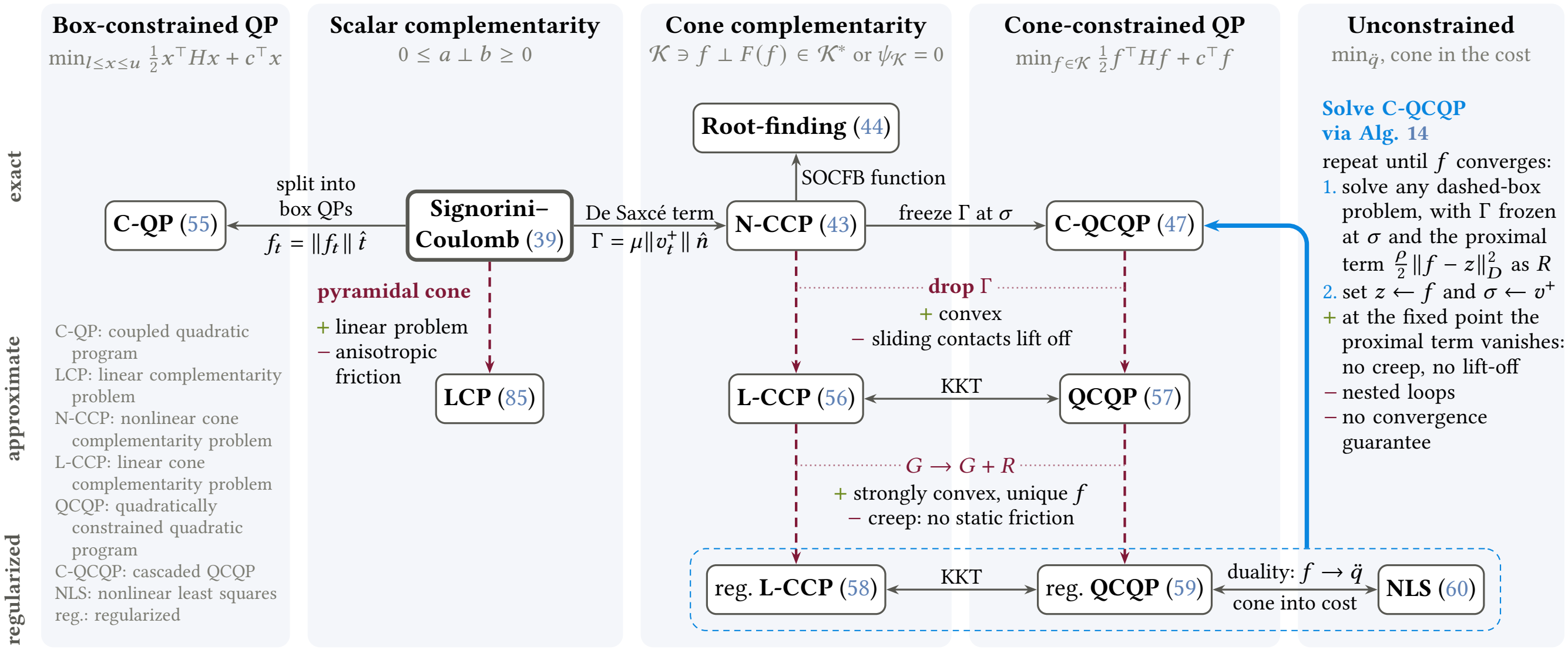
\begin{figure*}[t]
\centering
\newcommand{\ovpro}[1]{\par\hangindent=0.9em\hangafter=1\noindent\makebox[0.9em][l]{\textcolor{ourdarkgreen2}{$+$}}#1\par}
\newcommand{\ovcon}[1]{\par\hangindent=0.9em\hangafter=1\noindent\makebox[0.9em][l]{\textcolor{ourportwine}{$-$}}#1\par}
\newcommand{\ovstep}[2]{\par\hangindent=0.9em\hangafter=1\noindent\makebox[0.9em][l]{\textcolor{ourblue2}{#1.}}#2\par}
\newcommand{\ovabbr}[2]{\par\hangindent=0.9em\hangafter=1\noindent #1:~ #2\par}
\def\ovyRF{-0.95}\def\ovyE{-2.05}\def\ovyA{-4.0}\def\ovyR{-6.15}\def\ovyB{-6.8}
\begin{tikzpicture}[
    font=\footnotesize,
    band/.style={fill=ourblue!7, rounded corners=4pt},
    card/.style={fill=white, draw=ourlightgray2, rounded corners=3pt},
    header/.style={anchor=north, align=center, inner sep=1.5pt},
    tr/.style={draw=ourdarkgray2, fill=white, rounded corners=3pt, line width=0.6pt,
               font=\small, align=center, inner sep=2.5pt, minimum height=0.55cm},
    root/.style={line width=1.2pt},
    rewrite/.style={-{Stealth[length=4pt]}, line width=0.6pt, draw=ourdarkgray2},
    equiv/.style={{Stealth[length=4pt]}-{Stealth[length=4pt]}, line width=0.6pt, draw=ourdarkgray2},
    approx/.style={-{Stealth[length=4pt]}, line width=0.8pt, draw=ourportwine, dash pattern=on 3pt off 2pt},
    outer/.style={-{Stealth[length=5pt]}, line width=1.4pt, draw=ourblue2, rounded corners=6pt},
    lbl/.style={align=center, inner sep=1.5pt},
    steptitle/.style={font=\footnotesize\bfseries, text=ourportwine, inner sep=1.5pt},
    leader/.style={draw=ourportwine!60, line width=0.5pt, densely dotted},
    rowlbl/.style={rotate=90, font=\footnotesize\bfseries, text=ourdarkgray2},
    cardtxt/.style={anchor=north west, align=left, inner sep=0pt, text width=#1,
                    execute at begin node={\hyphenpenalty=10000\exhyphenpenalty=10000}},
  ]

  \begin{scope}[on background layer]
    \foreach \xa/\xb in {0.55/3.35, 3.55/7.10, 7.30/10.80, 11.00/14.50, 14.70/17.70}
      \fill[band] (\xa,0.45) rectangle (\xb,\ovyB);
  \end{scope}
  \node[header] at (1.95,0.38)   {{\small\bfseries Box-constrained QP}\\[1pt]{\color{ourgray2}$\min_{l\le x\le u}\tfrac12 x^\trsp H x+c^\trsp x$}};
  \node[header] at (5.325,0.38)  {{\small\bfseries Scalar complementarity}\\[1pt]{\color{ourgray2}$0\le a\perp b\ge 0$}};
  \node[header] at (9.05,0.38)   {{\small\bfseries Cone complementarity}\\[1pt]{\color{ourgray2}$\mathcal{K}\ni f\perp F(f)\in\mathcal{K}^*$ or $\psi_{\mathcal{K}}=0$}};
  \node[header] at (12.75,0.38)  {{\small\bfseries Cone-constrained QP}\\[1pt]{\color{ourgray2}$\min_{f\in\mathcal{K}}\tfrac12 f^\trsp H f+c^\trsp f$}};
  \node[header] at (16.20,0.38)  {{\small\bfseries Unconstrained}\\[1pt]{\color{ourgray2}$\min_{\ddot{q}}$, cone in the cost}};

  \node[rowlbl] at (0.25,{(\ovyRF+\ovyE)/2}) {exact};
  \node[rowlbl] at (0.25,\ovyA) {approximate};
  \node[rowlbl] at (0.25,\ovyR) {regularized};

  \node[tr] (CQP)      at (1.95,\ovyE)   {\textbf{C-QP} \eqref{eq:cqp}};
  \node[tr, root] (SC) at (5.60,\ovyE)   {\textbf{Signorini--}\\\textbf{Coulomb} \eqref{eq:mcp}};
  \node[tr] (NCCP)     at (9.05,\ovyE)   {\textbf{N-CCP} \eqref{eq:nccp}};
  \node[tr] (RF)       at (9.05,\ovyRF)  {\textbf{Root-finding} \eqref{eq:root_finding}};
  \node[tr] (CQCQP)    at (12.75,\ovyE)  {\textbf{C-QCQP} \eqref{eq:c_qcqp}};
  \node[tr] (LCP)      at (5.60,\ovyA)   {\textbf{LCP} \eqref{eq:lcp}};
  \node[tr] (LCCP)     at (9.05,\ovyA)   {\textbf{L-CCP} \eqref{eq:lccp_unreg}};
  \node[tr] (QCQP)     at (12.75,\ovyA)  {\textbf{QCQP} \eqref{eq:socqp_unreg}};
  \node[tr] (LCCPR)    at (9.05,\ovyR)   {reg.\ \textbf{L-CCP} \eqref{eq:lccp}};
  \node[tr] (QCQPR)    at (12.75,\ovyR)  {reg.\ \textbf{QCQP} \eqref{eq:socqp}};
  \node[tr] (NLS)      at (16.20,\ovyR)  {\textbf{NLS} \eqref{eq:nls}};

  \draw[rewrite] (SC) -- node[lbl, above] {split into\\box QPs} node[lbl, below] {$f_t=\|f_t\|\,\hat{t}$} (CQP);
  \draw[rewrite] (SC) -- node[lbl, above] {De Saxc\'e term} node[lbl, below] {$\Gamma=\mu\|v^+_t\|\,\hat{n}$} (NCCP);
  \draw[rewrite] (NCCP) -- node[lbl, right] {SOCFB function} (RF);
  \draw[rewrite] (NCCP) -- node[lbl, above] {freeze $\Gamma$ at $\sigma$} (CQCQP);
  \draw[equiv] (LCCP) -- node[lbl, above] {KKT} (QCQP);
  \draw[equiv] (LCCPR) -- node[lbl, above] {KKT} (QCQPR);
  \draw[equiv] (QCQPR) -- node[lbl, above] {duality: $f\to\ddot{q}$} node[lbl, below] {cone into cost} (NLS);

  \draw[approx] (SC) -- (LCP);
  \draw[approx] (NCCP) -- (LCCP);
  \draw[approx] (CQCQP) -- (QCQP);
  \draw[approx] (LCCP) -- (LCCPR);
  \draw[approx] (QCQP) -- (QCQPR);
  \node[steptitle, anchor=east] (pyrT) at (5.45,{\ovyE-0.75}) {pyramidal cone};
  \node[cardtxt=1.8cm, anchor=north east] at (pyrT.south east) {%
    \ovpro{linear problem}
    \ovcon{anisotropic friction}};
  \node[steptitle] (dropT) at (10.90,{\ovyE-0.7}) {drop $\Gamma$};
  \node[lbl, anchor=north] at (dropT.south) {\textcolor{ourdarkgreen2}{$+$} convex\\\textcolor{ourportwine}{$-$} sliding contacts lift off};
  \draw[leader] (dropT.west) -- (9.05,{\ovyE-0.7});
  \draw[leader] (dropT.east) -- (12.75,{\ovyE-0.7});
  \node[steptitle] (regT) at (10.90,{\ovyA-0.75}) {$G\to G+R$};
  \node[lbl, anchor=north] at (regT.south) {\textcolor{ourdarkgreen2}{$+$} strongly convex, unique $f$\\\textcolor{ourportwine}{$-$} creep: no static friction};
  \draw[leader] (regT.west) -- (9.05,{\ovyA-0.75});
  \draw[leader] (regT.east) -- (12.75,{\ovyA-0.75});

  \node[cardtxt=2.55cm, font=\scriptsize, text=ourgray2] at (0.70,{\ovyE-1.0}) {%
    \ovabbr{C-QP}{coupled quadratic program}
    \ovabbr{LCP}{linear complementarity problem}
    \ovabbr{N-CCP}{nonlinear cone complementarity problem}
    \ovabbr{L-CCP}{linear cone complementarity problem}
    \ovabbr{QCQP}{quadratically constrained quadratic program}
    \ovabbr{C-QCQP}{cascaded QCQP}
    \ovabbr{NLS}{nonlinear least squares}
    \ovabbr{reg.}{regularized}};

  \node[fit=(LCCPR)(QCQPR)(NLS), draw=ourblue2, dash pattern=on 2pt off 1.5pt,
        rounded corners=5pt, inner sep=5pt] (inner) {};
  \draw[outer] (14.80,0 |- inner.north) -- (14.80,\ovyE) -- (CQCQP.east);
  \node[cardtxt=2.7cm] at (14.97,-0.62) {%
    \noindent{\bfseries\color{ourblue2}Solve C-QCQP\\ via Alg.~\ref{alg:proximal-point-correction}}
    \par\vspace{1pt}\noindent repeat until $f$ converges:
    \ovstep{1}{solve any dashed-box problem, with $\Gamma$ frozen at $\sigma$ and the proximal term $\tfrac{\rho}{2}\|f-z\|_D^2$ as $R$}
    \ovstep{2}{set $z\gets f$ and $\sigma\gets v^+$}
    \ovpro{at the fixed point the proximal term vanishes: no creep, no lift-off}
    \ovcon{nested loops}
    \ovcon{no convergence guarantee}};
\end{tikzpicture}
\caption{\captitle{Overview of the transcriptions.} Columns group the transcriptions by mathematical form, and rows by physical fidelity: exact, approximate, and regularized. The Signorini--Coulomb transcription is the pedagogical starting point chosen in this tutorial. Solid arrows are exact rewrites that preserve the solution set. Dashed red arrows are approximations that change the physics, annotated with what they gain ($+$) and lose ($-$). The blue arrow is the outer loop of the final proximal blueprint proposed in this tutorial (Alg.~\ref{alg:proximal-point-correction}), which recovers the exact C-QCQP transcription from any solver for the regularized transcriptions in the dashed box. The abbreviations are spelled out in the first column, and \cref{tab:cmp_algorithms} presents the solution algorithms for each transcription.}
\label{fig:overview}
\end{figure*}

\smallbreak
\textbf{Differentiating Complementarity Functions.} The root-finding formulation introduced in the next section (\cref{sec:root_finding}) expresses contact complementarity as a system of nonlinear equations. Newton-type methods solve this system by repeatedly linearizing it, which requires the Jacobian of the complementarity residual (\cref{alg:nse-newton}). We therefore derive the partial derivatives of the SOCFB function; the chain rule then connects them to the contact forces and velocities. To simplify this differentiation, we write the Jordan product in matrix-vector form:
\begin{equation} \label{eq:arrow}
    x \circ y = \operatorname{Arw}(x)\, y = \operatorname{Arw}(y)\, x, \qquad
    \operatorname{Arw}(x) = \begin{bmatrix} x_n & x_t^\trsp \\ x_t & x_n I_2 \end{bmatrix},
\end{equation}
where $\operatorname{Arw}(x)$ is called the \emph{arrow matrix} of $x$. The Jacobians of the Jordan product with respect to its two arguments are therefore
\begin{equation}
    \partial_x (x \circ y) = \operatorname{Arw}(y), \qquad \partial_y (x \circ y) = \operatorname{Arw}(x),
\end{equation}
and in particular $\partial_x (x \circ x) = 2\operatorname{Arw}(x)$. The arrow matrix is symmetric. For $x_t \neq 0$, its eigenvalues are $x_n + \|x_t\|$ and $x_n - \|x_t\|$ with the eigenvectors $\left(1, \hat{x}_t\right)$ and $\left(1, -\hat{x}_t\right)$, and $x_n$ with the eigenvector $(0, u)$ for a unit vector $u$ orthogonal to $x_t$. It is therefore positive definite for $x$ in the interior of $\mathcal{L}$ and singular on its boundary.
Differentiating $s \circ s = x \circ x + y \circ y$, where $s = \sqrt{x \circ x + y \circ y}$, gives $\operatorname{Arw}(s)\, ds = \operatorname{Arw}(x)\, dx + \operatorname{Arw}(y)\, dy$. Wherever $s$ lies in the interior of $\mathcal{L}$, the Jacobians of the SOCFB function $\psi(x, y) = x + y - s$ can therefore be computed in closed-form as
\begin{align} \label{eq:socfb_jac}
    \partial_x \psi &= I - \operatorname{Arw}(s)^{-1}\operatorname{Arw}(x), \\
    \partial_y \psi &= I - \operatorname{Arw}(s)^{-1}\operatorname{Arw}(y).
\end{align}
Defining $s_\epsilon = \sqrt{x \circ x + y \circ y + \epsilon^2[1, 0, 0]}$, the corresponding Jacobians for the smoothed versions are simply
\begin{align} \label{eq:socfb_jac_eps}
    \partial_x \psi_\epsilon &= I - \operatorname{Arw}(s_\epsilon)^{-1}\operatorname{Arw}(x), \\
    \partial_y \psi_\epsilon &= I - \operatorname{Arw}(s_\epsilon)^{-1}\operatorname{Arw}(y).
\end{align}

\begin{figure*}[t]
    \centering
    \usetikzlibrary{arrows.meta,calc}
\begin{tikzpicture}[
    x=0.826mm, y=0.826mm, font=\footnotesize,
    >={Stealth[length=3.5pt]},
    title/.style={font=\small\bfseries, inner sep=0pt},
    panel/.style={draw=edgecolor, line width=0.5pt, rounded corners=2pt},
    orbadge/.style={draw=edgecolor, fill=fixedbodycolor!30, line width=0.5pt,
                    rounded corners=2pt, inner xsep=2pt, inner ysep=1.5pt},
    plane/.style={fill=fixedbodycolor!30},
    surface/.style={draw=edgecolor, line width=0.6pt},
    force/.style={->, line width=0.9pt, draw=forcecolor},
    forcebound/.style={->, line width=0.6pt, draw=forcecolor!50},
    vel/.style={->, line width=0.8pt, draw=velocitycolor},
    dist/.style={->, line width=0.5pt, draw=geometrycolor},
    helper/.style={line width=0.4pt, draw=annotationcolor, densely dashed},
    axis/.style={line width=0.4pt, draw=geometrycolor},
    coneedge/.style={draw=primalconecolor!75!black, line width=0.6pt},
    note/.style={font=\footnotesize\itshape, align=left, inner sep=0pt},
    pointer/.style={-{Straight Barb[length=2pt]}, line width=0.4pt, draw=annotationcolor},
    eq/.style={anchor=north west, inner sep=0pt},
    startpt/.style={fill=white, draw=geometrycolor, line width=0.3pt},
    contactpt/.style={fill=forcecolor!30, draw=geometrycolor, line width=0.3pt},
    frictioncone/.pic={
      \fill[plane] (-1.74,-7.92) -- (22.25,0.05) -- (1.24,7.80) -- (-22.54,0.48) -- cycle;
      \shade[left color=primalconecolor!12, right color=primalconecolor!40]
        (0,0) -- (10.607,19.108) arc[start angle=-9.74, end angle=189.74, x radius=10.76, y radius=3.33] -- cycle;
      \draw[axis] (0,25.6) -- (0,0);
      \shade[left color=primalconecolor!45, right color=primalconecolor!12, fill opacity=0.85]
        (0,0) -- (-10.607,19.108) arc[start angle=189.74, end angle=350.26, x radius=10.76, y radius=3.33] -- cycle;
      \draw[helper] (0,0) -- (10.63,-0.57);
      \draw[helper] (7.56,13.29) -- (7.56,-0.41);
      \draw[coneedge] (0,19.67) ellipse [x radius=10.76, y radius=3.33];
      \draw[coneedge, line join=round] (-10.607,19.108) -- (0,0) -- (10.607,19.108);
      \draw[force] (0,0) -- (0,13.7);
      \draw[forcebound] (0,13.7) -- (7.56,13.29);
      \node[text=forcecolor!75!black, left, inner sep=1.5pt] at (0,11.2) {$f_n$};
      \node[text=forcecolor, right, inner sep=1pt] at (8.4,14.0) {$\mu f_n$};
    },
  ]
  \node[title, anchor=north west] at (1.3,43.0) {Non-penetration constraint};
  \draw[panel] (3.8,20.92) rectangle (32.9,38.58);
  \draw[panel] (35.0,20.92) rectangle (64.1,38.58);
  \node[orbadge] at (33.95,29.75) {OR};
  \foreach \ox in {6.4,39.6} {
    \begin{scope}[shift={(\ox,31.17)}]
      \shade[top color=fixedbodycolor, bottom color=white] (0.7,0) rectangle (21.0,-9.6);
      \draw[surface] (0,0) -- (21.8,0);
      \draw[helper] (3.25,-7.54) -- (10.05,-7.54);
      \draw[dist, shorten >=0.55pt] (3.25,0) -- (3.25,-7.0);
      \fill[startpt] (3.25,-7.54) circle (0.55);
      \node at (5.3,-3.6) {$d$};
      \node[text=velocitycolor] at (15.4,-4.0) {$\Delta t\, v_n^+$};
    \end{scope}
  }
  \begin{scope}[shift={(6.4,31.17)}]
    \coordinate (P) at (10.05,3.66);
    \draw[vel] (10.05,-7.54) -- ($(P)+(0,-0.6)$);
    \fill[contactpt] (P) circle (0.55);
    \node[right, inner sep=1pt] at (11.6,3.66) {$f_n = 0$};
    \node[note, anchor=north west] at (-1.8,6.5) {tangent\\plane};
    \draw[pointer] (5.84,3.52) .. controls (6.2,3.13) and (7.52,2.52) .. (5.82,0.6);
  \end{scope}
  \begin{scope}[shift={(39.6,31.17)}]
    \coordinate (P) at (10.05,0);
    \draw[vel] (10.05,-7.54) -- ($(P)+(0,-0.6)$);
    \draw[force] ($(P)+(0,0.5)$) -- (10.05,6.6);
    \fill[contactpt] (P) circle (0.55);
    \node[right, inner sep=1pt] at (11.6,3.9) {$f_n \geq 0$};
  \end{scope}

  \node[title, anchor=north west, align=left] at (1.3,19.0) {Maximum dissipativity\\constraint};
  \fill[plane] (45.37,0.65) -- (69.36,8.62) -- (48.35,16.37) -- (24.57,9.06) -- cycle;
  \coordinate (C) at (47.1,8.6);
  \draw[axis, dash pattern=on 3pt off 1pt on 0.6pt off 1pt] (47.1,17.4) -- (C);
  \draw[vel] (C) -- ++(177.4:16.9);
  \draw[force] (C) -- ++(-2.6:7.4);
  \draw[->, line width=0.4pt, draw=geometrycolor, rotate around={-2.6:(C)}]
    ($(C)+(3.9,0)$) arc[start angle=0, end angle=180, x radius=3.9, y radius=1.35];
  \fill[contactpt] (C) circle (0.45);
  \node[text=velocitycolor] at (38.0,12.0) {$v_t^+$};
  \node[text=forcecolor] at (50.4,5.4) {$f_t$};
  \node at (54.6,11.6) {$180^\circ$};
  \node[anchor=west, inner sep=0pt] at (3.5,4.4) {$\|v_t^+\|\,\|f_t\| = -f_t^\trsp v_t^+$};

  \node[title, anchor=north] at (95.08,43.2) {Sticking contact};
  \draw[panel] (70.23,2.73) rectangle (119.94,38.54);
  \pic at (96.84,11.38) {frictioncone};
  \begin{scope}[shift={(96.84,11.38)}]
    \draw[force] (0,0) -- (3.9,-0.21);
    \fill[contactpt] (0,0) circle (0.45);
    \node[text=forcecolor] at (2.6,-3.4) {$f_t$};
  \end{scope}
  \node[eq] at (72.3,37.1) {$\|f_t\| \leq \mu f_n$};
  \node[eq] at (72.3,32.4) {$v_t^+ = 0$};
  \node[note, anchor=north west] at (109.9,37.2) {friction\\cone};
  \draw[pointer] (109.48,34.79) .. controls (109.04,34.49) and (108.36,34.67) .. (107.39,32.8);
  \node[note, anchor=north west] at (71.8,20.7) {contact\\plane};
  \draw[pointer] (73.82,15.3) .. controls (73.94,14.91) and (73.73,14.9) .. (74.96,12.86);
  \node[note, anchor=north west] at (71.2,9.4) {Can point\\in any direction};
  \draw[pointer] (92.4,5.0) .. controls (94.6,5.0) and (94.9,7.2) .. (96.2,9.2);

  \node[title, anchor=north] at (146.92,43.2) {Slipping contact};
  \draw[panel] (122.0,2.73) rectangle (171.71,38.54);
  \pic at (148.68,11.38) {frictioncone};
  \begin{scope}[shift={(148.68,11.38)}]
    \draw[force] (0,0) -- (7.56,-0.41);
    \fill[contactpt] (0,0) circle (0.45);
    \node[text=forcecolor] at (2.6,-3.4) {$f_t$};
  \end{scope}
  \node[eq] at (124.1,37.4) {$\mu f_n - \|f_t\| = 0$};
  \node[eq] at (124.1,32.7) {$v_t^+ \neq 0$};
  \node[note, anchor=west] at (124.0,4.9) {Opposes $v_t^+$};
  \draw[pointer] (142.0,4.8) .. controls (144.9,5.0) and (146.6,6.0) .. (148.1,9.2);

  \node[orbadge] at (120.97,21.23) {OR};
\end{tikzpicture}
    \caption{\captitle{Visualization of the Signorini--Coulomb conditions~\eqref{eq:mcp} for a single contact.} The non-penetration constraint allows a normal force $f_n > 0$ only if the contact point lands exactly on the surface after symplectic Euler integration, i.e., $d + \Delta t\, v_n^+ = 0$; if it ends above the surface, $f_n = 0$. The maximum dissipativity constraint requires the friction force $f_t$ to oppose the tangential end-of-step velocity $v_t^+$. The contact is ``sticking'' if $v_t^+ = 0$ and the contact force lies within the friction cone, $\|f_t\| \leq \mu f_n$. The contact is ``slipping'' if $v_t^+ \neq 0$ and the contact force lies on the boundary of the friction cone, $\|f_t\| = \mu f_n$.
    }
    \label{fig:mcp}
    \Description{Three panels. The first shows the non-penetration constraint as two alternatives, a contact point ending above the tangent plane with zero normal force or landing exactly on it with a nonnegative normal force, and below it the maximum dissipativity constraint with the friction force opposing the tangential velocity. The other two panels show a friction cone above the contact plane: for a sticking contact the tangential velocity is zero and the force lies inside the cone; for a slipping contact the force lies on the cone boundary and its tangential part opposes the tangential velocity.}
\end{figure*}
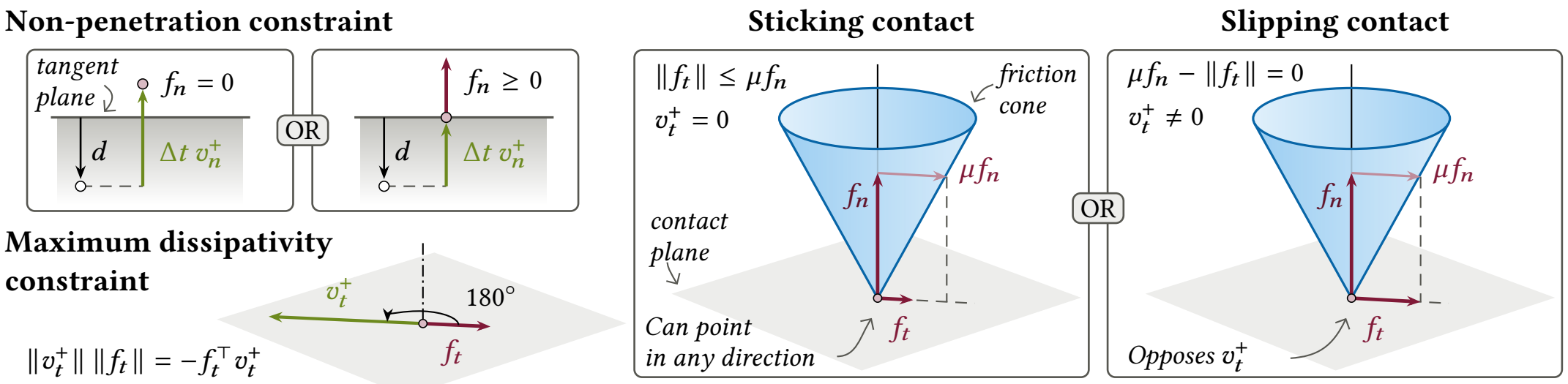
\section{Problem Formulation}
\label{sec:formulation}

Discrete time-stepping turns contact dynamics into an algebraic problem for the contact forces and end-of-step velocities. This problem admits several mathematical transcriptions, each exposing a different structure that can be exploited by numerical solvers. This section develops the main formulations used in robotics and simulation and explains their relationships, summarized in \cref{fig:overview}.

\smallskip
The starting point is the nonsmooth contact dynamics framework of Moreau and Jean~\cite{moreau1994numerical, Jean1999}, which provides a numerically tractable discretization of Moreau's measure differential inclusion~\cite{moreau1983liaisons, moreau1985standard, Moreau1988}. We distinguish two families: \emph{exact transcriptions}, which preserve the solution set of the discrete contact problem, and \emph{convex relaxations}, which modify the contact conditions to obtain problems that are easier to solve. This distinction separates changes in mathematical representation from changes in the modeled contact behavior.

\smallskip
We begin with the \emph{Signorini--Coulomb} transcription~\cite{moreau1994numerical, Jean1999, acary2018solving}, whose equations directly express non-penetration and frictional dissipation. We then derive the \emph{Nonlinear Cone Complementarity Problem (N-CCP)} transcription, building on the work of de Saxc\'e and Feng~\cite{DeSaxceFeng1991, DeSaxceFeng1998} and using the terminology of~\cite{lelidec2023contacts, carpentier2024compliant}. Deriving it from the Signorini--Coulomb conditions connects the abstract cone complementarity relation to its mechanical meaning. Complementarity functions then yield an equivalent root-finding formulation.

\smallskip
The remaining exact transcriptions connect contact dynamics to convex optimization. The \emph{Cascaded Quadratically Constrained Quadratic Program (C-QCQP)} formulation follows Acary~\cite{acary2011formulation, acary2018solving}, with terminology from~\cite{carpentier2024compliant}; the \emph{Coupled Quadratic Program (C-QP)} formulation builds on Panagiotopoulos~\cite{panagiotopoulos1975nonlinear} and Kaufman et al.~\cite{kaufman2008staggered}. Both express the contact problem through interdependent convex subproblems. Their coupling retains the difficulty of the original problem while also showing how to construct convex relaxations.

\smallskip
We next introduce the \emph{Linear Cone Complementarity Problem (L-CCP)} and \emph{Quadratically Constrained Quadratic Program (QCQP)} relaxations, developed from Anitescu's work~\cite{anitescu2004constraint, anitescu2006optimization} and named following~\cite{lelidec2023contacts, carpentier2024compliant}. They follow from the N-CCP and C-QCQP formulations by dropping selected terms. Finally, we derive the \emph{Nonlinear Least-Squares (NLS)} formulation associated with Todorov's work~\cite{todorov2014convex, MuJoCo150Release}, also called the reduced primal form~\cite{MuJoCo150Release, castro_sim}. Its derivation is more involved, but it enables efficient numerical methods used in MuJoCo and Drake. We close by examining the physical implications of these relaxations. \cref{sec:history} elaborates on the historical context surrounding these transcriptions.

\subsection{Exact Transcriptions}

We present five equivalent ways to express the same discrete frictional contact law. Starting from the mechanical Signorini--Coulomb conditions, we recast the problem as nonlinear cone complementarity and then as a root-finding problem. The C-QCQP and C-QP transcriptions subsequently express it through cascaded or coupled convex subproblems (although the individual convexity of such subproblems does not make the full problem convex). Each representation exposes a different structure for numerical solution while preserving the admissible contact forces and velocities.

\subsubsection{Signorini–Coulomb Transcription}
A natural starting point for describing complementarity-based contact dynamics is the \say{Signorini–Coulomb conditions}, which combine equations with scalar complementarity conditions~\cite{moreau1994numerical, Jean1999, acary2018solving}:
\begin{equationbox}{Signorini–Coulomb Transcription}
\begin{mysubequations} \label{eq:mcp}
\begin{align}
    \text{Find} \:\: & v^+, f \in \mathbb{R}^{3N}: \nonumber \\ 
    &v^+ = v_f + G f \Delta t, \label{eq:mcp_1} \\
    & \text{for} \ i = 1, \cdots, N: \nonumber \\
    &\quad 0 \leq (d_i/\Delta t + v^+_{i, n}) \perp f_{i,n} \geq 0, \label{eq:mcp_2} \\
    &\quad 0 \leq \left(\mu_i f_{i,n} -  \|f_{i,t} \| \right) \perp \|v^+_{i,t}\| \geq 0, \label{eq:mcp_3} \\
    &\quad \|v^+_{i,t}\| \ \|f_{i,t} \| = - f_{i,t}^\trsp v^{+}_{i,t}. \label{eq:mcp_4}
\end{align}
\end{mysubequations}
\end{equationbox}
This system determines the discrete contact velocities $v^+$ and contact forces $f$ coupled by the dynamics \mbox{$v^+ = v_f + G f \Delta t$}. Here, \mbox{$v^+ = J(q)\dot{q}^+$} \eqref{eq:vplus} is obtained from the generalized velocity at the next time instant, with the contact Jacobian evaluated at the current configuration $q$. Constraints \crefrange{eq:mcp_2}{eq:mcp_4} enforce the two physical requirements of the contact model: preventing penetration and dissipating energy.
As illustrated in \cref{fig:mcp}, \eqref{eq:mcp_2} states two things: 
i)~the normal component of the contact velocity has to be such that there is no penetration after time-stepping the simulation with $\Delta t$, and 
ii)~the normal contact force can only be nonzero if the contact point ends up exactly on the surface end-of-step. For the case of \eqref{eq:mcp_3} and \eqref{eq:mcp_4}, the term $\|v^+_{t}\|$ acts as a switch. When \mbox{$\|v^+_{t}\| = 0$} (i.e., sticking contact), \eqref{eq:mcp_4} is trivially satisfied and only the Coulomb law \mbox{$\|f_{i,t} \| \leq \mu_i f_{i,n}$} remains. When \mbox{$\|v^+_{t}\| \neq 0$} (i.e., sliding contact), \eqref{eq:mcp_3} and \eqref{eq:mcp_4} achieve a \say{maximal dissipation} effect: \eqref{eq:mcp_4} states that $f_{t}$ should look at the direct opposite direction of $v^+_{t}$, and \eqref{eq:mcp_3} states that \mbox{$\mu f_{n} = \|f_{t}\|$} (i.e., the tangential contact force should saturate its Coulomb upper-bound).

\smallbreak
\textbf{Constraint Stabilization.}
When \mbox{$d < 0$}, requiring the penetration to be resolved in a single timestep through \mbox{$d/\Delta t + v^+_n \geq 0$} can introduce large corrective velocities and cause jitter. Replacing $d/\Delta t$ with a more gradual correction $\gamma_n$ is a form of \emph{constraint stabilization}. This approach follows the longstanding principle of feeding constraint errors back into the dynamics, classically associated with Baumgarte stabilization~\cite{baumgarte1972stabilization}. In velocity-based solvers, a common implementation introduces a bias proportional to the positional error divided by the timestep. ODE~\cite{smith2006ode} and Bullet~\cite{coumans2015bullet} control this correction through an \emph{error reduction parameter} (ERP), which specifies the fraction of constraint error targeted for removal at each step.
For an existing penetration, this gives \mbox{$\gamma_n = \alpha d/\Delta t$}, where \mbox{$0 \leq \alpha \leq 1$} plays the role of ERP. When the normal constraint is active and solved exactly, the linearized gap satisfies \mbox{$d + \Delta t\,v_n^+ = (1-\alpha)d$}; thus, \mbox{$\alpha = 0.5$} targets a halving of the penetration per step. This geometric decay describes the linearized correction, rather than guaranteeing the same reduction for changing contact geometry or an inexact solve. Kamino~\cite{tsounis2026kamino} uses a correction of this type, written in our notation as \mbox{$\gamma_n = \alpha \min(0, d/\Delta t) + \max(0, d/\Delta t)$}, which relaxes penetration recovery while retaining the original non-penetration bound for separated surfaces.
The MuJoCo simulator~\cite{mujoco_docs} in turn uses a spring-damper term \mbox{$\gamma_n = b v_{n} + k d w(d)$} 
(with damping coefficient $b$, stiffness $k$, and a non-linear scaling term $w(d_i)$).\footnote{To be precise, Mujoco uses the nonlinear least-squares transcription of the problem (as detailed later in \cref{sec:formulation}), and sets \mbox{$a_{\text{ref}} = -b v - k d w(d) \hat{n} \in \mathbb{R}^3$}. Moving this expression into the dual-form using the relation \mbox{$a_{\text{ref}} = J\ddot{q}_f - {r}/{\Delta t}$} with \mbox{$r = v_{f} + \gamma$} in turn gives \mbox{$\gamma = \left(b v + k d w(d) \hat{n}\right)\Delta t - v$}.
} Generally, the construction of the term $\gamma_n$ is a matter of engineering design with pragmatic motivations and empirical justifications, and all the mathematical derivations in the following chapters are agnostic to the exact functional form of $\gamma_n$, as long as \mbox{$\gamma_n = 0$} holds whenever \mbox{$d=0$}. Hence, the rest of the tutorial proceeds assuming a properly constructed function $\gamma_n$ that implements penetration resolution is given, replacing the $d/\Delta t$ term in \eqref{eq:mcp_2} with $\gamma_n$.

\subsubsection{Nonlinear Cone Complementarity (N-CCP) Transcription}
\begin{figure}
    \centering
    \usetikzlibrary{arrows.meta}
\begin{tikzpicture}[
    font=\footnotesize,
    >={Stealth[length=3.5pt]},
    Kbd/.style={draw=primalconecolor, line width=0.5pt},
    Kdbd/.style={draw=dualconecolor, line width=0.5pt},
    Kpbd/.style={draw=polarconecolor, line width=0.5pt},
    vec/.style={-{Stealth[length=5pt, width=4pt]}, line width=1.1pt, draw=geometrycolor},
    axis/.style={->, line width=0.5pt, draw=geometrycolor},
    rmark/.style={line width=0.4pt, draw=geometrycolor},
    leader/.style={line width=0.4pt, draw=annotationcolor},
    apex/.style={fill=white, draw=geometrycolor, line width=0.4pt},
    nccp cones/.pic={
      \shade[top color=dualconecolor!10, bottom color=dualconecolor!30]
        (0,0) -- (30:1.44) -- (150:1.44) -- cycle;
      \shade[top color=polarconecolor!50, bottom color=polarconecolor!25]
        (0,0) -- (-30:1.44) -- (-150:1.44) -- cycle;
      \shade[top color=primalconecolor!20, bottom color=primalconecolor!50, fill opacity=0.85]
        (0,0) -- (60:2.02) -- (120:2.02) -- cycle;
      \draw[Kdbd] (150:1.44) -- (0,0) -- (30:1.44);
      \draw[Kpbd] (-150:1.44) -- (0,0) -- (-30:1.44);
      \draw[Kbd] (120:2.02) -- (0,0) -- (60:2.02);
      \node[anchor=south west, inner sep=1pt, text=primalconecolor!75!black]
        at ([shift={(0.02,-0.25)}]60:2.02) {$\mathcal{K}$};
      \draw[axis] (0,0) -- (0,0.38) node[above, inner sep=1pt] {$\hat{n}$};
      \draw[axis] (0,0) -- (0.38,0) node[right, inner sep=1pt] {$\hat{t}$};
    },
  ]
  \begin{scope}
    \pic {nccp cones};
    \draw[leader] (-150:1.15) -- ++(140:0.2)
      node[above, inner sep=1pt, text=polarconecolor!75!black] {$\boundary\mathcal{K}^\circ$};
    \node at (-0.4,-0.47) {$\mathcal{K}^\circ$};
    \draw[rmark] (60:0.14) -- ++(-30:0.14) -- ++(240:0.14);
    \draw[vec] (0,0) -- (60:1.58) node[pos=0.88, left, inner sep=2pt] {$x$};
    \draw[vec] (0,0) -- (-30:1.1) node[above right, inner sep=1pt] {$y$};
    \filldraw[apex] (0,0) circle (0.035);
    \node[anchor=base] at (0,-1.15) {$\mathcal{K} \ni x \perp y \in \mathcal{K}^\circ$};
  \end{scope}

  \begin{scope}[shift={(3.35,0)}]
    \pic {nccp cones};
    \node[anchor=south west, inner sep=1pt, text=dualconecolor!75!black]
      at ([shift={(-0.1,0.02)}]30:1.44) {$\mathcal{K}^*$};
    \draw[leader] (30:1.0) -- ++(-40:0.3)
      node[below, inner sep=1pt, text=dualconecolor!75!black] {$\boundary\mathcal{K}^*$};
    \draw[rmark] (60:0.14) -- ++(150:0.14) -- ++(240:0.14);
    \draw[vec] (0,0) -- (60:1.58) node[pos=0.88, left, inner sep=2pt] {$x$};
    \draw[vec] (0,0) -- (150:1.1) node[below left, inner sep=1pt] {$y$};
    \filldraw[apex] (0,0) circle (0.035);
    \node[anchor=base] at (0,-1.15) {$\mathcal{K} \ni x \perp y \in \mathcal{K}^*$};
  \end{scope}

  \node[anchor=base] at (1.675,-1.58) {(the case where $x, y \neq 0$)};
\end{tikzpicture}
    \caption{\captitle{Friction direction from cone complementarity.} Orthogonal vectors in a circular cone and its dual have opposing collinear tangential components. For contact forces and corrected velocities, this gives the friction direction required by maximum dissipation; saturation of the Coulomb bound during sliding completes the friction law.} \label{fig:nccp}
    \Description{Two 2D sketches of a circular cone with its dual and polar cones around the normal direction. In each, a vector x on the boundary of the cone is perpendicular to a vector y on the boundary of the polar cone (left) or of the dual cone (right), and the tangential components of x and y point in opposite directions.}
\end{figure}
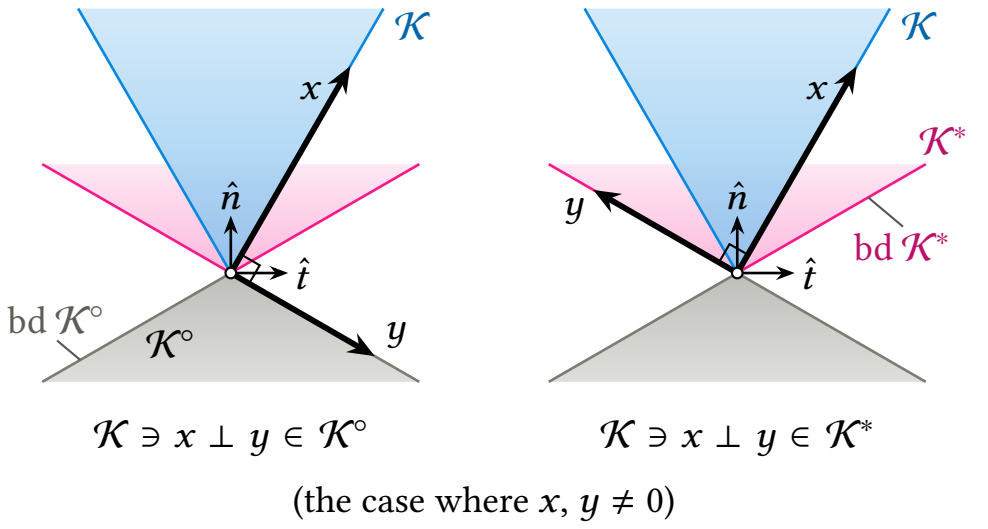

\begin{figure}
    \centering
    \usetikzlibrary{arrows.meta,calc}
\begin{tikzpicture}[
    scale=1.3, font=\footnotesize,
    >={Stealth[length=3.5pt]},
    Kbd/.style={draw=primalconecolor, line width=0.5pt},
    Kdbd/.style={draw=dualconecolor, line width=0.5pt},
    plane/.style={draw=fixedbodycolor, line width=0.8pt},
    vec/.style={->, line width=0.9pt},
    axis/.style={->, line width=0.6pt, draw=geometrycolor},
    helper/.style={line width=0.5pt, draw=annotationcolor, densely dashed},
    rmark/.style={line width=0.4pt, draw=geometrycolor},
    dimline/.style={line width=0.4pt, draw=annotationcolor},
    leader/.style={line width=0.4pt, draw=annotationcolor},
    apex/.style={fill=white, draw=geometrycolor, line width=0.4pt},
  ]
  \def\vtlen{1.6}
  \coordinate (O) at (0,0);
  \coordinate (KL) at (120:2.54);                %
  \coordinate (KR) at (60:2.54);
  \coordinate (DL) at (150:2.04);                %
  \coordinate (DR) at (30:2.04);
  \coordinate (F) at (60:2.34);                  %
  \coordinate (Ft) at (F |- O);                  %
  \coordinate (Vt) at (-\vtlen,0);               %
  \coordinate (P) at ($(Vt)+(0,{\vtlen*tan(30)})$); %
  \coordinate (Q) at ($(P)+(0,0.82)$);           %

  \shade[top color=dualconecolor!10, bottom color=dualconecolor!30] (O) -- (DR) -- (DL) -- cycle;
  \shade[top color=primalconecolor!25, bottom color=primalconecolor!45, fill opacity=0.85] (O) -- (KR) -- (KL) -- cycle;
  \draw[Kdbd] (DL) -- (O) -- (DR);
  \draw[Kbd] (KL) -- (O) -- (KR);
  \node[anchor=west, inner sep=1pt, text=primalconecolor!75!black] at ($(KR)+(0.08,-0.12)$) {$\mathcal{K}$};
  \node[anchor=south, inner sep=1pt, text=dualconecolor!75!black] at ($(DR)+(-0.12,0.03)$) {$\mathcal{K}^*$};
  \draw[leader] ($(O)+(30:1.68)$) -- ++(-40:0.3) node[anchor=north west, inner sep=1pt, text=dualconecolor!75!black] {$\boundary\mathcal{K}^*$};

  \draw[plane] (-1.72,0) -- (1.85,0);

  \draw[helper] (F) -- (Ft);
  \draw[rmark] ($(Ft)+(-0.17,0)$) -- ++(0,0.17) -- ++(0.17,0);
  \draw[vec, draw=forcecolor] (O) -- (Ft) node[below, inner sep=2pt, text=forcecolor] {$f_t$};
  \draw[vec, draw=forcecolor] (O) -- (F) node[pos=0.86, left, inner sep=2pt, text=forcecolor] {$f$};

  \coordinate (A) at (150:1.21);
  \draw[dimline] (A) -- ++(0,-0.231) coordinate (C) -- ++(0.4,0);
  \node[left, inner sep=1pt] at ($(A)!0.5!(C)$) {$\mu$};
  \node[below, inner sep=1.5pt] at ($(C)+(0.2,0)$) {$1$};

  \draw[rmark] ($(Vt)+(0.17,0)$) -- ++(0,0.17) -- ++(-0.17,0);
  \draw[vec, draw=velocitycolor] (O) -- (Vt) node[pos=0.72, below, inner sep=3pt, text=velocitycolor] {$v_t^+$};
  \draw[vec, draw=correctioncolor] (Vt) -- (P) node[midway, left, inner sep=2pt, text=correctioncolor!75!black] {$\mu\|v_t^+\|\hat{n}$};
  \draw[helper] ($(Q)+(-0.14,0)$) -- ++(0.42,0);
  \draw[helper] (P) -- ++(0.3,0);
  \draw[vec, draw=velocitycolor] (P) -- (Q) node[midway, left, inner sep=2pt, text=velocitycolor] {$v_n^+$};
  \draw[vec, draw=stabilizationcolor] ($(Q)+(0.165,0)$) -- ($(P)+(0.165,0)$) node[midway, right, inner sep=2pt, text=stabilizationcolor] {$\gamma_n\hat{n}$};
  \fill[geometrycolor] (P) circle (0.035);

  \draw[rmark] ($(O)+(60:0.2)$) -- ++(150:0.2) -- ++(240:0.2);
  \draw[->, line width=0.4pt, draw=geometrycolor] (0.4,0) arc (0:-180:0.4);
  \node[inner sep=0pt] at (0,-0.22) {$180^\circ$};

  \draw[axis] (O) -- (0,0.5) node[above, inner sep=1pt] {$\hat{n}$};
  \draw[axis] (O) -- (0.5,0) node[below right, inner sep=1pt] {$\hat{t}$};
  \filldraw[apex] (O) circle (0.035);

  \coordinate (mid) at (current bounding box.center);
  \node[anchor=base] at (mid |- {(0,-0.68)}) {$\mathcal{K} \ni f \perp \bigl(v^+ + (\mu\|v_t^+\| + \gamma_n)\hat{n}\bigr) \in \mathcal{K}^*$};
  \node[anchor=base] at (mid |- {(0,-0.97)}) {(the case where $v_t^+ \neq 0$)};
\end{tikzpicture}
    \caption{\captitle{De Saxcé correction.} Including the De-Saxcé correction $\mu \|v^+_t\| \hat{n}$ allows purely tangential contact velocities $v^+ = v^+_t = -\|v^+_t\| \hat{t} \notin \mathcal{K}^*$ to be valid solutions, as $v_t^+ + \mu \|v^+_t\| \hat{n} \in \mathcal{K}^*$ holds.}
    \vspace{-5mm}
    \label{fig:nccp_correction}
    \Description{A 2D sketch of the friction cone and its dual cone around the normal. A purely tangential end-of-step velocity along the contact plane lies outside the dual cone. Adding the de Saxc\'e correction, a normal vector of length mu times the tangential speed, together with the stabilization term moves the sum onto the boundary of the dual cone, perpendicular to the contact force on the boundary of the friction cone.}
\end{figure}
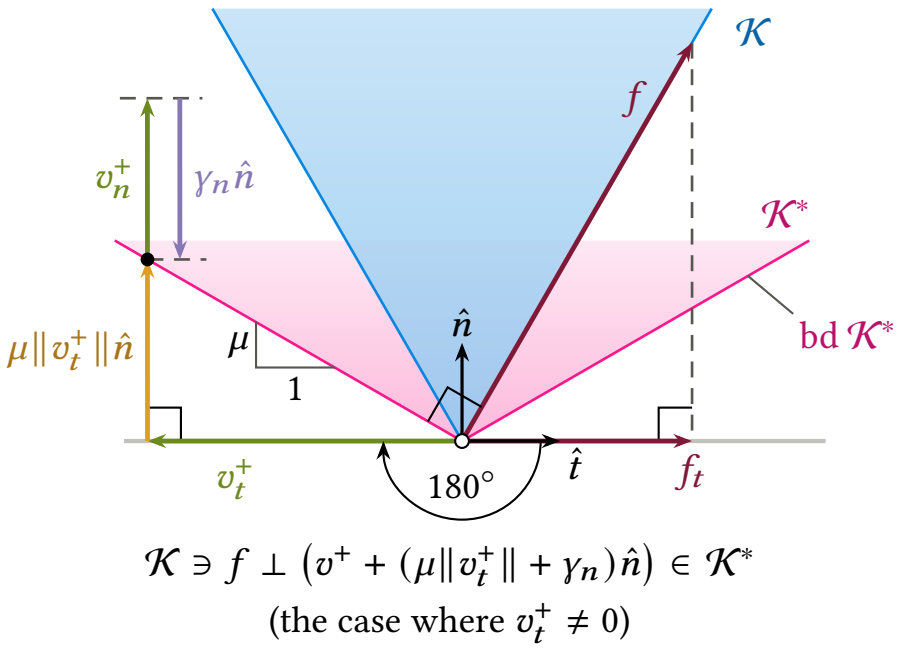
The N-CCP transcription is an exact reformulation of the Signorini–
Coulomb transcription that unifies the scalar nonlinear complementarity conditions \crefrange{eq:mcp_2}{eq:mcp_3} and the friction-direction condition \eqref{eq:mcp_4} into a single nonlinear cone complementarity condition~\cite{DeSaxceFeng1991, DeSaxceFeng1998}. This reformulation yields what is commonly called a \say{variational inequality}~\cite{Fichera1963}, enabling the use of numerical methods developed for this general class of problems, such as the projected Gauss-Seidel method. The reformulation is achieved by replacing the conditions \crefrange{eq:mcp_2}{eq:mcp_4} with the following equivalent set (written for a single contact point):
\begin{mysubequations}
\begin{align}
    \text{Find} \:\: & v^+, f \in \mathbb{R}^3: \nonumber \\ 
    & f \in \mathcal{K}, \label{eq:nccp1} \\
    & v^+ + (\mu\left\|v^+_{t}\right\| + \gamma_n) \hat{n} \in \mathcal{K}^*, \label{eq:nccp2} \\
    & f^\trsp \left[v^+ + (\mu\left\|v^+_{t}\right\| + \gamma_n) \hat{n} \right]= 0, \label{eq:nccp3}
\end{align}
\end{mysubequations}%
where $\hat{n} \in \mathbb{R}^3$ is the contact normal and $\mathcal K$ is the friction cone. \Crefrange{eq:nccp1}{eq:nccp3} are often notated as a single cone complementarity condition $\mathcal{K} \ni f \perp (v^+ + \Gamma(v^+) + \gamma) \in \mathcal{K}^*$, where the nonlinear term $\Gamma(v^+) = \mu\left\|v^+_{t}\right\| \hat{n} \in \mathbb{R}^3$ is called the \say{de Saxcé correction}, and the term $\gamma = \gamma_n \hat{n} \in \mathbb{R}^3$ is simply the constraint stabilization lifted to 3D~\cite{lelidec2023contacts, carpentier2024compliant}.
The specific role of the de Saxc\'e correction is to recover the Signorini normal-contact condition within this cone formulation~\cite{DeSaxceFeng1998, lelidec2023contacts}. Maximum dissipation in the tangential directions follows from the friction-force direction and saturation of the Coulomb bound during sliding, as detailed below.
To understand the equivalence with the Signorini--Coulomb transcription, consider the following:
\begin{itemize}[leftmargin=*, label={\color{bulletcolor}\textbf{\textbullet}}]
    \item Since $\Gamma(v^+)$ and $\gamma$ act only in the normal direction, the corrected velocity has tangential component $v_t^+$. Cone membership and orthogonality therefore enforce \mbox{$f_t^\trsp v_t^+ = -\|f_t\|\,\|v_t^+\|$}, as illustrated in \cref{fig:nccp} and expressed by \eqref{eq:mcp_4}. Together with \eqref{eq:mcp_3}, which imposes \mbox{$\|f_t\| = \mu f_n$} during sliding, this gives maximum tangential dissipation for a given normal force $f_n$.
    \item The dual cone is characterized by \mbox{$y_n \geq \mu\|y_t\|$}. Consequently, with the de Saxc\'e correction, \eqref{eq:nccp2} reduces to \mbox{$v_n^+ + \gamma_n \geq 0$}. Combined with orthogonality, it yields the full stabilized Signorini condition \mbox{$0 \leq v_n^+ + \gamma_n \perp f_n \geq 0$}. Without the correction, dual-cone membership would instead require \mbox{$v_n^+ + \gamma_n \geq \mu\|v_t^+\|$}, coupling tangential slip to normal separation. In particular, for $\gamma_n=0$, $f_n>0$, and $\mu>0$, uncorrected cone complementarity would impose \mbox{$v_n^+ = \mu\|v_t^+\|$} during sliding, an artificial separation known as \emph{dilatancy}~\cite{lelidec2023contacts}. The correction removes this coupling and permits sliding with zero normal velocity (\cref{fig:nccp_correction}).
    \item Condition \eqref{eq:nccp1} ensures that $f_{n} \geq 0$ and $\|f_{t} \| \leq \mu f_{n}$ both hold.
    \item Condition \eqref{eq:nccp3} ensures that \eqref{eq:mcp_2} and \eqref{eq:mcp_3} both hold, since: 
    \begin{align*}
        &f^\trsp \left[v^+ + (\mu\left\|v^+_{t}\right\| + \gamma_n) \hat{n}\right] \\
        &= f_{t}^\trsp \ v_{t}^+ + f_{n}(\mu \|v_{t}^+\| + v_{n}^+ + \gamma_n) \\
        &= - \|v_{t}^+\| \, \|f_{t} \| + \|v_{t}^+\| \, \mu f_{n} + f_{n}(v_{n}^+ + \gamma_n) \\
        &= \|v_{t}^+\|\left(\mu f_{n} -  \|f_{t} \| \right) + f_{n}(v_{n}^+ + \gamma_n),
    \end{align*}
    and we have already established that the relations $\|v_{t}^+\| \geq 0$, $(\mu f_{n} -  \|f_{t} \|) \geq 0$, $f_{n} \geq 0$ and $(v_{n}^+ + \gamma_n) \geq 0$ all hold, meaning the relation $f^\trsp \left[v^+ + (\mu\left\|v^+_{t}\right\| + \gamma_n) \hat{n}\right] = 0$ implies that its two constituents $\|v_{t}^+\|\left(\mu f_{n} -  \|f_{t} \| \right)$ and $f_{n}(v_{n}^+ + \gamma_n)$ are both zero.
\end{itemize}
Putting together conditions \crefrange{eq:nccp1}{eq:nccp3} and the Newton-Euler equation $v^+ = v_f + G f \Delta t$ results in the N-CCP transcription:
\begin{align}
    \text{Find} \:\: & v^+, f \in \mathbb{R}^{3N}: \nonumber \\ 
    &v^+ = v_f + G f \Delta t, \label{eq:nccp_long_1} \\
    & \text{for} \ i = 1, \cdots, N: \nonumber \\
    & \quad \mathcal{K}_i \ni f_i \perp (v^+_i + \Gamma(v^+_i) + \gamma_i) \in \mathcal{K}_i ^*. \label{eq:nccp_long_2}
\end{align}
Defining the additional notation $\Gamma(v^+) = \text{concat}(\left\{\Gamma(v^+_i)\right\}) \in \mathbb{R}^{3N}$, $\gamma = \text{concat}(\left\{\gamma_i\right\}) \in \mathbb{R}^{3N}$, and $\mathcal{K} = \prod_i \mathcal{K}_i$ (i.e., the cartesian set product of $\mathcal{K}_i$), we can shorthand the final transcription as:
\begin{equationbox}{N-CCP Transcription}
\begin{mysubequations} \label{eq:nccp}
\begin{align}
    \text{Find} \:\: & v^+, f \in \mathbb{R}^{3N}: \nonumber \\ 
    &v^+ = v_f + G f \Delta t, \label{eq:nccp_1} \\
    &\mathcal{K} \ni f \perp (v^+ + \Gamma(v^+) + \gamma) \in \mathcal{K} ^*. \label{eq:nccp_2}
\end{align}
\end{mysubequations}
\end{equationbox}

\subsubsection{Root-Finding Transcription}
\label{sec:root_finding}

\Cref{sec:cone-complementarity} introduced complementarity functions as a way to express cone membership and orthogonality through nonlinear equations. We now apply this construction to the N-CCP transcription. For each contact, the second-order-cone Fischer--Burmeister (SOCFB) function vanishes exactly when the contact force and the corrected velocity satisfy cone complementarity, as stated in \eqref{eq:socfb}. Replacing the complementarity condition in \eqref{eq:nccp_2} by this zero-residual condition gives the following equivalent root-finding problem:
\begin{equationbox}{Root-Finding Transcription}
\begin{mysubequations} \label{eq:root_finding}
\begin{align}
    \text{Find} \:\: & v^+, f \in \mathbb{R}^{3N}: \nonumber \\ 
    &v^+ = v_f + G f \Delta t, \label{eq:nse_1} \\
    &\psi_{\mathcal{K}}(h(f \,, \, v^+ + \Gamma(v^+) + \gamma)) = 0, \label{eq:nse_2}
\end{align}
\end{mysubequations}
\end{equationbox}
where $\psi_{\mathcal{K}}$ stacks one three-dimensional SOCFB residual per contact, each using the rescaling function $h$ \eqref{eq:lorentz_map} associated with its friction coefficient $\mu_i$. A zero residual thus enforces the complete Signorini--Coulomb law at every contact, and the dynamics \cref{eq:nse_2} couple these local conditions through $G$.

Eliminating $v^+$ with \eqref{eq:nse_1} leaves $3N$ nonlinear equations in the $3N$ contact-force components:
\begin{equation}
    \mathcal{R}(f) := \psi_{\mathcal{K}}\left(f,\; v_f + Gf\Delta t + \Gamma\left(v_f + Gf\Delta t\right) + \gamma\right) = 0.
\end{equation}
This form allows Newton-type methods to update the forces by linearizing the residual, as described in \cref{alg:nse-newton}. The reformulation preserves the nonsmoothness of contact transitions, so generalized derivatives or a smoothed complementarity function are used where ordinary derivatives are unavailable.

\subsubsection{Cascaded Quadratically Constrained Quadratic Program (C-QCQP) Transcription}

The C-QCQP transcription is an exact reformulation of the N-CCP transcription as a cascade of quadratically constrained quadratic programs (QCQP). The main benefit of this reformulation lies in the way that it establishes a direct connection between contact simulation and convex optimization~\cite{acary2011formulation, acary2018solving}, enabling the transfer of solution methods from the latter literature.
To obtain this transcription, let us first trivially expand the N-CCP transcription via a redundant variable $\sigma$ as follows:
\begin{mysubequations} \label{eq:c_qcqp_precursor}
\begin{align}
    \text{Find} \:\: & v^+, \sigma, f \in \mathbb{R}^{3N}: \nonumber \\ 
    &v^+ = \sigma, \\
    &v^+ = v_f + G f \Delta t, \label{eq:c_qcqp_pre_1} \\
    &\mathcal{K} \ni f \perp (v^+ + \Gamma(\sigma) + \gamma) \in \mathcal{K} ^*. \label{eq:c_qcqp_pre_2}
\end{align}
\end{mysubequations}%
The main reason for introducing the term $\sigma$ is to decouple the de Saxcé term $\Gamma(\sigma)$ from the end-of-step velocity $v^+$. For a fixed $\sigma$, substituting \eqref{eq:c_qcqp_pre_1} into \eqref{eq:c_qcqp_pre_2} gives the complementarity relation $\mathcal{K} \ni f \perp (G \Delta t) f + (v_f  + \Gamma(\sigma) + \gamma) \in \mathcal{K} ^*$. A direct comparison to \eqref{eq:cone_kkt_gI} with $\nabla f(f) = (G \Delta t) f + (v_f  + \Gamma(\sigma) + \gamma)$ identifies this complementarity relation as the optimality condition for a QCQP of the form $\min_{f\in \mathcal{K}} \: \frac{1}{2}f^\trsp (G \Delta t)f + (v_f + \Gamma(\sigma) + \gamma)^\trsp f$~\cite{acary2018solving, carpentier2024compliant}. Therefore,  \eqref{eq:c_qcqp_precursor} can be equivalently stated as:
\begin{equationbox}{C-QCQP Transcription}
\begin{mysubequations} \label{eq:c_qcqp}
\begin{align}
    \text{Find} \:\: & \sigma, f^* \in \mathbb{R}^{3N}: \nonumber \\ 
    &\sigma = v_f + G f^* \Delta t, \label{eq:c_qcqp_1} \\
    &f^* = \argmin_{f\in \mathcal{K}} \: \frac{1}{2}f^\trsp (G \Delta t)f + (v_f + \Gamma(\sigma) + \gamma)^\trsp f. \label{eq:c_qcqp_2}
\end{align}
\end{mysubequations}
\end{equationbox}

\subsubsection{Coupled Quadratic Program (QP) Transcription} 

The C-QP transcription is an exact reformulation of the Signorini--Coulomb transcription as two box-constrained QPs, one in the normal and one in the tangential force magnitudes, which are coupled through their linear terms and through the Coulomb bound $f_t \leq \mu f_n$. Just as the C-QCQP transcription connects the N-CCP transcription to convex optimization, C-QP transcription establishes a similar connection for the Signorini--Coulomb transcription. This time the cone involved is the set of non-negative real numbers $\mathbb{R}_+$, which is self-dual (i.e.,\ \mbox{$\mathbb{R}_+^* = \mathbb{R}_+$}).
To obtain the C-QP transcription, let us write the tangential force of each contact as $f_{i,t} = \norm{f_{i,t}}\hat{t}_i$ along a unit tangent $\hat{t}_i$. By the maximal dissipation condition \eqref{eq:mcp_4}, $\hat{t}_i$ can be chosen such that the tangential velocity points the opposite way, $v^+_{i,t} = -\norm{v^+_{i,t}}\hat{t}_i$. Since $\hat{t}_i$ is a unit vector orthogonal to $\hat{n}_i$, this choice is captured by the direction conditions:
\begin{equation}
    \hat{n}_i^\trsp \hat{t}_i = 0, \qquad \norm{\hat{t}_i} = 1, \qquad \hat{t}_i^\trsp v^+_i + \norm{v^+_{i,t}} = 0. \label{eq:cqp_dir}
\end{equation}
To write all contacts at once, let us stack the normals and tangents into block-diagonal matrices, $\bar{n} = \operatorname{diag}\left(\hat{n}_i\right)$ and $\bar{t} = \operatorname{diag}\left(\hat{t}_i\right)$, both in $\mathbb{R}^{3N \times N}$, so that $f = \bar{n} f_n + \bar{t}\norm{f_t}$, where, with a slight abuse of notation, $f_n, \norm{f_t}, v^+_n, \norm{v^+_t} \in \mathbb{R}^N$ denote the stacked vectors of the per-contact normal components and tangential norms. Projecting the dynamics equation \eqref{eq:mcp_1} onto $\bar{n}$ and $\bar{t}$, and using the direction condition \eqref{eq:cqp_dir} to replace $\bar{t}^\trsp v^+$ with $-\norm{v^+_t}$, gives
\begin{align}
    v_n^+ &= \left(\bar{n}^\trsp G \bar{n}\Delta t\right)f_n + \bar{n}^\trsp\big(v_f + (G\Delta t) \bar{t} \norm{f_t}\big), \label{eq:decoupled_n} \\
    -\norm{v_t^+} &= \left(\bar{t}^\trsp G \bar{t}\Delta t\right)\norm{f_t} + \bar{t}^\trsp\big(v_f + (G\Delta t) \bar{n} f_n\big). \label{eq:decoupled_t}
\end{align}
Adding $\gamma_n$ to both sides of \cref{eq:decoupled_n}, adding $\bar{t}^\trsp\gamma = 0$ to the right side of \cref{eq:decoupled_t}, and combining them with the remaining conditions \crefrange{eq:mcp_2}{eq:mcp_3} gives:
\begin{align}
    &v_n^+ + \gamma_n = \left(\bar{n}^\trsp G \bar{n}\Delta t\right)f_n + \bar{n}^\trsp\big(v_f + \gamma + (G\Delta t) \bar{t} \norm{f_t}\big), \label{eq:dqp_n_1} \\
    &0 \leq (v^+_{n} + \gamma_n) \perp f_{n} \geq 0, \label{eq:dqp_n_2} \\
    \nonumber\\
    &-\norm{v_t^+} = \left(\bar{t}^\trsp G \bar{t}\Delta t\right)\norm{f_t} + \bar{t}^\trsp\big(v_f + \gamma + (G\Delta t) \bar{n} f_n\big), \label{eq:dqp_t_1} \\
    &0 \leq \left(\mu f_{n} - \norm{f_{t}} \right) \perp \norm{v^+_{t}} \geq 0, \label{eq:dqp_t_2}
\end{align}
where $\mu = \operatorname{diag}(\mu_1, \dots, \mu_N)$, and the complementarity conditions hold for each contact. A direct comparison to \eqref{eq:cone_kkt_gI} identifies \crefrange{eq:dqp_n_1}{eq:dqp_n_2} as the optimality conditions of a QP with the decision variable $f_n$, the primal constraint $f_n \in \mathbb{R}_+$, and the dual variable $s_n = v_n^+ + \gamma_n \in \mathbb{R}^*_+$. Similarly, a comparison to \crefrange{eq:cone_kkt_1}{eq:cone_kkt_4} with the constraint $g(\norm{f_t}) = \mu f_n - \norm{f_t} \in \mathbb{R}_+$ (with the Jacobian $J_g = -I$) identifies \crefrange{eq:dqp_t_1}{eq:dqp_t_2} as the optimality conditions of a QP with the decision variable $\norm{f_t}$ and the dual variable $s_t = \norm{v_t^+} \in \mathbb{R}^*_+$. Therefore, together with the direction condition \eqref{eq:cqp_dir} and the dynamics, \crefrange{eq:dqp_n_1}{eq:dqp_t_2} can be written equivalently as two coupled box-constrained QPs:
\begin{equationbox}{C-QP Transcription}
\begin{mysubequations} \label{eq:cqp}
\begin{align}
    \text{Find} \:\: & v^+ \in \mathbb{R}^{3N},\ \bar{t} \in \mathbb{R}^{3N \times N},\ f_n^*, \norm{f_t}^* \in \mathbb{R}^{N}: \nonumber \\
    & v^+ = v_f + (G\Delta t)\big(\bar{n} f_n^* + \bar{t} \norm{f_t}^*\big), \label{eq:cqp_1} \\
    & \text{for} \ i = 1, \cdots, N: \nonumber \\
    & \quad \hat{n}_i^\trsp\hat{t}_i = 0, \quad \norm{\hat{t}_i} = 1, \quad \hat{t}_i^\trsp v^+_i + \norm{v^+_{i,t}} = 0, \label{eq:cqp_2} \\
    &f_n^* = \argmin_{f_n \geq 0} \: \tfrac{1}{2}f_n^\trsp \left(\bar{n}^\trsp G \bar{n}\Delta t\right)f_n \label{eq:cqp_3} \\
    & \quad\quad\quad\quad\quad\:\:  + \big(v_f + \gamma + (G\Delta t)\, \bar{t} \norm{f_t}^*\big)^\trsp \bar{n} f_n, \nonumber \\
    &\norm{f_t}^* = \argmin_{0 \leq \norm{f_t} \leq \mu f_{n}^*} \: \tfrac{1}{2}\norm{f_t}^\trsp \left(\bar{t}^\trsp G \bar{t}\Delta t\right)\norm{f_t} \label{eq:cqp_4} \\
    & \quad\quad\quad\quad\quad\quad\:\: + \big(v_f + \gamma + (G\Delta t)\, \bar{n} f_n^*\big)^\trsp \bar{t} \norm{f_t}. \nonumber
\end{align}
\end{mysubequations}
\end{equationbox}
This transcription is particularly insightful because it isolates what makes the contact dynamics problem hard mechanistically: finding the tangential directions $\bar{t}$ and the Coulomb bounds $\mu f_n$ that couple the two QPs. In other words, deciding which contacts slip (in which direction), and which contacts stick (with which Coulomb upperbound on the friction force it can sustain). If both were known, with the bounds frozen at their values $s = \mu f_n^*$, the two QPs \crefrange{eq:cqp_3}{eq:cqp_4} would combine and the entire contact dynamics problem would simply reduce into a single box-constrained, convex QP:
\begin{align*}
    \min_{f_n \geq 0,\ 0 \leq \norm{f_t} \leq s} \: &\tfrac{1}{2}\big(\bar{n} f_n + \bar{t}\norm{f_t}\big)^\trsp (G\Delta t) \big(\bar{n} f_n + \bar{t}\norm{f_t}\big) \\
    & + \big(v_f + \gamma \big)^\trsp \big(\bar{n} f_n + \bar{t}\norm{f_t}\big).
\end{align*}

\subsection{Convex Relaxations}
Convex relaxations simplify numerical solution by modifying the contact law to obtain a convex optimization problem. We first show how omitting the de Saxc\'e correction yields two equivalent descriptions: the L-CCP conditions and a cone-constrained QCQP in the contact forces. We then introduce regularization to obtain a unique force solution, before deriving the equivalent NLS formulation in generalized accelerations. These steps distinguish the modeling changes introduced by relaxation and regularization from the subsequent changes in mathematical representation. The next subsection examines their physical implications.

\subsubsection{Linear Cone Complementarity (L-CCP) and QCQP Transcriptions}

The equivalent exact transcriptions of complementarity-based contact dynamics all correspond to a nonconvex problem with no polynomial-time algorithm known at the velocity level~\cite{baraff1993issues}. A convex program in turn can be solved to any prescribed accuracy in polynomial time and, in practice, by the standard methods of convex optimization~\cite{boyd2004convex}. Therefore, most modern simulation frameworks ignore the De-Saxcé term $\Gamma(v^+)$ in the exact transcriptions to make the problem convex~\cite{anitescu2004constraint, anitescu2006optimization}, resulting in the linear cone complementarity (L-CCP) and QCQP transcriptions:
\begin{equationbox}{L-CCP Transcription}
\begin{align}
    \text{Find} \:\: & f \in \mathbb{R}^{3N}: \nonumber \\ 
    &\mathcal{K} \ni f \perp ((G \Delta t) f  + v_f + \gamma) \in \mathcal{K} ^*, \label{eq:lccp_unreg}
\end{align}
\end{equationbox}

\begin{equationbox}{QCQP Transcription}
\begin{align}
    \text{Find} \:\: & f^* \in \mathbb{R}^{3N}: \nonumber \\ 
    &f^* = \argmin_{f\in \mathcal{K}} \: \frac{1}{2}f^\trsp (G \Delta t) f + (v_f + \gamma)^\trsp f. \label{eq:socqp_unreg}
\end{align}
\end{equationbox}
It is typical in multi-contact settings for $G$ to be rank deficient, and hence these problems are often only positive semi-definite.
Therefore, standard relaxations add a regularization matrix $R$ to the Delassus matrix $G$ such that $G + R \succ 0$, usually via a diagonal $R \succ 0$, making the problem strictly positive-definite and strongly convex~\cite{todorov2014convex},
This regularization, combined with the shorthand $r = v_f + \gamma$ for notational brevity, results in the following transcriptions:
\begin{equationbox}{L-CCP Transcription (Regularized)}
\begin{align}
    \text{Find} \:\: & f \in \mathbb{R}^{3N}: \nonumber \\ 
    &\mathcal{K} \ni f \perp \left((G + R)f \Delta t  + r \right) \in \mathcal{K} ^*, \label{eq:lccp}
\end{align}
\end{equationbox}

\begin{equationbox}{QCQP Transcription (Regularized)}
\begin{align}
    \text{Find} \:\: & f^* \in \mathbb{R}^{3N}: \nonumber \\ 
    &f^* = \argmin_{f\in \mathcal{K}} \: \frac{1}{2}f^\trsp \left((G + R)\Delta t \right)f + r^\trsp f. \label{eq:socqp}
\end{align}
\end{equationbox}

\subsubsection{Nonlinear Least Squares (NLS) Transcription}\label{sec:primal-form}
The nonlinear least-squares (NLS) transcription is an equivalent reformulation of the regularized QCQP transcription as an unconstrained strongly convex program in the generalized accelerations. It is obtained by moving the cone-constraints into the objective via a sequence of algebraic manipulations of the optimality conditions~\cite{todorov2014convex, MuJoCo150Release}. The main benefit of this reformulation is that it is very efficient to solve. As such, it is used by some of the most well-adopted frameworks such as Mujoco~\cite{todorov2012mujoco} and Drake~\cite{drake}. It takes the following form:
\begin{equationbox}{NLS Transcription}
\begin{align}
    \text{Find} \:\: & \ddot{q}^* \in \mathbb{R}^{n_v}: \nonumber \\ 
    \min_{\ddot{q}} \:& \tfrac{1}{2}\left(\ddot{q} - \ddot{q}_f\right)^\trsp M \left(\ddot{q} - \ddot{q}_f\right) \nonumber\\ + &\tfrac{1}{2} \left[\Pi_{\mathcal{K}}\left(a_\text{ref} - J\ddot{q}\right)\right]^\trsp R^{-1}  \left[\Pi_{\mathcal{K}}\left(a_\text{ref} - J\ddot{q}\right)\right], \label{eq:nls}\\
    \nonumber\\
    (a_\text{ref} =& J\ddot{q}_f - {r}/{\Delta t}\in \mathbb{R}^{3N} \text{ with } r=v_f+\gamma_n \text{ and } v_f = J \dot q_f). \nonumber
\end{align}
\end{equationbox}
To show the equivalence between the regularized QCQP and NLS transcriptions, let us expand $\left[\Pi_{\mathcal{K}}\left(a_\text{ref} - J\ddot{q}\right)\right]^\trsp R^{-1} \left[\Pi_{\mathcal{K}}\left(a_\text{ref} - J\ddot{q}\right)\right]$ as a serial sum $\sum_{i=1}^N R^{-1}_{ii}\left\|\Pi_{\mathcal{K}_i}\left(a_\text{ref, i} - J_i \ddot{q}\right)\right\|^2$, where $a_\text{ref, i} \in \mathbb{R}^3$ is the 3D vector corresponding to the $i$th contact in $a_\text{ref} \in \mathbb R^{3N}$. Using the property $\left\|\Pi_{\mathcal{K}}(x)\right\|^2 = \min_{w \in \mathcal{K}^*} \| x + w\|^2$, we can then write:
\begin{align}
    \left\|\Pi_{\mathcal{K}_i}\left(a_\text{ref, i} - J_i \ddot{q}\right)\right\|^2 &= \min_{w \in \mathcal{K}^*_i} \left\|w + a_\text{ref, i} - J_i \ddot{q}\right\|^2 \nonumber \\
    & = \min_{w \in \mathcal{K}^*_i} \left\|a_\text{ref, i} - \left(J_i \ddot{q} - w\right)\right\|^2 \nonumber \\
    & = \min_{J_i \ddot{q} - y \in \mathcal{K}^*_i} \|a_\text{ref, i} - y\|^2, \nonumber
\end{align}
where we have made the definition $y = J_i \ddot{q} - w$. This means:
\begin{align}
    \sum_{i=1}^N R^{-1}_{ii}\left\|\Pi_{\mathcal{K}_i}\left(a_\text{ref, i} - J_i \ddot{q}\right)\right\|^2 &= \sum_{i=1}^N R_{ii}^{-1} \left(\min_{J_i \ddot{q} - y \in \mathcal{K}^*_i} \|a_\text{ref, i} - y\|^2\right) \nonumber \\
    & =  \sum_{i=1}^N \min_{J_i \ddot{q} - y \in \mathcal{K}^*_i} R_{ii}^{-1}  \|a_\text{ref, i} - y\|^2 \nonumber \\
    &= \min_{\left\{ J_i \ddot{q} - y \in \mathcal{K}^*_i\right\}_{i=1}^N} \sum_{i=1}^N R_{ii}^{-1}  \|a_\text{ref, i} - y\|^2 \nonumber \\
    &= \min_{J\ddot{q} - y \in \mathcal{K}^*} \left(y - a_\text{ref}\right)^\trsp R ^{-1} \left(y - a_\text{ref}\right), \nonumber
\end{align}
where in the last line, we defined $y = \text{concat}(\left\{y_i\right\}_{i=1}^N) \in \mathbb{R}^{3N}$. Therefore, the NLS transcription can be equivalently expressed in what is called the \say{primal-form} of the contact dynamics problem:
\begin{equationbox}{Primal-Form Transcription}
\begin{align}
    \min_{\ddot{q}, y} \:& \tfrac{1}{2}\left(\ddot{q} - \ddot{q}_f\right)^\trsp M \left(\ddot{q} - \ddot{q}_f\right) \nonumber \\ + &\tfrac{1}{2} \left(y - a_\text{ref}\right)^\trsp R^{-1}  \left(y - a_\text{ref}\right), \label{eq:primal_cons} \\
    \text{s.t.} \: & J\ddot{q} - y \in \mathcal{K}^*. \nonumber
\end{align}
\end{equationbox}
Referring to the optimality conditions \crefrange{eq:cone_kkt_1}{eq:cone_kkt_4}, and noticing that $x = \left[ \ddot{q}^\trsp \: y^\trsp\right]^\trsp \in \mathbb{R}^{n_v + 3N}$ and $g(x) =\left[ J \: -\Id\right] x$, we can write the optimality conditions for \eqref{eq:primal_cons} as:
\begin{mysubequations} \label{eq:primal_kkt}
\begin{align}
    \text{Find} \:\: &\ddot{q} \in \mathbb{R}^{n_v}, y \in \mathbb{R}^{3N}, f \in \mathbb{R}^{3N}: \nonumber \\ 
    &M \left(\ddot{q} - \ddot{q}_f\right) - J^\trsp f = 0, \label{eq:primal_kkt1} \\
    &R^{-1}  \left(y - a_\text{ref}\right) + f = 0, \label{eq:primal_kkt2} \\
    &J\ddot{q} - y \in \mathcal{K}^*,  \label{eq:primal_kkt3}\\
    &f \in \mathcal{K}, \label{eq:primal_kkt4} \\
    &f^\trsp \left(J\ddot{q} - y\right) = 0, \label{eq:primal_kkt5}
\end{align}
\end{mysubequations}%
where the dual variable is denoted as $f$ rather than $s$, and we have used $(\mathcal{K}^*)^* = \mathcal{K}$. Using \eqref{eq:primal_kkt1}, we can write \mbox{$J\ddot{q} = J\ddot{q}_f + Gf$}. Similarly, using equation \eqref{eq:primal_kkt2}, we can write $y = a_\text{ref} - R f$. Therefore, \crefrange{eq:primal_kkt1}{eq:primal_kkt2} can be replaced with a single equation $J\ddot{q} - y =  (G + R) f + \left(J\ddot{q}_f - a_\text{ref}\right)$ to obtain equivalent conditions:
\begin{align}
    \text{Find} \:\: &\ddot{q} \in \mathbb{R}^{n_v}, y \in \mathbb{R}^{3N}, f \in \mathbb{R}^{3N}: \nonumber \\ 
    & (G + R) f + \left(J\ddot{q}_f - a_\text{ref}\right) - \left(J\ddot{q} - y\right) = 0, \label{eq:primal_kkt_2_1} \\
    &J\ddot{q} - y \in \mathcal{K}^*,  \label{eq:primal_kkt_2_2}\\
    &f \in \mathcal{K}, \label{eq:primal_kkt_2_3} \\
    &f^\trsp \left(J\ddot{q} - y\right) = 0. \label{eq:primal_kkt_2_4}
\end{align}
Defining $s = J\ddot{q} - y$ and comparing \crefrange{eq:primal_kkt_2_1}{eq:primal_kkt_2_4} with the optimality conditions \crefrange{eq:cone_kkt_1}{eq:cone_kkt_4}, we see that they are the optimality conditions for the following problem:
\begin{align}
    \min_{f} \:& \tfrac{1}{2}f^\trsp(G  + R)f + \left(J\ddot{q}_f - a_\text{ref}\right)^\trsp f, \\
    \text{s.t.} \: & f \in \mathcal{K}. \nonumber
\end{align}
Finally, substituting $a_\text{ref} = J\ddot{q}_f - {r}/{\Delta t}$ in the objective term, and multiplying it with $\Delta t$ gives back the familiar regularized SOCQP:
\begin{align}
    \min_{f} \:& \tfrac{1}{2}f^\trsp \left((G + R)\Delta t\right)f + r^\trsp f, \\
    \text{s.t.} \: & f \in \mathcal{K}. \nonumber
\end{align}

\subsection{Physical Interpretations of the Convex Relaxation}

\subsubsection{NLS Optimality Conditions as Inverse Dynamics}
\label{sec:inverse_dynamics}

Let us notate the NLS transcription more compactly as:
\begin{equation}
    \min_{\ddot{q}} \: \frac{1}{2}\left(\ddot{q} - \ddot{q}_f\right)^\trsp M \left(\ddot{q} - \ddot{q}_f\right) + \frac{1}{2}\sum_{i=1}^N R^{-1}_{ii}\left\|\Pi_{\mathcal{K}_i}\left(a_\text{ref, i} - J_i \ddot{q}\right)\right\|^2. \label{eq:nls_2}
\end{equation}
Since this is an unconstrained optimization problem, its optimality condition can be expressed simply by taking the gradient with respect to $\ddot{q}$ and setting it to zero. Using the cone projection identity $\nabla \frac{1}{2}\left\| \Pi_{\mathcal{K}}(x) \right\|^2 = \Pi_{\mathcal{K}}(x)$, this results in the following equation:
\begin{align}
    M \left(\ddot{q}^* - \ddot{q}_f\right) - \sum_{i=1}^N J_i^\trsp \ R^{-1}_{ii} \ \Pi_{\mathcal{K}_i}\left(a_\text{ref, i} - J_i \ddot{q}^*\right) = 0, \label{eq:primal_optimality}
\end{align}
where $\ddot{q}^*$ is the optimal solution of \eqref{eq:nls_2}. Expanding $\ddot{q}_f = M^{-1}(\tau_u - b)$ and rearranging the terms gives:
\begin{align}
    M\ddot{q}^* + b = \tau_u + \sum_{i=1}^N J_i^\trsp \ \left[R^{-1}_{ii} \ \Pi_{\mathcal{K}_i}\left(a_\text{ref, i} - J_i \ddot{q}^*\right) \right]. \label{eq:primal_inverse_dyn}
\end{align}
Comparing this optimality condition with the Newton-Euler equation $M\ddot{q} + b = \tau_u + \sum_i J_i^\trsp f_i$ \eqref{eq:dynamics_with_forces} reveals that once the optimization problem is solved to find $\ddot{q}^*$, the rigid-body system acts exactly as if it is driven by contact forces $R^{-1}_{ii} \ \Pi_{\mathcal{K}_i}\left(a_\text{ref, i} - J_i \ddot{q}^*\right)$. This shows that the inverse dynamics under the convex relaxation of the complementarity based contact-model is governed by the equation:
\begin{align}
    \tau\left(q, \dot{q}, \ddot{q}\right) &= M(q)\ddot{q} + b(q, \dot{q}) \\ &- \sum_{i=1}^N J_i(q)^\trsp \left[R^{-1}_{ii}(q, \dot{q}) \ \Pi_{\mathcal{K}_i}\left(a_\text{ref, i}(q, \dot{q}) - J_i(q) \ddot{q}\right) \right], \label{eq:primal_inverse_dyn2}
\end{align}
and the optimization problem \eqref{eq:nls_2} can in turn be interpreted as implementing the forward-dynamics via:
\begin{align}
    \ddot{q}(q, \dot{q}, u) = \arg\min_{\ddot{q}} \left\|\tau\left(q, \dot{q}, \ddot{q}\right) - u \right\|^2.  \label{eq:forward_dyn}
\end{align}

\subsubsection{Kinetic Energy Minimization Perspective}\label{sec:kinmin}
The relationship $\dot{q}^+ = \dot{q}_f + (M^{-1}J^\trsp) f \Delta t$ expresses the end-of-step velocity $\dot{q}^+(f)$ as a linear function of the contact force $f$. Therefore, the end-of-step kinetic energy $\frac{1}{2} (\dot{q}^+)^\trsp M (\dot{q}^+)$ of the rigid-body system can be expressed (after substituting $\dot{q}^+(f)$ and performing some straight-forward algebraic manipulation) as:
\begin{align}
    \frac{1}{2}f^\trsp Gf + (v_f \ \Delta t^{-1})^\trsp f. \label{eq:kinmin}
\end{align}
Comparing this expression with the QCQP transcription \eqref{eq:socqp_unreg}, we can see that they are identical in the absence of the term $\gamma$. The main role of $\gamma$ is to resolve penetrations that happen due to discrete time-stepping with a non-zero $\Delta t$, and $\gamma \to 0$ as $\Delta t \to 0$ (since if $\Delta t$ is infinitesimal, any penetration would be caught at the exact moment it first happens, and the penetration depth $d$ would hence also be infinitesimal). Therefore, from a continuous time physics perspective, the complementarity-based formulation can be said to operate on the \emph{principle of kinetic energy minimization}.
If two surfaces in penetration move with a nonzero relative normal velocity, and forces between them are constrained to be exact opposites in the normal direction and always repulsive, then minimizing the total kinetic energy after an Euler update will create the necessary forces that will make their relative velocity zero, preventing further penetration. Similarly, if two surfaces are sliding on top of each other, and forces between them are constrained to be exact opposites in the tangential direction and always dissipative, then minimizing the total kinetic energy after an Euler update will create frictional forces unless the tangential velocities are equal.

\section{Algorithmic Solutions}
\label{sec:solution}

The formulations of \cref{sec:formulation} expose different mathematical structures in the contact problem; this section shows how numerical solvers exploit them. The central challenges are to enforce the cone constraints, handle coupling between contacts, and compute reliable updates when the underlying matrices are ill-conditioned or rank-deficient. The selected algorithms illustrate complementary ways of addressing these challenges, from inexpensive contact-wise updates to Newton steps that account for all contacts simultaneously. Throughout, we emphasize the tradeoffs between iteration cost, convergence speed, and robustness.

\smallskip
We begin with solvers for the convex relaxations, which provide building blocks for methods that solve the exact transcriptions. We then explain how these methods extend to the nonlinear contact conditions and where additional difficulties arise. The section closes with a general construction that combines a convex solver with proximal regularization and updates of the de Saxc\'e correction (\cref{sec:asocqp_sol}). Its fixed points satisfy the exact contact conditions, connecting the computational advantages of convex subproblems to the original frictional contact model.

\smallskip
The algorithms are presented as pedagogical blueprints that make their essential numerical mechanisms explicit. Implementing them efficiently also requires choices about matrix storage and assembly, linear algebra routines, line searches, and stopping criteria. These implementation details lie beyond the scope of this tutorial; the aim here is to explain why each method works, what it costs, and which difficulties it addresses.

\subsection{Convex Relaxations}
\subsubsection{L-CCP Transcription} Consider a general nonlinear cone complementarity relation of the form
\begin{equation}
    \mathcal{K} \ni x \perp F(x) \in \mathcal{K} ^*. \label{eq:pgs_1}
\end{equation}
For any $\rho > 0$, this relation is equivalent to the fixed-point relation
\begin{equation}
    x = \Pi_{\mathcal K}(x - \rho F(x)). \label{eq:pgs_2}
\end{equation}
To see this, apply the characterization \eqref{eq:proj_intuition} of the projection to the point $x - \rho F(x)$: the relation \eqref{eq:pgs_2} holds if and only if $x \in \mathcal{K}$ and $\left\langle F(x), p - x \right\rangle \geq 0$ for all $p \in \mathcal{K}$. Choosing $p = x + y$ for any $y \in \mathcal{K}$ gives $F(x) \in \mathcal K^*$, and choosing $p = 0$ and $p = 2x$ gives $\left\langle F(x), x \right\rangle = 0$. Conversely, if $x \in \mathcal{K}$, $F(x) \in \mathcal K^*$ and $\left\langle F(x), x \right\rangle = 0$, then $\left\langle F(x), p - x \right\rangle = \left\langle F(x), p \right\rangle \geq 0$ for all $p \in \mathcal{K}$, which proves the equivalence. Applying the fixed-point relation \eqref{eq:pgs_2} contact by contact to the L-CCP transcription \eqref{eq:lccp} gives the well-known projected Gauss--Seidel (PGS) algorithm:
\begin{algorithm}[htb]
\caption{Projected Gauss--Seidel (Anitescu--Tasora)}
\label{alg:lccp-pgs}
\begin{algorithmic}[1]
\State Choose $\rho$, initial $f\in\mathcal K$
\While{not converged}:
    \For{$i$ in active contacts}:
        \State $\displaystyle w_i\gets r_i
                +\Delta t\sum_{j=1}^N(G + R)_{ij}f_j$
        \State $f_i\gets\Pi_{\mathcal K_i}(f_i-\rho w_i)$
    \EndFor
\EndWhile
\State \Return $f$
\end{algorithmic}
\end{algorithm}
For the L-CCP, the algorithm goes back to Anitescu and Tasora~\cite{AnitescuTasora2010}. The main insight it introduces is that complementarity relations can be written as fixed-point relations. PGS-type solvers are available in Chrono~\cite{tasora2016chrono}, MuJoCo~\cite{todorov2012mujoco, mujoco_docs}, Simple~\cite{simple_simulator_2025}, and PhysX~\cite{nvidia2024physx}, as well as through Bullet's sequential-impulse solver~\cite{coumans2015bullet} and ODE's QuickStep solver~\cite{smith2006ode}. These implementations share the projected Gauss--Seidel principle, but differ in their contact updates and friction models; they do not all use the whole-cone projection of \cref{alg:lccp-pgs}. The main benefits of PGS are that it requires no linear solve with $G$, is easy to implement, needs minimal memory, and is robust: with suitably chosen step sizes, its iterates remain bounded and their accumulation points solve the problem even when $G$ is only positive semi-definite~\cite{AnitescuTasora2010}. Its main downside is slow convergence when the effective matrix \mbox{$(G + R)\Delta t$} is poorly conditioned. Since \mbox{$G = JM^{-1}J^\trsp$}, its spectrum depends on both the generalized inertia and the contact Jacobian. Poor conditioning can therefore arise from disparities in masses or rotational inertias, contact geometry and lever arms, or nearly dependent contact constraints. Exactly redundant constraints make $G$ singular, and weak regularization can leave the resulting system poorly conditioned even when the bodies have comparable masses.

In settings such as body stacks, where large numbers of contacts are coupled, the ordering and organization of local PGS updates based on the contact graph can also have a significant impact. For example, Erleben's velocity-based shock propagation~\cite{erleben2007velocity} processes stack layers from the supporting bodies upward, temporarily fixing the lower bodies while solving each layer with PGS. This improves the treatment of large stacks by modifying the mass coupling within each local solve, illustrating how the organization of the solver can matter alongside the choice of contact formulation.

Applying Newton's method to the (nonsmooth) fixed-point residual equation $x - \Pi_{\mathcal K}(x - \rho F(x)) = 0$ instead gives an algorithm that updates all active contacts at once rather than looping over them, whose pros and cons are the exact opposite of those of PGS:

\begin{algorithm}[htb]
\caption{Nonsmooth Newton (Alart--Curnier)}
\label{alg:lccp-newton}
\begin{algorithmic}[1]
\State Choose $\rho$, initial $f\in\mathcal K$
\While{not converged}:
    \State $w\gets r+(G + R)f\Delta t$
    \State $y\gets f-\rho w$
    \State $e\gets f-\Pi_{\mathcal K}(y)$
    \State Assemble $K=\partial\Pi_{\mathcal K}(y)$
    \State $A\gets \Id-K(\Id-\rho(G + R)\Delta t)$
    \State Solve $Ad=-e$
    \State Choose $\alpha\in(0,1]$ via line search on the projection residual
    \State $f\gets f+\alpha d$
\EndWhile
\State \Return $f$
\end{algorithmic}
\end{algorithm}
The Jacobian $\partial\Pi_{\mathcal K}$ of the cone projection is assembled contact by contact from the closed-form expressions of \cref{sec:circular_cone}; where the projection is not differentiable, any element of its generalized Jacobian can be used. A PGS sweep is usually used as a fallback when the Newton system is singular, or the line search fails.

The algorithm is a variant of the generalized Newton method of Alart and Curnier~\cite{AlartCurnier1991}, who wrote the contact law as projection equations with separate projections of the normal and tangential components. Projecting onto the whole friction cone at once, as done here, gives the formulation known as the natural map~\cite{acary2018solving}. The main insight it introduces is that setting the fixed-point residual to zero turns the fixed-point relation into a root-finding problem, which Newton-type methods can solve. As a semismooth Newton method, it converges locally superlinearly near solutions where its generalized Jacobian is nonsingular~\cite{facchinei2003finite}, so it typically needs far fewer iterations than PGS. In exchange, each iteration needs a slow linear solve, and the method is less robust.

\subsubsection{QCQP Transcription}
Because the QCQP is a convex program, any standard convex optimization method for second-order cone programs~\cite{boyd2004convex} can be used to solve it, and convergence results from the optimization literature transfer directly. The most commonly used methods are presented here. The objective, gradient, and Hessian of the regularized QCQP transcription \eqref{eq:socqp} are:
\begin{mysubequations} \label{eq:socqp_opt}
\begin{align}
    L(f) &= \tfrac12f^\trsp((G + R)\Delta t)f
        +{r}^\trsp f, \label{eq:socqp_objective} \\
    \nabla L(f) &= r+(G + R)f\Delta t, \\
    H(f) &= (G + R)\Delta t.
\end{align}
\end{mysubequations}

Perhaps the simplest method to understand and implement is projected gradient descent (PGD), which takes a gradient step on the contact forces and then projects the forces of every contact onto its friction cone. Comparing \cref{alg:socqp-pgd} with \cref{alg:lccp-pgs} shows that PGS for the L-CCP transcription is a block-coordinate (i.e. Gauss--Seidel) version of PGD.

\begin{algorithm}[htb]
\caption{Projected Gradient Descent}
\label{alg:socqp-pgd}
\begin{algorithmic}[1]
\State Choose $\rho$, initial $f\in\mathcal K$
\While{not converged}:
    \State $w\gets r+(G + R)f\Delta t$
    \State $f\gets\Pi_{\mathcal K}(f-\rho w)$
\EndWhile
\State \Return $f$
\end{algorithmic}
\end{algorithm}
The PGD algorithm is a special case (with simultaneous updates and a uniform step size) of the matrix-free method of Tasora and Anitescu~\cite{TasoraAnitescu2011}. A variant with Nesterov acceleration was proposed by Mazhar et al.~\cite{mazhar2015nesterov} and is implemented in the Chrono simulator~\cite{tasora2016chrono}. Its main downside is slow convergence on ill-conditioned problems. The step size must satisfy $\rho < 2/\lambda_{\max}$ for the PGD algorithm to converge, with $\lambda_{\max}$ denoting the largest eigenvalue of $(G + R)\Delta t$. This means progress along directions of small curvature is slow, and the number of iterations grows with the condition number of $(G + R)\Delta t$, which reflects both inertial disparities and the geometry and redundancy of the contact constraints, as well as the regularization strength.

The alternating direction method of multipliers (ADMM)~\cite{boyd2011distributed} greatly reduces this dependence, because it handles the curvature of the objective with an exact linear solve instead of a gradient step. It introduces an unconstrained decision variable $\hat{f}$, coupled to $f$ through the equality constraint $\hat{f} = f$, which is imposed progressively through a (scaled) Lagrange multiplier $u$ and an augmented Lagrangian penalty with weight $\beta$. Each iteration then splits into solving an unconstrained quadratic problem in $\hat{f}$, and a projection of $\hat{f} + u$ onto the friction cone:

\begin{algorithm}[htb]
\caption{Alternating Direction Method of Multipliers (ADMM)}
\label{alg:socqp-admm}
\begin{algorithmic}[1]
\State Choose initial $f\in\mathcal K$
\State $u\gets0$
\State Factorize $(G + R)\Delta t+\beta \Id$
\While{not converged}:
    \State Solve $\displaystyle
        ((G + R)\Delta t+\beta \Id)\hat{f}
        =\beta(f-u)-r$ for $\hat{f}$
    \State $f\gets\Pi_{\mathcal K}\left(\hat{f}+u\right)$
    \State $u\gets u+\hat{f}-f$
\EndWhile
\State \Return $f$
\end{algorithmic}
\end{algorithm}
The matrix $(G + R)\Delta t+\beta \Id$ does not change between iterations, so it is factorized only once, as long as $\beta$ is fixed. The convergence speed depends strongly on $\beta$, and adaptive updates of $\beta$ require a refactorization~\cite{carpentier2024compliant}. ADMM converges for any $\beta > 0$ on convex problems~\cite{boyd2011distributed}, and it is markedly more robust than PGS and PGD on ill-conditioned problems~\cite{tasora2021solving, carpentier2024compliant}. For the cone complementarity problem of contact dynamics, it was used by Tasora et al.~\cite{tasora2021solving} and is available in Chrono~\cite{tasora2016chrono}. Carpentier et al.~\cite{carpentier2024compliant} combined it with the proximal and de Saxc\'e updates of \cref{sec:asocqp_sol}, as implemented in Simple~\cite{simple_simulator_2025} and later by Kamino~\cite{tsounis2026kamino}.

ADMM handles the curvature of the objective exactly, but treats cone constraints only through projections and makes no use of their first- or second-order information. Interior-point methods take this next step and typically converge in far fewer iterations~\cite{NocedalWright2006}. They enforce the cone constraints smoothly through a convex logarithmic barrier term added to the objective. For $\mu_i>0$, the barrier term is
\begin{equation}
    B(f)=-\sum_{i=1}^N
        \log\bigl(\mu_i^2 f_{i,n}^2-\|f_{i,t}\|^2\bigr),
    \qquad f\in\interior\mathcal K.
\end{equation}
The combined objective is $L_\tau(f)=L(f)+\tau B(f)$, where $\tau>0$ is reduced toward zero each time Newton's method has solved the barrier subproblem; as $\tau \to 0$, the minimizers of $L_\tau$ approach the solution of the QCQP. Frictionless contacts instead use the scalar variable $f_{i,n}>0$ and the barrier $-\log f_{i,n}$.

\begin{algorithm}[htb]
\caption{Interior-Point Newton}
\label{alg:socqp-interior-point}
\begin{algorithmic}[1]
\State Choose initial $f\in\interior\mathcal K$ and $\tau>0$
\While{not converged}:
    \While{barrier subproblem not converged}:
        \State $w\gets r+(G + R)f\Delta t$
        \State $g\gets w+\tau\nabla B(f)$
        \State $H\gets(G + R)\Delta t+\tau\nabla^2 B(f)$
        \State Solve $Hd=-g$
        \State Choose $\alpha\in(0,1]$ by feasible line search
        \State on the barrier objective $L_\tau$
        \State $f\gets f+\alpha d$
    \EndWhile
    \State Reduce $\tau$ if the QCQP has not converged
\EndWhile
\State \Return $f$
\end{algorithmic}
\end{algorithm}
An early application of interior-point methods to contact simulation is the custom primal interior-point method of Todorov~\cite{Todorov2011}. Acary et al.~\cite{acary2024second} develop and evaluate a primal--dual interior-point algorithm for the convex second-order cone relaxation of frictional contact, providing a contact-specific treatment of the QCQP approach described here. In robotics, Dojo~\cite{howell2022dojo} applies a primal--dual interior-point method to the exact N-CCP transcription, and Pang et al.~\cite{pang2023global} use the barrier relaxation to smooth a quasi-dynamic contact model. The main benefit of an interior-point method is that, for a fixed $\tau > 0$, the barrier solution depends smoothly on the problem parameters~\cite{tracy2024differentiability, pang2023global}, which is particularly valuable whenever the contact dynamics need to be differentiated. An interior-point iteration makes more progress toward the optimum than an ADMM or PGD iteration, but the Hessian changes in every iteration and must be refactorized, and interior-point methods are hard to warm-start~\cite{carpentier2024compliant}. With a single factorization and cheap iterations, ADMM therefore often strikes a balance between the cheap but slowly converging PGD and the expensive but quickly converging interior-point method.

\subsubsection{NLS Transcription}
The NLS transcription is an unconstrained, strongly convex and continuously differentiable program, which makes it a prime target for Newton's method.

\begin{algorithm}[htb]
\caption{Nonsmooth Newton (Todorov)}
\label{alg:nls-newton}
\begin{algorithmic}[1]
\State Choose initial $\ddot{q}\in\mathbb R^{n_v}$
\While{not converged}:
    \State $y\gets a_{\mathrm{ref}}-J\ddot{q}$
    \State $f\gets R^{-1}\Pi_{\mathcal K}(y)$
    \State Assemble $K=\partial\Pi_{\mathcal K}(y)$
    \State $g\gets M\left(\ddot{q}-\ddot{q}_f\right)-J^\trsp f$
    \State $H\gets M+J^\trsp R^{-1}KJ$
    \State Solve $Hd=-g$
    \State Choose $\alpha\in(0,1]$ via line search on the NLS objective
    \State $\ddot{q}\gets\ddot{q}+\alpha d$
\EndWhile
\State \Return $\ddot{q}$
\end{algorithmic}
\end{algorithm}
The main benefit of this algorithm is that, unlike the algorithms introduced so far, its decision variable is $\ddot{q} \in \mathbb R^{n_v}$ instead of $f \in \mathbb R^{3N}$, so its linear systems have size $n_v \times n_v$ and inherit the sparsity of $M$. In contact-rich settings, $3N > n_v$ often holds, which makes these solves much cheaper than their force-space counterparts. The algorithm goes back to Todorov~\cite{MuJoCo150Release} and is the default solver of MuJoCo~\cite{todorov2012mujoco, mujoco_docs}. The SAP solver of Drake uses a similar construction in generalized velocities~\cite{castro_sim, drake}. The main numerical insight is to solve, instead of the force-space QCQP, its Lagrange dual (called the primal problem in MuJoCo), which is unconstrained and lives in the space of generalized accelerations.

\subsection{Exact Transcriptions}
\subsubsection{Signorini--Coulomb Transcription} As a nonlinear system of equations with inequality constraints, the Signorini--Coulomb transcription \eqref{eq:mcp} can also be solved by a Gauss--Seidel scheme, one that solves the local problem of each contact exactly.

\begin{algorithm}[htb]
\caption{Gauss--Seidel with Exact Local Solves (Moreau--Jean)}
\label{alg:mcp-gs}
\begin{algorithmic}[1]
\State Choose initial $f\in\mathcal K$
\While{not converged}:
    \For{$i$ in active contacts}:
        \State $v_i^{\circ}\gets(v_f)_i+\gamma_i+\Delta t\sum_{j\neq i}G_{ij}f_j$ \Comment{$f_i$ removed}
        \If{$v^{\circ}_{i,n}\geq 0$} \Comment{separating}
            \State $f_i\gets0$
        \Else
            \State $f_i\gets-(\Delta t\,G_{ii})^{-1}v_i^{\circ}$ \Comment{sticking candidate}
            \If{$f_{i,n}<0$ or $\|f_{i,t}\|>\mu_i f_{i,n}$} \Comment{sliding}
                \State Find $f_{i,n}>0$ and unit tangential $\hat{s}_i$ with
                \Statex \hspace{5.75em} $f_i = f_{i,n}\left(\hat{n}_i-\mu_i\hat{s}_i\right)$, $w_i=v_i^{\circ}+\Delta t\,G_{ii}f_i$,
                \Statex \hspace{5.75em} $w_{i,n}=0$, $w_{i,t}=\lambda \hat{s}_i$, $\lambda>0$
            \EndIf
        \EndIf
    \EndFor
\EndWhile
\State \Return $f$
\end{algorithmic}
\end{algorithm}
The algorithm works as follows. For contact $i$, it holds the forces of all other contacts fixed, computes the velocity $v_i^{\circ}$ that contact $i$ would have without its own force, and solves the local problem \crefrange{eq:mcp_2}{eq:mcp_4} for $f_i$ exactly. The local problem has three mutually exclusive outcomes, which are tried in order: the contact separates, sticks, or slides. The sliding branch is a nonlinear system in two unknowns, the normal force and the slip direction. In 2D it has a closed-form solution: with $\hat{s}_i = \pm 1$, the condition $w_{i,n}=0$ gives $f_{i,n} = -v^{\circ}_{i,n}/\big(\Delta t\,(G_{nn} \mp \mu_i G_{nt})\big)$ for each sign, and the admissible sign is the one for which $w_{i,t}$ has the sign of $\hat{s}_i$. In 3D it can be solved either by a few fixed-point iterations on the direction, $\hat{s}_i \gets w_{i,t}/\|w_{i,t}\|$ with $f_{i,n}$ recomputed from $w_{i,n}=0$ at each pass, or by a local Newton method.

The algorithm goes back to the nonsmooth contact dynamics framework of Moreau and Jean~\cite{moreau1994numerical, Jean1999}. The main idea it introduces is to break the problem down contact by contact and to treat it with exact local solves in a Gauss--Seidel fashion. Siconos implements it as its nonsmooth Gauss--Seidel solver with a local Newton solver~\cite{acary2007siconos}. RaiSim~\cite{hwangbo2018per} also sweeps contact by contact with an exact local solve (a bisection), but for a modified local law that keeps the Signorini condition and the friction cone and replaces the maximal-dissipation principle~\cite{lelidec2023contacts}. Compared to its L-CCP counterpart, the main benefits of it are that it addresses the exact transcription, does not require a schedule for $\rho$, and handles early termination more gracefully, since each contact satisfies its local law with respect to the current forces of its neighbors at the moment it is updated.

\subsubsection{N-CCP Transcription: Projected Gauss--Seidel}
\label{sec:nccp-pgs-methods}
The fixed-point equivalence \eqref{eq:pgs_2} used to solve the L-CCP transcription also applies when $F(x)$ is nonlinear. This gives a PGS method for the N-CCP transcription \eqref{eq:nccp} using the de Saxc\'e correction $\Gamma(v^+)$. We present two variants: using a single joint projection onto the full friction cone, and using separate normal and tangential projections. Both sweep through the contacts, updating each individual force using the latest values of all the other forces.

\smallbreak
\textbf{Projection onto the Full Friction Cone.}
The PGS scheme of \cref{alg:nccp-pgs} replaces the three-branch local solve of \cref{alg:mcp-gs} with one inexpensive, closed-form projection onto the friction cone. The de Saxc\'e correction is required in this full-cone formulation to recover the Signorini--Coulomb law~\cite{DeSaxceFeng1991, DeSaxceFeng1998}. Siconos provides this update as a local cone-projection solver within its nonsmooth Gauss--Seidel framework~\cite{acary2007siconos, siconosdocs}.

\begin{algorithm}[htb]
\caption{Projected Gauss--Seidel (Full-Cone Projection)}
\label{alg:nccp-pgs}
\begin{algorithmic}[1]
\State Choose initial $f\in\mathcal K$
\While{not converged}:
    \For{$i$ in active contacts}:
        \State $\displaystyle v_i^+\gets(v_f)_i
            +\Delta t\sum_{j=1}^NG_{ij}f_j$
        \State $w_i\gets v_i^++\Gamma(v_i^+)+\gamma_i$
        \State $f_i\gets\Pi_{\mathcal K_i}(f_i-\rho_iw_i)$
    \EndFor
\EndWhile
\State \Return $f$
\end{algorithmic}
\end{algorithm}
\smallbreak

\smallbreak
\textbf{Separate Normal and Tangential Projections.}
An alternative, used in the SORprox approach of Studer and Glocker~\cite{StuderGlocker2007}, treats the normal and tangential conditions separately. For each contact, it first projects the normal force onto the non-negative half-line, then projects the tangential force onto the friction disk whose radius is set by the updated normal force. Define \mbox{$\mathcal D_i(\eta) = \left\{z\in\mathbb R^2 : \|z\|\leq\mu_i\eta\right\}$} for \mbox{$\eta\geq0$}. A sweep with positive scalar step sizes proceeds as follows:
\begin{algorithm}[htb]
\caption{Projected Gauss--Seidel (Normal and Tangential Projections)}
\label{alg:nccp-pgs-split}
\begin{algorithmic}[1]
\State Choose initial $f\in\mathcal K$ and step sizes $\rho_{i,n},\rho_{i,t}>0$
\While{not converged}:
    \For{$i$ in active contacts}:
        \State $\displaystyle v_i^+\gets(v_f)_i+\Delta t\sum_{j=1}^N G_{ij}f_j$
        \State $f_{i,n}\gets\max\left(0,\;f_{i,n}-\rho_{i,n}\left(v_{i,n}^++\gamma_{i,n}\right)\right)$
        \State $\displaystyle v_i^+\gets(v_f)_i+\Delta t\sum_{j=1}^N G_{ij}f_j$ \Comment{use the updated normal force}
        \State $f_{i,t}\gets\Pi_{\mathcal D_i(f_{i,n})}\left(f_{i,t}-\rho_{i,t}v_{i,t}^+\right)$
    \EndFor
\EndWhile
\State \Return $f$
\end{algorithmic}
\end{algorithm}
The velocity is recomputed between the two projections to account for coupling between normal and tangential forces. At a fixed point, the normal projection enforces the stabilized Signorini condition, while the disk projection enforces maximum tangential dissipation for the current normal force. Thus, the fixed points satisfy the exact contact law without an explicit de Saxc\'e term. A single update need not solve the local contact problem exactly, and convergence still depends on the step sizes and coupling between contacts.

Simple~\cite{simple_simulator_2025} implements this normal-then-tangential disk projection through Pinocchio's PGS solver~\cite{carpentier2019pinocchio}. Related normal/friction splitting is used by Bullet's sequential-impulse solver~\cite{catto2005iterative, coumans2015bullet} and ODE's QuickStep~\cite{smith2006ode}; their componentwise friction bounds differ from the Euclidean disk projection shown here.

\subsubsection{N-CCP Transcription: Nonsmooth Newton}
\label{sec:nccp-newton-methods}
The same full-cone projection relation can be solved by applying a nonsmooth Newton method to its fixed-point residual. Each iteration updates all contact forces simultaneously by solving a linearized system that includes their coupling through $G$ (\cref{alg:nccp-newton}). Linearizing this residual requires the Jacobian $\partial\Gamma$ of the de Saxc\'e correction, assembled contact by contact from the closed-form expression
\begin{equation}
    \partial\Gamma(v^+_i)=
    \begin{bmatrix}
        0 & \mu_i\hat{s}_i^\trsp \\
        0 & 0_{2\times2}
    \end{bmatrix},
    \qquad
    \hat{s}_i=
    \begin{cases}
        v^+_{i,t}/\|v^+_{i,t}\|, & v^+_{i,t}\neq0, \\
        0, & v^+_{i,t}=0
    \end{cases},
\end{equation}
where, at $v^+_{i,t} = 0$, $\Gamma$ is not differentiable and the choice $\hat{s}_i = 0$ selects an element of its generalized Jacobian.
The Newton method of \cref{alg:nccp-newton} is the natural-map variant of the method of Alart and Curnier~\cite{AlartCurnier1991, acary2018solving}. Bertails-Descoubes et al.~\cite{bertails2011nonsmooth} apply this nonsmooth root-finding approach to exact Coulomb friction in fiber assemblies, illustrating how the same contact formulation extends beyond rigid bodies. Siconos provides an implementation as part of its nonsmooth Newton solvers~\cite{acary2007siconos}.

\begin{algorithm}[t]
\caption{Nonsmooth Newton (Alart--Curnier, Exact)}
\label{alg:nccp-newton}
\begin{algorithmic}[1]
\State Choose initial $f\in\mathcal K$
\While{not converged}:
    \State $v^+\gets v_f+Gf\Delta t$
    \State $w\gets v^++\Gamma(v^+)+\gamma$
    \State $y\gets f-\rho w$
    \State $e\gets f-\Pi_{\mathcal K}(y)$
    \State Assemble $K = \partial\Pi_{\mathcal K}(y)$
    \State Assemble $C = \partial\Gamma(v^+)$
    \State $A\gets \Id-K\bigl(\Id-\rho(\Id+C)G\Delta t\bigr)$
    \State Solve $Ad=-e$
    \State Choose $\alpha\in(0,1]$ via residual line search
    \State $f\gets f+\alpha d$
\EndWhile
\State \Return $f$
\end{algorithmic}
\end{algorithm}

Interior-point methods can also be applied directly to the exact N-CCP transcription. Acary, Armand, and Nguyen~\cite{acary2026solving} combine a primal--dual interior-point method with semismooth Newton iterations and path-following techniques to handle the nonsmooth, non-associated Coulomb law. Under their stated assumptions, they establish an analytic central path converging to a contact solution. Their treatment of redundant constraints and continuation along this path provides a further approach to the robustness difficulties caused by rank deficiency.

\subsubsection{Root-Finding Transcription}

The root-finding transcription \eqref{eq:root_finding} is a system of $6N$ equations with $6N$ unknowns $v^+$ and $f$ that has no inequality constraints, so Newton's method applies to it directly. Since the dynamics \eqref{eq:nse_1} are linear, substituting \mbox{$v^+ = v_f + G f \Delta t$} into \eqref{eq:nse_2} eliminates the velocity and leaves $3N$ equations in the contact forces alone. To write them in terms of the SOCFB function $\psi$ on the Lorentz cone, let us collect the rescalings \eqref{eq:lorentz_map} of all contacts in the block-diagonal matrices $P$ and $Q$, with the blocks $\operatorname{diag}(\mu_i, 1, 1)$ and $\operatorname{diag}(1, \mu_i, \mu_i)$, so that $x = Pf$ and $y = Qw$ with $w = v^+ + \Gamma(v^+) + \gamma$. The equations of contact $i$ then read
\begin{equation} \label{eq:fb_residual}
    b_i = \psi_{\mathcal K}(x_i, y_i) = 0,
\end{equation}
where $b_i$ is the residual of contact $i$. Collecting the Jacobians \eqref{eq:socfb_jac} of all contacts in the block-diagonal matrices $\Psi_x$ and $\Psi_y$, with the blocks $\partial_x\psi(x_i, y_i)$ and $\partial_y\psi(x_i, y_i)$, the chain rule through $x$, $y$, $w$, and $v^+$ gives the Newton matrix
\begin{equation} \label{eq:fb_newton_matrix}
    A = \Psi_x P + \Psi_y Q\,(\Id + C)\,G\Delta t,
\end{equation}
where $C = \partial\Gamma(v^+)$ is the Jacobian of the de Saxc\'e correction, assembled contact by contact. The resulting algorithm is:

\begin{algorithm}[t]
\caption{Nonsmooth Newton (Fischer--Burmeister)}
\label{alg:nse-newton}
\begin{algorithmic}[1]
\State Choose initial $f$, smoothing $\epsilon \geq 0$

\While{not converged}:
    \State $v^+\gets v_f+Gf\Delta t$
    \State $w\gets v^++\Gamma(v^+)+\gamma$
    \State $x\gets Pf$, \ $y\gets Qw$
    \State $b_i\gets\psi_\epsilon(x_i,y_i)$ for every contact $i$ \Comment{\eqref{eq:socfb_smooth}}
    \State Assemble $\Psi_x$, $\Psi_y$ via \eqref{eq:socfb_jac_eps}, and $C = \partial\Gamma(v^+)$
    \State $A\gets\Psi_xP+\Psi_yQ(\Id+C)G\Delta t$
    \State Solve $Ad=-b$
    \State Choose $\alpha\in(0,1]$ via line search
    \State $f\gets f+\alpha d$
\EndWhile
\State \Return $f$
\end{algorithmic}
\end{algorithm}
The smoothing factor $\epsilon$ is usually decreased toward zero over the iterations. \Cref{alg:nse-newton} and \cref{alg:nccp-newton} are both nonsmooth Newton methods on a root-finding reformulation of the N-CCP conditions. They differ mainly in the function that turns the cone complementarity condition into equations: the natural map uses the projection residual $f - \Pi_{\mathcal K}(f - \rho w)$, which depends on the parameter $\rho$, whereas the SOCFB function needs no such parameter. Newton methods on Fischer--Burmeister reformulations go back to Fischer~\cite{fischer1992special} for scalar complementarity problems. For exact Coulomb friction, Daviet et al.~\cite{daviet2011hybrid} use a modified SOC Fischer--Burmeister residual in a hybrid Gauss--Seidel solver: each contact is solved by a local Newton method, with an analytical enumerative solver as a fallback. Siconos provides an implementation as part of its family of nonsmooth Newton solvers~\cite{acary2007siconos, acary2018solving}. The main insights the algorithm adds are that complementarity conditions can be converted to equality conditions and can be smoothed as such. With the smoothed SOCFB function at a fixed $\epsilon > 0$, the solution depends smoothly on the problem data, again enabling differentiability like the interior-point method. As for the other Newton methods, convergence is only local, and the Newton matrix can still be singular when $G$ is rank deficient. A PGS sweep is again commonly employed as a fallback.

\subsubsection{Coupled Quadratic Program (C-QP) Transcription} The C-QP transcription \eqref{eq:cqp} splits the unknowns into three blocks: normal forces $f_n$, friction magnitudes $\norm{f_t}$, and tangential directions $\bar{t}$. With the other two blocks held fixed, each block is easy to update. The QPs \eqref{eq:cqp_3} and \eqref{eq:cqp_4} are convex and box-constrained, so any solver for this well-studied problem class can be applied to them. The direction updates follow from the conditions \eqref{eq:cqp_2} interpreted as a fixed point iteration. This gives a staggered scheme:

\begin{algorithm}[t]
\caption{Staggered Box QPs}
\label{alg:cqp-staggered}
\begin{algorithmic}[1]
\State Choose initial $f_n$, $\norm{f_t}$ 
\State Choose unit tangents $\hat{t}_i$ \Comment{e.g.,\ against $(v_f)_{i,t}$}
\While{not converged}:
    \State $f_n \gets$ solution of \eqref{eq:cqp_3} \Comment{$\norm{f_t}$, $\bar{t}$ fixed}
    \State $\norm{f_t} \gets$ solution of \eqref{eq:cqp_4} \Comment{$f_n$, $\bar{t}$ fixed}
    \State $v^+ \gets v_f + G\big(\bar{n} f_n + \bar{t}\norm{f_t}\big)\Delta t$
    \For{$i$ in active contacts}:
        \State $u_i \gets \norm{f_{i,t}}\hat{t}_i - \rho_i v^+_{i,t}$
        \If{$u_i \neq 0$}
            \State $\hat{t}_i \gets u_i / \norm{u_i}$
        \EndIf
    \EndFor
\EndWhile
\State \Return $f = \bar{n} f_n + \bar{t}\norm{f_t}$
\end{algorithmic}
\end{algorithm}

The direction update points $\hat{t}_i$ along $u_i = \norm{f_{i,t}}\hat{t}_i - \rho_i v^+_{i,t}$, the tangential part of a gradient step on the friction force $f_{i,t} = \norm{f_{i,t}}\hat{t}_i$ with a step size $\rho_i > 0$. It leaves $\hat{t}_i$ unchanged at a sticking contact (where $v^+_{i,t} = 0$) as well as at a sliding contact whose slip velocity already points against $\hat{t}_i$. Otherwise, it turns $\hat{t}_i$ toward $-v^+_{i,t}$. The fixed points of the iteration therefore satisfy the direction conditions \eqref{eq:cqp_2} and solve the C-QP transcription. That being said, the iteration itself has no general convergence guarantee~\cite{kaufman2008staggered}.
Splitting the problem into a normal and a tangential problem goes back to Panagiotopoulos~\cite{panagiotopoulos1975nonlinear}. Siconos implements it as its Panagiotopoulos fixed-point solver, which solves the normal problem with PGS and the tangential problem with a projection-based convex QP solver~\cite{acary2007siconos, siconosdocs}. The staggered projections of Kaufman et al.~\cite{kaufman2008staggered} alternate the same two convex problems. Unlike \cref{alg:cqp-staggered}, both implementations directly optimize the full tangential force (over the friction disk, or over a polygon inscribed in it), therefore they need no direction update.
 The main insight of \cref{alg:cqp-staggered} is the decomposition of the contact dynamics problem into sub-problems that are box-constrained QPs, a well-studied and particularly easy to solve class of convex optimization problems.

\subsubsection{Rank Deficiency and Proximal Minimization}

As discussed in \cref{sec:formulation}, the Delassus matrix $G$ is often rank-deficient, which can make linear solves singular or poorly conditioned. The methods above address this difficulty in different ways: convex relaxations introduce the regularization matrix $R$, while ADMM and interior-point methods add curvature through the augmented Lagrangian and barrier terms, respectively. PGS and PGD avoid a linear solve with the full matrix $G$ and can therefore serve as fallbacks when a Newton system is singular.

\smallbreak
Proximal optimization combines regularization with an iterative solution strategy: each subproblem penalizes deviations from the previous iterate, and updating this reference point allows the method to approach a solution of the original problem while retaining regularization in every subproblem. Building on Moreau's proximity operator~\cite{moreau1965proximite}, Martinet~\cite{martinet1970regularisation} introduced this iterative construction, whose general convergence theory, including inexact subproblem solves, was subsequently developed by Rockafellar~\cite{rockafellar1976monotone}. In robotics, Carpentier et al.~\cite{carpentier2021proximal} applied this principle to constrained dynamics with redundant or singular kinematic constraints, combining a proximal formulation with a sparse factorization that exploits the robot's kinematic structure.

\begin{algorithm}[htb]
\caption{Proximal Minimization}
\label{alg:proximal-point}
\begin{algorithmic}[1]
\State Choose $\rho > 0$, $D \succ 0$, $x^0\in\mathcal C$; set $z \gets x^0$
\While{not converged}:
    \For{$k = 1, \dots, m$}
        \State Do an optimization update on
        \State $\min_{x\in\mathcal C}\;L(x) +\tfrac\rho2\|x-z\|_{D}^2$
        \State using any constrained optimization
        \State algorithm to obtain $x^*$.
    \EndFor
    \State $z \gets x^*$
\EndWhile
\State \Return $x^*$
\end{algorithmic}
\end{algorithm}
The algorithm repeatedly adds a quadratic penalty centered at the previous iterate $z$ and applies $m$ optimization updates to the resulting problem $\min_{x\in\mathcal C} L(x) +\tfrac\rho2\|x-z\|_{D}^2$. The penalty adds \mbox{$\rho D \succ 0$} to the Hessian, so every inner problem is strongly convex. At a fixed point $z = x^*$, the penalty vanishes, so $x^*$ solves the original problem. For convex $L$, the outer iteration converges to a minimizer even when $L$ is only positive semi-definite, provided the inner problems are solved accurately enough~\cite{rockafellar1976monotone}. The main benefit of the framework is its flexibility, as any optimization algorithm can be wrapped in an outer proximal minimization loop.

\subsubsection{C-QCQP with Proximal Minimization}\label{sec:asocqp_sol}

Adding the proximal penalty $\tfrac\rho2\|f-z\|_{D}^2$ to the QCQP objective \eqref{eq:c_qcqp_2} and collecting linear and quadratic terms together gives
\begin{equation}\label{eq:socqp_prox}
    \min_{f \in \mathcal{K}} \: \frac{1}{2}f^\trsp \left(\left(G + \frac{\rho}{\Delta t}D\right) \Delta t\right)f + (v_f + \Gamma(\sigma) + \gamma - \rho Dz)^\trsp f.
\end{equation}

For fixed $\sigma$ and $z$, and for a diagonal $D \succ 0$, this problem has exactly the strictly positive-definite and strongly convex form of \eqref{eq:socqp}, with $R = \frac{\rho}{\Delta t}D$ and $r = v_f + \Gamma(\sigma) + \gamma - \rho Dz$. Therefore, all the solution algorithms and numerical insights described above for the regularized QCQP transcription, and for the equivalent regularized L-CCP and NLS transcriptions, can be used to solve \eqref{eq:socqp_prox}. Combining this observation with a fixed-point scheme that iterates \crefrange{eq:c_qcqp_1}{eq:c_qcqp_2} of the C-QCQP transcription \eqref{eq:c_qcqp} gives the following general blueprint for deriving solution algorithms:
\begin{algorithm}[t] %
\caption{Proximal Minimization with de Saxc\'e Correction}
\label{alg:proximal-point-correction}
\begin{algorithmic}[1]
\State Choose $\rho > 0$, diagonal $D \succ 0$, $R = \frac{\rho}{\Delta t}D$
\State Initialize $z\in\mathcal K$, $\sigma \gets v_f$
\While{not converged}:
    \State $r \gets v_f + \Gamma(\sigma) + \gamma - \rho Dz$
    \For{$k = 1, \dots, m$}
        \State Do a solver iteration on any of the
        \State three equivalent relaxed transcriptions:
        \State (i)~$\min_{f\in \mathcal{K}} \: \frac{1}{2}f^\trsp  \left([(G + R)\Delta t]\right)f + r^\trsp f$
        \State (ii)~$\text{find} \ f: \ \mathcal{K} \ni f \perp ([(G + R)\Delta t] f  + r) \in \mathcal{K} ^*$
        \State (iii)~$\min_{\ddot{q}} \: \frac{1}{2}\left(\ddot{q} - \ddot{q}_f\right)^\trsp M \left(\ddot{q} - \ddot{q}_f\right)$
        \State $\quad \quad \quad \quad + \frac{1}{2}\sum_{i=1}^N R^{-1}_{ii}\left\|\Pi_{\mathcal{K}_i}\left(a_\text{ref, i} - J_i \ddot{q}\right)\right\|^2$
        \State using any of the corresponding algorithms
        \State covered in previous sections, to obtain $f^*$.
    \EndFor
    \State $z \gets f^*$
    \State $\sigma \gets v_f+Gf^*\Delta t$
\EndWhile
\State \Return $f^*$
\end{algorithmic}
\end{algorithm}
The algorithm wraps the solver iterations for the approximate transcriptions in an outer loop that updates the proximal anchor $z$ and the de Saxc\'e correction velocity $\sigma$. At a fixed point, $z = f^*$ and \mbox{$\sigma = v_f + G f^* \Delta t$} both hold, so the proximal terms cancel, and the optimality condition of the inner problem, \mbox{$\mathcal{K} \ni f^* \perp \big((G\Delta t) f^* + v_f + \Gamma(\sigma) + \gamma\big) \in \mathcal{K}^*$}, is exactly the N-CCP transcription \eqref{eq:nccp}: the fixed points of the blueprint are the solutions of the exact problem. For example, running ADMM on the regularized QCQP transcription~(i) with a single iteration per outer iteration ($m = 1$) gives the routine proposed by Carpentier et al.~\cite{carpentier2024compliant} (up to their adaptive update of the ADMM penalty), which is also implemented in the Kamino simulator~\cite{tsounis2026kamino}.

\section{Comparative Analysis}\label{sec:comparison}
\Cref{sec:formulation,sec:solution} presented the transcriptions and the solution algorithms one at a time. The goal of this section is to draw comparisons between them. We first ask whether each transcription has a solution and whether that solution is unique (\cref{sec:cmp_existence}). We then compare the transcriptions by the physical behavior they produce (\cref{sec:cmp_physics}), the algorithms by their cost, convergence, and robustness (\cref{sec:cmp_solvers}), and the simulators that implement them on CPUs and GPUs (\cref{sec:cmp_implementations}). We finally provide guidelines for choosing among them (\cref{sec:cmp_choice}).

\begin{table*}[t]
\centering\small\setlength{\tabcolsep}{4pt}
\begin{tabular}{P{4.0cm} P{1.2cm} P{3.4cm} P{1.6cm} P{1.9cm} P{3.6cm}}
\toprule
Transcriptions & Convex & Solution exists & Unique motion & Unique forces & Physical artifact \\
\midrule
Signorini--Coulomb, N-CCP, root-finding, C-QCQP, C-QP (exact) & no & under a strict feasibility assumption~\cite{acary2011formulation} & not in general & not in general & none \\
L-CCP, QCQP & yes & if admissible velocities exist, e.g.,\ for $\gamma_n \geq 0$ & yes & if $G$ is nonsingular & sliding contacts lift off \\
Regularized L-CCP and QCQP, NLS & strongly & always & yes & yes & lift-off, creep, compliant normal response \\
LCP (\cref{sec:lcp}) & no & always~\cite{anitescu1997formulating} & not in general & not in general & anisotropic friction \\
\bottomrule
\end{tabular}
\caption{\captitle{Existence and uniqueness of solutions.} For each group of transcriptions, the table lists the following information: whether the problem is convex, under which conditions a solution exists, whether the motion (i.e., the generalized velocity after the step) and the contact forces are unique, and which physical artifact the transcription introduces relative to the exact contact law (\cref{sec:cmp_existence,sec:cmp_physics}).}
\label{tab:cmp_existence}
\end{table*}

\subsection{Existence and Uniqueness of Solutions}\label{sec:cmp_existence}

Before comparing physical behavior or computational cost, we must ask whether each transcription admits a solution and whether that solution is unique. Two distinct notions of uniqueness matter here: the motion of the bodies and the contact forces that produce it. Redundant contacts can admit several force distributions for the same motion, whereas friction can also lead to different admissible motions. These properties belong to the mathematical formulation; when several solutions exist, the numerical algorithm can influence which one is selected. We examine these questions for the exact transcriptions, the convex relaxations, and the LCP, then discuss this role of the solver. \Cref{tab:cmp_existence} summarizes the comparison.

\smallbreak
\textbf{Exact transcriptions.} The five exact transcriptions are equivalent, so the discussion applies equally to all of them. The existence of solutions is not guaranteed without further assumptions. Using the C-QCQP form, Acary et al.~\cite{acary2011formulation} proved that a solution can be found whenever there exists a generalized velocity $\dot{q}$ that places $J_i\dot{q} + \gamma_i$ in the interior of $\mathcal{K}_i^*$ at every contact (i.e., opens every contact with a normal velocity larger than $\mu_i$ times its tangential velocity). This sufficient condition holds in most practical situations, for example a box on a table can be lifted off all its contacts at once, and fails mainly for bodies wedged between opposing contacts. Uniqueness can fail in two different ways~\cite{stewart2000rigid, acary2018solving}. First, when $G$ is rank deficient (e.g., when the contacts outnumber the degrees of freedom they constrain) many force distributions can produce the same motion. For example, for a box resting on four corner contacts, only the sum of the four contact forces is uniquely determined. Second, with friction, even when $G$ is full rank the motion after the step can have several solutions, the discrete-time counterpart of the non-uniqueness found by Painlev\'e in continuous time (\cref{sec:history}).

\smallbreak
\textbf{Convex relaxations.} As the kinetic energy perspective of \cref{sec:kinmin} suggests, the L-CCP and QCQP transcriptions can be interpreted as the optimality conditions of a convex projection problem. The end-of-step generalized velocity $\dot{q}^+$ is the projection (under the metric induced by $M$) of the free velocity $\dot{q}_f$ onto the set of velocities $\dot{q}$ with $J\dot{q} + \gamma \in \mathcal{K}^*$. Therefore, there exists a unique $\dot{q}^+$ whenever this constraint set is nonempty. The cases where the admissible set is empty again involve non-physical configurations such as bodies wedged between opposing walls.
The set of admissible forces in turn form a convex set, which reduces to a single point whenever $G$ is nonsingular. The regularized transcriptions are strongly convex and therefore always have a unique solution. Among the many force distributions that produce the same motion, the regularization selects the one with the smallest weighted norm $f^\trsp R f$, which spreads the load over redundant contacts.

\smallbreak
\textbf{LCP.} For the velocity-level LCP of \cref{sec:lcp}, a solution always exists, and Lemke's pivoting algorithm finds one for any number of contacts~\cite{anitescu1997formulating}. Uniqueness again fails in general.

\smallbreak
\textbf{How the algorithms choose.} When a transcription does not identify a unique solution, it is the implementation details of the numerical routine that determines which one is selected. Warm-started iterative methods, such as PGS, PGD, ADMM, and the proximal minimization approach of \cref{alg:proximal-point-correction}, return a solution that depends on their initialization. For Gauss--Seidel type of methods, the final solution that is picked also depends on the order of the updates. Interior-point methods approach the solution set from its interior, and tend to spread the force over redundant contacts due to the barrier penalty. For Newton methods applied to the unregularized transcriptions, the choice of the linear solver for handling underdetermined systems also effects the solution obtained. Regularization removes such problems due to non-uniqueness, at the price of the physical artifacts discussed next.

\subsection{Physical Trade-offs Between the Transcriptions}\label{sec:cmp_physics}
The exact transcriptions model the same discrete-time contact law without approximation, so choosing among them only effects which solution algorithms are suitable and not the solution itself (\cref{sec:cmp_solvers}). The other transcriptions trade physical fidelity for computational convenience in three distinct ways, listed in \cref{tab:cmp_existence}.

\textbf{Dropping the de Saxc\'e term} makes the problem convex, but changes the normal law at sliding contacts. At a sliding contact, cone complementarity places $v^+_i + \Gamma(v^+_i) + \gamma_i$ on the boundary of $\mathcal{K}^*_i$, where its normal component equals $\mu_i$ times its tangential norm. In the exact transcriptions, the de Saxc\'e term supplies this normal component, so that $v^+_{i,n} + \gamma_{i,n} = 0$. Without it, the normal velocity has to supply it, so that $v^+_{i,n} + \gamma_{i,n} = \mu_i\|v^+_{i,t}\|$ holds. This means sliding contacts separate at a rate proportional to their tangential speed. A box pushed across a table therefore rises off the table while it slides, instead of staying in contact. The effect grows with the friction coefficient and the tangential speed, and it disappears at sticking contacts, where the relaxed and exact contact laws coincide.

\textbf{Regularization} makes the problem strongly convex, so that its solution is unique, and it turns the dynamics into the invertible form of \cref{sec:inverse_dynamics}. Its price is a compliant contact law. In the normal direction, a contact penetrates in proportion to the force it transmits. In the tangential direction, static friction is lost. 
This loss of static friction can be read off directly from the inverse dynamics \eqref{eq:primal_inverse_dyn}. As discussed in \cref{sec:inverse_dynamics}, the regularized transcriptions drive the system with the contact forces $R^{-1}_{ii}\,\Pi_{\mathcal{K}_i}\left(a_\text{ref, i} - J_i \ddot{q}\right)$, and the reference acceleration $a_\text{ref, i}(q, \dot{q})$ is always defined such that its tangential component is zero in the absence of any velocity (i.e., $a_\text{ref, i, t}(q, 0) = 0$). Whenever $R \succ 0$, $\dot{q} = 0$, and $\ddot{q} = 0$ all hold, the contact forces therefore cannot contain a tangential component. Only when $R = 0$, for which the equivalence between the QCQP \eqref{eq:socqp} and NLS \eqref{eq:nls} transcriptions no longer holds, can a contact transmit a tangential force in a state with zero velocity and zero acceleration. As a consequence, every contact that needs to transmit a tangential force to maintain staticity instead slips. For example, a pen held between two fingers under gravity slips downward, regardless of how strongly the fingers press against it. The same mechanism can be observed directly in the QCQP transcription. Without regularization, \eqref{eq:socqp_unreg} solves for the exact force needed to prevent any acceleration. The term $f^\trsp R f$ added in \eqref{eq:socqp} scales the optimal contact forces down below this value, so the pen slips.

\textbf{Replacing the circular cone by a pyramid} turns the problem into an LCP (\cref{sec:lcp}), which always has a solution and can be solved exactly by pivoting (i.e., not merely to numerical convergence like other methods). The cost is anisotropic friction: the tangential contact force depends on the direction of sliding relative to the faces of the pyramid, and it no longer opposes the slip exactly.

\begin{table*}[t]
\centering\footnotesize\setlength{\tabcolsep}{3pt}
\begin{tabular}{P{2.3cm} P{1.5cm} P{1.8cm} P{2.0cm} P{2.2cm} P{1.9cm} P{1.0cm} P{1.6cm} P{1.2cm}}
\toprule
Algorithm & Transcription & Linear solve per iteration & Convergence & Singular $G$; large mass ratios & GPU suitability & Warm start & Early termination & Smooth solutions \\
\midrule
PGS (Alg.~\ref{alg:lccp-pgs}) & L-CCP & none & global; at best linear & robust; slow & Jacobi or coloring variants & yes & feasible forces & no \\
Newton, natural map (Alg.~\ref{alg:lccp-newton}) & L-CCP & $3N \times 3N$ & local, superlinear & may be singular; robust & batched solves & yes & may leave $\mathcal{K}$ & no \\
PGD (Alg.~\ref{alg:socqp-pgd}) & QCQP & none & global; linear, grows with the condition number & robust; slow & fully parallel & yes & feasible forces & no \\
ADMM (Alg.~\ref{alg:socqp-admm}) & QCQP & one factorization, reused & global; speed depends on $\beta$ & robust; robust & batched solves & yes & feasible forces & no \\
Interior point (Alg.~\ref{alg:socqp-interior-point}) & QCQP & $3N \times 3N$, refactorized & global; few iterations & robust; robust & batched solves & hard & strictly feasible & yes, for $\tau > 0$ \\
Newton (Alg.~\ref{alg:nls-newton}) & NLS & $n_v \times n_v$, sparse & global; few iterations & robust; robust & batched solves & yes & feasible forces & no \\
\midrule
Exact local GS (Alg.~\ref{alg:mcp-gs}) & Signorini--Coulomb & none & no guarantee; at best linear & robust; slow & sequential, branching & yes & graceful & no \\
PGS (Alg.~\ref{alg:nccp-pgs}) & N-CCP & none & no guarantee; at best linear & robust; slow & Jacobi or coloring variants & yes & feasible forces & no \\
Newton, natural map (Alg.~\ref{alg:nccp-newton}) & N-CCP & $3N \times 3N$ & local, superlinear & may be singular; robust & batched solves & yes & may leave $\mathcal{K}$ & no \\
Newton, SOCFB (Alg.~\ref{alg:nse-newton}) & Root-finding & $3N \times 3N$ & local & may be singular; robust & batched solves & yes & may leave $\mathcal{K}$ & yes, for $\epsilon > 0$ \\
Staggered box QPs (Alg.~\ref{alg:cqp-staggered}) & C-QP & inner solver & no guarantee & inner solver & inner solver & yes & feasible forces & no \\
Proximal blueprint (Alg.~\ref{alg:proximal-point-correction}) & C-QCQP & inner solver & no guarantee; fixed points are exact & robust; inner solver & inner solver & yes & inner solver & no \\
\midrule
Proximal minimization (Alg.~\ref{alg:proximal-point}) & any & inner solver & global for convex problems & robust; inner solver & inner solver & yes & inner solver & no \\
Pivoting (App.~\ref{sec:lcp}) & LCP & rank-one update per pivot & finite & direct method & sequential & limited & not applicable & no \\
\bottomrule
\end{tabular}
\caption{\captitle{Properties of the solution algorithms.} The rows group the algorithms of \cref{sec:solution} into those for the convex relaxations (top), those for the exact transcriptions (middle), and the remaining ones (bottom): proximal minimization, which applies to any transcription, and pivoting for the LCP. The columns give the size of the linear system solved per iteration, for $N$ contacts and $n_v$ degrees of freedom; the convergence guarantee and rate; the behavior when $G$ is singular and when the mass ratios are large; the suitability for batched execution on GPUs; whether the algorithm can be warm-started; what it returns when stopped early; and whether its solutions depend smoothly on the problem data (\cref{sec:cmp_solvers}). ``Inner solver'' means that the property is inherited from the algorithm used for the inner problems. ``Feasible forces'' means that every iterate lies in the friction cone, so the algorithm returns admissible forces when it is stopped early, whereas the iterates of the Newton methods can leave it before convergence.}
\label{tab:cmp_algorithms}
\end{table*}

\begin{table*}[t]
\centering\small\setlength{\tabcolsep}{4pt}
\begin{tabular}{P{2.6cm} P{5.6cm} P{4.6cm} P{2.5cm}}
\toprule
Engine & Transcription & Solver & Hardware \\
\midrule
MuJoCo~\cite{todorov2014convex, mujoco_docs}; MJX, MuJoCo Warp~\cite{mjx, mujoco_warp} & Regularized QCQP \eqref{eq:socqp} (dual), NLS \eqref{eq:nls} (primal). & Newton (default) or CG on the primal; PGS on the dual. & CPU (MuJoCo); GPU (MJX, MuJoCo Warp) \\
Drake, SAP~\cite{castro_sim, drake} & NLS \eqref{eq:nls}. & Newton with exact line search. & CPU \\
Bullet~\cite{coumans2015bullet} & C-QP \eqref{eq:cqp}, LCP \eqref{eq:lcp}. & PGS, Pivoting. & CPU \\
ODE~\cite{smith2006ode}; DART~\cite{lee2018dart} & Boxed LCP~\cite{andrews2022contact}. & PGS, Pivoting. & CPU \\
PhysX, Isaac~\cite{makoviychuk2021isaac} & LCP \eqref{eq:lcp}. & PGS, temporal Gauss--Seidel (TGS). & CPU, GPU \\
Chrono~\cite{tasora2016chrono} & L-CCP/QCQP \eqref{eq:lccp}. & PGS, PGD , ADMM. & CPU \\
Siconos~\cite{acary2007siconos, acary2018solving} & N-CCP \eqref{eq:nccp}, root-finding \eqref{eq:root_finding}, C-QCQP \eqref{eq:c_qcqp}, C-QP \eqref{eq:cqp}. & Nonsmooth Gauss--Seidel; Alart--Curnier and Fischer--Burmeister Newton; ADMM. & CPU \\
RaiSim~\cite{hwangbo2018per} & N-CCP \eqref{eq:nccp}. & Contact-by-contact exact solves inside a Gauss--Seidel sweep. & CPU \\
Dojo~\cite{howell2022dojo} & N-CCP \eqref{eq:nccp}. & Primal-dual interior point. & CPU \\
Simple~\cite{carpentier2024compliant, lelidec2023contacts} & C-QCQP \eqref{eq:c_qcqp}. & Proximal ADMM with $\Gamma$ updates, \cref{sec:asocqp_sol}. & CPU \\
Kamino~\cite{tsounis2026kamino} & C-QCQP \eqref{eq:c_qcqp}. & Proximal ADMM with $\Gamma$ updates. & GPU (batched) \\
\bottomrule
\end{tabular}
\caption{\captitle{Contact dynamics routines in current simulators.} The transcription and solution algorithm employed by each simulator, as well as its target hardware. Note that in practice, each simulator ends up implementing its own highly customized contact dynamics routine. Consequently, the assignments inevitably have to correspond to the nearest-neighbors presented in this tutorial, rather than a perfectly accurate description of the exact implementation.}
\label{tab:engines}
\end{table*}

\subsection{Trade-offs Between the Solution Algorithms}\label{sec:cmp_solvers}
\Cref{tab:cmp_algorithms} summarizes the solution algorithms of \cref{sec:solution} along the criteria that matter most in practice. Several patterns stand out.

\smallbreak
\textbf{Cost per iteration versus number of iterations.} First-order methods, such as PGS, PGD, and the Gauss--Seidel schemes for the exact transcriptions, need no linear solve with $G$, so their iterations are cheap, but they converge at best linearly, and slowly when $G$ is ill-conditioned, for example in the presence of large mass ratios. Newton and interior-point methods solve a linear system in every iteration, but they account for the curvature of the problem and need far fewer iterations. ADMM sits in between: it factorizes once and then takes cheap iterations.

\smallbreak
\textbf{Force space versus acceleration space.} The linear systems of the force-space Newton and interior-point methods have size $3N \times 3N$, whereas Newton's method on the NLS transcription solves systems of size $n_v \times n_v$ that inherit the sparsity of $M$. The NLS transcription is therefore the cheaper choice when the contacts outnumber the degrees of freedom ($3N > n_v$), as is common for robots in contact-rich tasks, and the force-space methods are cheaper otherwise.

\smallbreak
\textbf{Rank deficiency.} PGS and PGD never solve a linear system with $G$, and ADMM, interior-point methods, regularization, and proximal minimization all involve adding a positive-definite term to the Hessian as part of the numerical routine. The Newton methods for the exact transcriptions have no such safeguard, so they commonly fall back to PGS sweeps when their Newton matrix becomes singular.

\smallbreak
\textbf{Convergence guarantees.} The algorithms for the convex relaxations inherit the convergence guarantees of convex optimization. The algorithms for the exact transcriptions do not, since the exact problem is nonconvex: the Newton methods converge only locally, and fixed-point schemes such as the staggered box QPs have no general convergence guarantee, although their fixed points are exact solutions. The proximal blueprint of \cref{alg:proximal-point-correction} narrows the gap between the two groups, since it solves the exact transcription with any solver for the regularized convex relaxations as its inner loop, at the cost of nested loops.

\smallbreak
\textbf{Warm starts and early termination.} In real-time applications, solvers are typically warm-started from the previous step and stopped after a fixed number of iterations. First-order methods warm-start well, and their iterates remain in the friction cone, so stopping them early returns admissible, although inaccurate, forces. The exact local Gauss--Seidel scheme is particularly graceful, since each contact satisfies its local law with respect to the current forces of its neighbors. Interior-point methods keep their iterates strictly feasible but are hard to warm-start, and the iterates of the Newton methods can leave the friction cone before convergence.

\smallbreak
\textbf{Smooth solutions.} The interior-point method at a fixed barrier weight $\tau > 0$ and the Fischer--Burmeister Newton method at a fixed smoothing factor $\epsilon > 0$ return solutions that depend smoothly on the problem data, at the cost of relaxing the complementarity conditions, which makes them natural choices when the contact dynamics need to be differentiated (\cref{sec:future}). The other algorithms return solutions of the original nonsmooth problems, whose derivatives jump whenever a contact changes its mode.

\subsection{Implementations on CPUs and GPUs}\label{sec:cmp_implementations}
\Cref{tab:engines} relates widely used simulators to the transcriptions and solution algorithms of this tutorial, and to the hardware they run on. Simulators designed to run on the CPU traditionally rely either on Gauss--Seidel sweeps over the contacts, as in Bullet, ODE, RaiSim, and Siconos, or on Newton methods with sparse factorizations, as in MuJoCo and Drake. GPU-based simulators are designed for throughput, often over several thousand simulations run in parallel, and favor algorithms whose work is independent across contacts, regular across different simulation instances in a batch, and generally involve minimal branching control follow. PhysX~\cite{nvidia2024physx, makoviychuk2021isaac} runs its PGS and TGS~\cite{macklin2019small} solvers on the GPU, MJX~\cite{mjx} and MuJoCo Warp~\cite{mujoco_warp} batch the Newton and conjugate-gradient solvers of MuJoCo over many simulations, and Kamino~\cite{tsounis2026kamino} batches the proximal ADMM version of the blueprint \cref{alg:proximal-point-correction}. Gauss--Seidel sweeps are sequential, so GPU implementations either replace them with Jacobi-type sweeps~\cite{macklin2014unified} or update contacts that do not share a body in parallel~\cite{harada2011parallel, fratarcangeli2015scalable}.

Macklin et al.~\cite{macklin2019non} demonstrate a GPU implementation of nonsmooth Newton methods for coupled rigid and deformable dynamics. Their method combines complementarity preconditioning with an iterative conjugate-residual linear solver, illustrating how second-order methods can exploit GPUs; their formulation uses a smooth isotropic friction model.

\subsection{Choosing a Transcription and a Solver}\label{sec:cmp_choice}
No single combination of transcription and algorithm is best for all applications. The choice depends on which physical artifacts an application can tolerate and on its computational budget.
\begin{itemize}[leftmargin=*, label={\color{bulletcolor}\textbf{\textbullet}}]
    \item \textbf{Throughput and robustness:} When many simulations must run in real time or faster, as in reinforcement learning, the regularized relaxations are the default choice~\cite{todorov2012mujoco}. They always have a unique solution, and Newton's method on the NLS transcription solves them in a few iterations. That being said, lift-off artifacts at sliding contacts and creeping artifacts at sticking contact make them less suitable for applications that demand physical fidelity.
    \item \textbf{Simplicity on a fixed budget:} First-order methods, PGS above all, need no linear solves, warm-start well, and return admissible forces whenever they are stopped. They remain the workhorse of game engines~\cite{catto2005iterative, erleben2005physics}, at the cost of slow convergence for large mass ratios and tall stacks.
    \item \textbf{Physical fidelity:} Applications that depend on static friction or on sliding without lift-off, such as grasping and in-hand manipulation, call for the exact transcriptions. The proximal blueprint reaches them with any solver for the regularized relaxations, the staggered box QPs reach them with solvers for box-constrained QPs, and the Newton methods converge fast near a solution.
    \item \textbf{Differentiable simulation:} Interior-point methods and Newton's method on the smoothed SOCFB function return solutions that depend smoothly on the problem data, which makes them natural choices for gradient-based trajectory optimization and system identification (\cref{sec:future}).
\end{itemize}

\section{Beyond Forward Simulation: Ongoing Developments and Future Perspectives}
\label{sec:future}

The previous sections treated contact dynamics as a forward simulation problem: given the current state and control inputs, compute the state at the next time step. In robotics, contact models are also used inside larger optimization problems, for example to plan motions that make and break contact, or to identify physical parameters from contact interactions. This section briefly surveys two such directions, trajectory optimization and differentiable simulation. 

\subsection{Contact Models for Planning and Optimal Control}
The applications of complementarity-based contact models extend beyond forward simulation. In a robotics context, planning and control constitute important applications, where the introduction of a control objective leads to a non-convex or bi-level optimization problem, even when combined with a convex or regularized contact model. Given the breadth of this topic, we only briefly touch upon it in this section, primarily to direct interested readers to existing pedagogical resources~\cite{tedrake2024underactuated, manchester2023optimalcontrol, toussaint2017tutorial, toussaint2020describing, pavone2026aa203, buchli2017optimal, todorov2006optimal}.

Broadly speaking, there are two approaches to incorporating contact models into trajectory optimization: direct methods and bi-level methods. Direct methods transcribe the contact model equations as constraints in the trajectory optimization problem~\cite{mordatch2012discovery, posa2014direct, manchester2019variational, toussaint2020describing}. This means the equations are gradually solved together with the controls over successive trajectory optimization iterations, and are only guaranteed to be satisfied at the end once the entire trajectory optimization routine converges. When complementarity conditions are retained explicitly, the resulting problems belong to the class of mathematical programs with complementarity constraints (MPCCs). Their usual smooth reformulations violate standard constraint qualifications, motivating specialized optimality conditions and solution methods. Nurkanovi\'c, Pozharskiy, and Diehl~\cite{nurkanovic2025solving} review this theory and relaxation-based algorithms, and introduce the NOSBENCH benchmark to assess their performance on problems arising from nonsmooth optimal control.

In practice, this computational bottleneck can be circumvented by introducing additional regularization or approximate solution strategies to obtain rapid solutions at a frequency sufficient for real-time model-predictive control~\cite{bui2026push, li2026impact}. For example, Consensus Complementarity Control ($C^3$)~\cite{aydinoglu2024consensus} uses an ADMM-based consensus formulation to separate trajectory optimization from parallel contact-mode selection subproblems, enabling online control without a prescribed contact schedule.

Direct methods based on contact-implicit optimization can also be used for instantaneous task-space control. M\'enager et al.~\cite{menager2025frictional} formulate frictional inverse dynamics as a quadratic program with complementarity constraints, combining task objectives, actuator limits, and frictional contact laws. Their augmented Lagrangian solver alternates smooth convex optimization with projections onto actuator and contact constraint sets, allowing the control problem to select contact modes for both rigid and soft robots.

Bi-level methods instead transcribe the constraints of the trajectory optimization problem using the forward simulation function~\cite{tassa2012synthesis, howell2022trajectory, pang2023global}. This means the contact model equations are solved to convergence as an inner optimization problem (to evaluate the forward simulation function) at each outer trajectory optimization iteration. 
As an example, the bi-level method of Dietz et al.~\cite{dietz2026robust} illustrates the use of this approach for robust assembly planning: they smooth both collision detection and contact resolution, compute exact first- and second-order derivatives of the smoothed problems, and optimize over multiple scenarios to account for model uncertainty.

Both direct and bi-level approaches require differentiating through the contact model equations, introducing additional challenges beyond those encountered in forward simulation alone~\cite{toussaint2020describing}.

\subsection{Differentiable Simulation}
The challenges involved in computing the derivatives of the forward simulation function can be broadly grouped into four categories: (i) differentiating collision detection, (ii) differentiating contact dynamics, (iii) obtaining nonzero derivatives in the absence of contact between bodies associated with actuated and unactuated degrees of freedom, and (iv) addressing pathologies arising from time discretization. Differentiable simulation is an active area of research, with numerous works proposing different approaches to tackling these challenges~\cite{Coros2021Differentiable}. In this section, we provide a brief, high-level survey of representative works, with the aim of helping interested readers navigate the literature and identify promising directions for further exploration.

\smallbreak
\textbf{Differentiating collision detection.}
The most common approach to collision detection, adopted by many widely used collision engines \cite{libccd, pan2012fcl, coal2024}, is to approximate geometries using collections of convex primitives. Contact information, including the signed distance $d$, witness points $(p_1, p_2)$, and contact normal $\hat{n}$, is then computed for each pair of primitives using the GJK~\cite{gilbert1988fast} and EPA~\cite{van2003collision} algorithms (see \cref{sec:collision}). The signed distance evolves according to the normal component of the relative contact velocity, $\dot{d} = \hat{n}^\trsp J \dot{q} = v_n$,
so its gradient with respect to the system configuration can be obtained directly from the normal row of the contact Jacobian, $\hat{n}^\trsp J$~\cite{schulman2014motion}. However, since both the normal $\hat{n}$ and the Jacobian $J$ depend on the configuration $q$, differentiating the collision detection routine further requires computing the derivatives of the witness points from which these two quantities are constructed.

For strictly convex primitives, such as spheres and super-ellipsoids \cite{beker2026smoothly, beker2026novel}, the witness points are unique, and change continuously with respect to the system configuration $q$, rendering the collision detection routine differentiable. In contrast, for non-strictly convex primitives, such as capsules, convex polytopes, and sphere-swept convex polytopes (i.e., Minkowski sums of a convex core and a sphere) \cite{tracy2023differentiable}, witness points may be non-unique and can exhibit discontinuous jumps as the configuration changes and the closest pair of features switches (e.g. face--face configurations in polytope collisions~\cite{werling2021fast, schulman2014motion}). As a result, the gradients of the contact information are only piecewise continuous, requiring the downstream gradient-based optimization algorithms to be robust to such cases.

One way to avoid such challenges is to resort to an approximation of the underlying signed distance by a smooth, closed-form function with a sufficiently accurate proximity measure~\cite{wang2013harmonic, beker2025, beker2026smoothly, beker2026novel}. The derivatives of these functions can then be computed analytically, without explicitly solving a collision detection problem. Such constructions have, for example, been used to plan reaching and grasping motions without expensive collision detection~\cite{sandilands2013dexterous}.

When the original nonsmooth geometry must be retained, an alternative is to smooth the underlying nonsmooth optimization problem rather than the geometry itself. One approach is randomized smoothing~\cite{montaut2023differentiable}, which involves solving the optimization problem under randomized perturbations and estimating the sensitivity of its solution using a Monte Carlo estimator~\cite{mohamed2020monte}. Another approach is implicit differentiation, in which the optimality conditions are smoothed, for example, through an interior-point method~\cite{tracy2023differentiable, tracy2024differentiability}, and the implicit function theorem is applied to the resulting system of equations~\cite{jaxopt_implicit_diff}. As an example, Dietz et al.~\cite{dietz2025smoothed} construct smooth growth-distance functions between polytopes by perturbing the optimality conditions of a parametric linear program. By evaluating these distances and their derivatives separately, their approach avoids introducing the variables and constraints of the distance computation into the optimal control problem, thereby reducing the size of the resulting optimization problem.

\smallbreak
\textbf{Differentiating contact dynamics.}
As discussed in this tutorial, computing contact dynamics also involves solving an optimization problem. Consequently, randomized smoothing~\cite{suh2022bundled, suh2022differentiable, lelidec2024leveraging} and implicit differentiation~\cite{lelidec2021differentiable, howell2022dojo, pang2023global, lidec2024end} are again commonly employed approaches. Alternatively, contact forces can be formulated as closed-form functions of the system state, allowing analytical computation of derivatives~\cite {geilinger2020add, kurtz2026inverse, beker2025}. A further approach is to unroll the solver iterations and apply automatic differentiation~\cite{paulus2025hard}. In this context, nonsmooth operations, such as minimum functions, can be replaced with smooth surrogates in the backward pass to obtain well-behaved gradients~\cite{paulus2026softjax}.

Access to exact dynamics derivatives does not, however, guarantee reliable gradients for policy optimization. Suh et al.~\cite{suh2022differentiable} show that averaging trajectory derivatives can give biased estimates of the gradient of an expected return when the underlying trajectories or costs depend discontinuously on policy parameters. Even continuous but stiff models can produce misleading finite-sample estimates when rare, large gradients are missed. Their $\alpha$-order estimator combines first-order derivatives with zeroth-order estimates, which use sampled returns without differentiating the dynamics, to balance accuracy and variance.
Scheffler et al.~\cite{scheffler2026unbiased} address this bias in randomized smoothing: averaging classical derivatives omits contributions from jumps across discontinuity surfaces. They derive a corrected formula that includes these contributions explicitly. When the discontinuity structure is unknown, their corrected first-order bundle gradient estimator combines classical gradients with a zeroth-order residual, remaining unbiased for the gradient of the Gaussian-smoothed objective under the stated assumptions. This distinction matters: unbiased estimation of a smoothed objective's gradient does not imply that smoothing preserves the original objective. Together, these works emphasize that the usefulness of simulator derivatives depends on both the objective being differentiated and the bias and variance of its gradient estimator.

\smallbreak
\textbf{Obtaining nonzero gradients in the absence of contact.}
Here, the gradients of interest measure how an object's predicted motion changes with the robot's configuration or controls. For example, if a robot remains separated from an object throughout a simulated trajectory, small changes to its controls leave the object's motion unchanged, so the corresponding derivatives are zero. An objective based only on the object's motion then provides no gradient signal to guide the robot toward making contact. Smoothing or relaxing the contact model can introduce this sensitivity before contact occurs, providing a surrogate gradient for optimization~\cite{pang2023global}.

One common approach is to introduce a log-barrier term into the optimization objective used by the contact dynamics routine~\cite{howell2022dojo, pang2023global, tracy2024differentiability}. Another approach is to modify the functional form of the penetration resolution term, $\gamma$, so that it remains nonzero even when no penetration occurs~\cite{beker2025, paulus2025hard}. Finally, auxiliary decision variables can be introduced into the contact dynamics optimization problem to allow contact forces to act at a distance~\cite{mordatch2012discovery, toussaint2020describing}.

A complementary strategy retains the hard-contact forward simulation and relaxes the contact conditions only when computing derivatives. Kim et al.~\cite{kim2025contactimplicit} use derivatives of relaxed normal complementarity conditions to guide contact creation and release in contact-implicit model predictive control. Their subsequent \emph{DiffContact} framework~\cite{kim2026diffcontact} applies a central-path relaxation to the full de Saxc\'e-corrected cone complementarity conditions, providing sensitivities across separation, sticking, and sliding. These methods supply surrogate derivatives that help optimization explore neighboring contact modes, while the simulated trajectories continue to satisfy the hard-contact model.

\smallbreak
\textbf{Addressing pathologies arising from time discretization.}
Stewart and Anitescu~\cite{stewart2010optimal} show that, for systems with discontinuous dynamics such as Coulomb friction, differentiating a time-discrete simulation can produce sensitivity errors that do not vanish as the timestep decreases. They establish convergence for a smoothed approximation when the timestep is sufficiently small relative to the smoothing parameter, under suitable assumptions. Thus, accurate trajectories alone do not guarantee accurate derivatives of the continuous-time dynamics.

One approach is to use implicit time integration (i.e., the backward Euler method), which involves solving a root-finding problem whose solution can be differentiated using the implicit function theorem~\cite{geilinger2020add}. Alternatively, adaptive time integration can be implemented using an automatic differentiation library, with gradients computed through backpropagation over its iterations~\cite{diffrax, paulus2025hard}. Finally, adaptive time integration itself can be formulated as a convex optimization problem, allowing sensitivities to be computed using any approach for differentiating convex programs~\cite{kurtz2025cenic}.

For higher-order integration, accurately locating contact transitions is essential: applying a Runge--Kutta scheme across unresolved discontinuities can reduce its accuracy to first order and yield incorrect sensitivities. Finite Elements with Switch Detection (FESD)~\cite{nurkanovic2024fesd} addresses this by treating integration step sizes as unknowns and imposing cross-complementarity conditions that restrict mode transitions to step boundaries. Under suitable assumptions, this recovers the underlying Runge--Kutta method's accuracy and yields convergent numerical sensitivities. Its extension to rigid-body impacts and Coulomb friction, FESD-J~\cite{nurkanovic2024fesdj}, additionally imposes impact complementarity conditions at these boundaries to determine normal and tangential impulses and the associated velocity jumps. These formulations incorporate event detection into the discretized equations, making them suitable for direct contact-implicit optimal control without prescribing the contact sequence.

\section{Conclusion}\label{sec:conclusion}
In this tutorial, we have covered the most common approaches to transcribing cone complementarity-based contact models and solving the resulting problems. Starting from the two phenomenological characterizations of contact interactions—preventing interpenetration and dissipating energy—we have built up, step by step, from rigid-body dynamics and convex analysis to the discrete time-stepping formulations as they are employed by modern simulators, keeping the presentation self-contained throughout.
This discussion led to a central trade-off: exact transcriptions reproduce the Signorini--Coulomb law but lead to nonconvex problems, whereas regularized convex relaxations always admit a unique solution and global convergence, at the price of physical artifacts. The proximal blueprint of \cref{alg:proximal-point-correction} offers a way to bridge this gap: its fixed points solve the exact problem, while its inner iterations can be performed using any solver for the regularized relaxations.

\smallskip

Readers who have worked through this tutorial should now be able to consult the documentation or source code of simulators such as MuJoCo, Bullet, Drake, Kamino, or Simple, identify which transcription each employs and how it is solved, and anticipate the resulting artifacts, such as sliding contacts that lift off or grasped objects that slowly slip.
Furthermore, they should be well equipped to work on their own contact solvers, follow the research literature on contact simulation, and take the next steps toward trajectory optimization and differentiable simulation outlined in \cref{sec:future}.

\section*{Acknowledgements}
This work was supported by the ERC - 101045454 REALRL grant and the German Federal Ministry of Education and
Research (BMBF) through the Tübingen AI Center (FKZ:
01IS18039B) and the Robotics Institute Germany. Georg Martius is a member of the Machine
Learning Cluster of Excellence, EXC number 2064/1 – Project
number 390727645. The authors thank the International Max
Planck Research School for Intelligent Systems (IMPRS-IS)
for supporting Onur Beker.
Michael Posa was supported by an NSF CAREER Award under Grant No. FRR-2238480.
Justin Carpentier was supported by “PR[AI]RIE-PSAI” (ANR-23-IACL-0008), the ERC - 101165695 ARTIFACT, and the France 2030 program, with the references Organic Robotics Program (PEPR O2R).

\newpage
\bibliographystyle{unsrtnat_links} %
{\hypersetup{urlcolor=biblinkcolor}\bibliography{main}}

\newpage
\appendix
\crefalias{section}{appendix}\crefalias{subsection}{subappendix}\crefalias{subsubsection}{subsubappendix}
\section{The Linear Complementarity Based Formulation}\label{sec:lcp}
\begin{figure*}
    \centering
    \usetikzlibrary{arrows.meta,calc,decorations.pathreplacing}
\begin{tikzpicture}[
    font=\footnotesize,
    >={Stealth[length=3.5pt]},
    title/.style={font=\small, inner sep=0pt},
    basis/.style={->, line width=0.8pt, draw=forcecolor},
    vel/.style={->, line width=0.8pt, draw=velocitycolor},
    slip/.style={->, line width=0.8pt, draw=correctioncolor},
    axis/.style={line width=0.6pt, draw=geometrycolor},
    helper/.style={line width=0.4pt, draw=annotationcolor, densely dashed},
    frame/.style={line width=0.5pt, draw=edgecolor},
    note/.style={font=\footnotesize\itshape},
    cpoint/.style={circle, draw=geometrycolor, fill=white, line width=0.5pt, inner sep=0pt, minimum size=3pt},
  ]
  \begin{scope}[shift={(2.18,0.81)}, x={(1cm,0cm)}, y={(0cm,0.35cm)}, z={(0cm,1cm)}]
    \def\coneH{1.79}\def\coneR{0.98}  %
    \coordinate (O) at (0,0,0);
    \coordinate (C) at (0,0,\coneH);                       %
    \foreach \k/\a in {1/225, 2/45, 3/315, 4/135} {   %
      \coordinate (T\k) at ({\coneR*cos(\a)},{\coneR*sin(\a)},\coneH);
      \coordinate (b\k) at ({\coneR*cos(\a)},{\coneR*sin(\a)},0);  %
      \coordinate (t\k) at ({0.72*\coneR*cos(\a)},{0.72*\coneR*sin(\a)},0); %
      \coordinate (l\k) at ({1.6*cos(\a)},{1.6*sin(\a)},0);  %
    }

    \shade[bottom color=fixedbodycolor!36, top color=fixedbodycolor!22]
      (-2.03,0.11,0) -- (-0.05,2.2,0) -- (1.99,-0.06,0) -- (-0.15,-2.06,0) -- cycle;
    \draw[helper] (O) ellipse [x radius=0.98cm, y radius=0.343cm];
    \draw[helper] (b1) -- (b3) -- (b2) -- (b4) -- cycle;
    \draw[axis] (l1) -- (l2);
    \draw[axis] (l3) -- (l4);

    \draw[line width=0.8pt, forcecolor] (O) -- (C);
    \foreach \k in {1,...,4} \draw[line width=0.4pt, geometrycolor] (O) -- (T\k);
    \shade[left color=primalconecolor!45, right color=primalconecolor!20, fill opacity=0.82]
      (O) -- ($(C)+(191.05:0.98cm and 0.343cm)$)
      arc[start angle=191.05, end angle=348.95, x radius=0.98cm, y radius=0.343cm] -- cycle;
    \shade[left color=primalconecolor!15, right color=primalconecolor!42, fill opacity=0.9]
      (C) ellipse [x radius=0.98cm, y radius=0.343cm];
    \begin{scope}
      \clip (C) ellipse [x radius=0.98cm, y radius=0.343cm];
      \draw[line width=0.6pt, geometrycolor] (T2) -- (O) (T4) -- (O);
    \end{scope}
    \draw[line width=0.7pt, geometrycolor, line join=round] (T1) -- (T3) -- (T2) -- (T4) -- cycle;
    \draw[line width=0.8pt, primalconecolor, line join=round]
      ($(C)+(191.05:0.98cm and 0.343cm)$) -- (O) -- ($(C)+(-11.05:0.98cm and 0.343cm)$);
    \draw[line width=0.8pt, primalconecolor] (C) ellipse [x radius=0.98cm, y radius=0.343cm];
    \draw[axis] (C) -- (0,0,2.42);
    \draw[->, line width=0.8pt, forcecolor] ($(C)-(0,0.98,0)$) -- (C)
      node[right, inner sep=2pt, text=forcecolor] {$\hat{n}$};

    \foreach \k in {1,...,4} \draw[basis] (O) -- (t\k);
    \draw[vel] (O) -- (1.18,-0.27,0) node[right, inner sep=2pt, text=velocitycolor] {$v_t$};
    \node[cpoint] at (O) {};
    \node[text=forcecolor] at (-0.58,1.26,0) {$\hat{t}_4$};
    \node[text=forcecolor] at (0.58,1.26,0)  {$\hat{t}_2$};
    \node[text=forcecolor] at (-0.50,-1.3,0) {$\hat{t}_1$};
    \node[text=forcecolor] at (0.62,-1.3,0)  {$\hat{t}_3$};

    \node[note, align=left, anchor=west, text=mutedtextcolor] at (-2.0,1.4,0) {contact\\[-1pt]plane};
    \node[note, anchor=south] (polylbl) at (-1.28,0,2.28) {Polyhedral cone};
    \draw[line width=0.4pt, geometrycolor] (polylbl.south east) ++(-0.15,0.02) -- ($(T1)!0.45!(T4)$);
  \end{scope}

  \node[title, anchor=west] at (3.38,3.30) {Basis vectors};
  \node[anchor=east] at (5.08,2.62) {$B_K =$};
  \draw[line width=0.5pt, line join=round] (5.13,3.07) -- (5.06,3.07) -- (5.06,2.17) -- (5.13,2.17);
  \draw[line width=0.5pt, line join=round] (6.80,3.07) -- (6.87,3.07) -- (6.87,2.17) -- (6.80,2.17);
  \foreach \k/\x in {1/5.30, 2/5.75, 3/6.20, 4/6.65} {
    \draw[line width=0.6pt, forcecolor] (\x,2.24) -- (\x,3.00);
    \node[fill=white, inner sep=1pt, text=forcecolor] at (\x+0.035,2.62) {$\hat{t}_{\k}$};
  }

  \node[title, anchor=west, align=left] at (4.40,1.74) {Projection onto $B_K$\\via dot product};
  \fill[fixedbodycolor!30] (4.40,0.01) rectangle (6.92,1.45);
  \coordinate (P) at (6.24,0.57);
  \coordinate (V) at (4.63,1.36);
  \draw[helper] (V) -- (V |- P) -- (P);
  \draw[vel] (P) -- (V) node[pos=0.45, above right, inner sep=1pt, text=velocitycolor] {$v_t$};
  \draw[basis] (P) -- ++(0.55,0) node[above, inner sep=2pt, pos=0.75, text=forcecolor] {$\hat{t}_k$};
  \node[cpoint] at (P) {};
  \node[above, inner sep=3pt] at (P) {$p$};
  \draw[decorate, decoration={brace, mirror, amplitude=3pt}, line width=0.4pt]
    ($(V |- P)+(0,-0.08)$) -- ($(P)+(0,-0.08)$) node[midway, below=3pt] {$\hat{t}_k^\trsp v_t$};

  \node[title] at (9.48,3.27) {Maximum dissipativity constraint};
  \draw[frame] (7.21,0.01) rectangle (11.75,3.06);
  \node[font=\small] at (9.48,2.70) {$0 \leq {\color{correctioncolor}\beta \mathbf{1}_K} + B_K^\trsp v_t \perp \lambda_t \geq 0$};
  \def\zy{1.16}\def\slipb{0.44}
  \draw[->, axis] (7.46,0.53) -- (7.46,1.93);
  \draw[line width=0.5pt, geometrycolor] (7.46,\zy) -- (10.0,\zy);
  \node[left, inner sep=2pt] at (7.46,\zy) {$0$};
  \foreach \xv/\xb/\tv in {7.71/7.87/-0.275, 8.38/8.56/0.275, 9.07/8.96/0.44, 9.73/9.89/-0.44} {
    \draw[helper] (\xv,\zy+\tv) -- (\xb,\zy+\tv);
    \draw[vel] (\xv,\zy) -- (\xv,\zy+\tv);
    \draw[slip] (\xb,\zy+\tv) -- (\xb,\zy+\tv+\slipb);
  }
  \fill[forcecolor] (9.89,\zy) circle (0.03);
  \node at (7.88,0.66) {$\hat{t}_1^\trsp v_t$};
  \node at (8.50,0.98) {$\hat{t}_2^\trsp v_t$};
  \node[anchor=west] at (9.10,1.34) {$\hat{t}_3^\trsp v_t$};
  \node[anchor=west, text=forcecolor] at (9.93,1.34) {${\color{correctioncolor}\beta} + \hat{t}_4^\trsp v_t = 0$};
  \node[note] (msv) at (8.40,2.26) {Max slip velocity};
  \draw[->, line width=0.4pt] (9.17,2.12) to[bend left=25] (9.03,1.86);
  \draw[->, line width=0.4pt] (9.32,2.14) to[bend left=30] (9.20,1.66);
  \draw[-{Triangle[length=4pt, width=5pt]}, line width=1.2pt] (10.82,1.12) -- (10.82,0.86);
  \node[draw=edgecolor, line width=0.5pt, align=center, inner sep=2pt] at (10.83,0.43)
    {{\color{forcecolor}$\lambda_t \geq 0$}\\{\itshape along} $\hat{t}_4$};
  \node[draw=edgecolor, line width=0.5pt, inner sep=2.5pt, fill=white] (eqb) at (8.95,0.22)
    {$\hat{t}_4^\trsp v_t = -\hat{t}_3^\trsp v_t$};
  \draw[line width=0.4pt, geometrycolor] (9.62,0.40) -- (9.70,0.70);

  \foreach \K/\cx/\a in {4/12.95/0, 6/14.18/30, 8/15.44/0} {
    \begin{scope}[shift={(\cx,1.64)}]
      \pgfmathsetmacro{\step}{360/\K}
      \fill[primalconecolor!25] (\a:0.546)
        \foreach \i in {1,...,\K} { -- ({\a+\i*\step}:0.546) } -- cycle;
      \draw[line width=0.8pt, primalconecolor] (0,0) circle (0.546);
      \foreach \i in {1,...,\K} \draw[basis] (0,0) -- ({\a+\i*\step}:0.546);
      \node[cpoint] at (0,0) {};
      \node at (0,0.75) {$K = \K$};
    \end{scope}
  }
  \node[align=left, anchor=west] at (12.32,0.62) {Increasing $K$\\improves approximation};
\end{tikzpicture}
    \caption{\captitle{LCP approximation of the friction cone.} The N-CCP conditions can be approximated with an LCP by representing the friction force $f_t$ as a positive linear combination of the basis vectors $\hat{t}_k$ in the columns of $B_K$. The maximum dissipativity constraint of the LCP formulation introduces the maximum slip velocity $\beta$ as a decision variable, and dictates that the friction force $f_t$ should only point along a basis vector $\hat{t}_k$ for which $\hat{t}_k^\trsp v_t + \beta = 0$ holds. This in turn implies that the friction force can only point along a single basis vector at a time or lie exactly in-between two basis vectors.}
    \label{fig:lcp}
    \Description{Left: a polyhedral cone inscribed in the circular friction cone, with four tangential basis vectors and the tangential velocity in the contact plane. Middle: the basis matrix with the basis vectors as columns, the projection of the tangential velocity onto a basis vector by a dot product, and a bar chart of the maximum dissipativity constraint in which the friction force is nonzero only along the basis vector that attains the maximum slip velocity. Right: polygonal approximations of the circular cone section with 4, 6 and 8 basis vectors, which get closer to the circle as the number grows.}
\end{figure*}
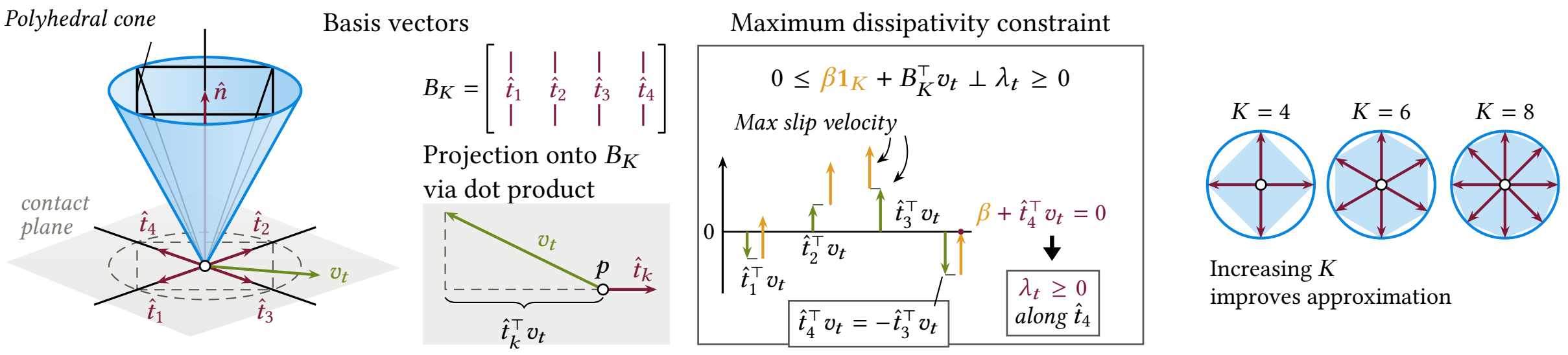

As further traced in \cref{sec:history}, both the historical development and the mathematical structure of the linear-complementarity based formulation of contact dynamics is highly related to the elliptic cone complementarity based formulation discussed in the main body of this tutorial. It is also used by many well-adopted simulators. Therefore, this section presents a compressed summary for completeness, as a more elaborate treatment (which this section heavily borrows material from) can already be found in~\cite{andrews2022contact}, which we refer to interested reader to.
\subsection{Linear Complementarity Problems}
A linear complementarity problem (LCP) is a special case of the linear cone complementarity relation where the cone $\mathcal{K}$ involved is the set of positive real numbers $\mathbb{R}^+$ (which is self-dual):
\begin{align}
    \text{Find} \:\: & x: \nonumber \\ 
    &x \geq 0, \nonumber \\
    &A x + b \geq 0, \nonumber \\
    &\ x^\trsp (A x + b) = 0, \nonumber 
\end{align}
often notated as $0 \leq A x + b \perp x \geq 0$ similar to the general cone complementarity setting. 
A special property of LCP's is that they can be transcribed as an equivalent quadratic program (QP):
\begin{align}
    \ \min_{x} \: &x^\trsp (A x + b), \nonumber \\
    \ \text{s.t.} \: & A x + b \geq 0, \nonumber \\
    & x \geq 0, \nonumber
\end{align}
since positivity constraints on $A x + b$ and $x$ ensure that \mbox{$\min_{x} \ x^\trsp (A x + b)$} is lower-bounded by $0$.\footnote{Note that the QP is nonconvex unless $A \succeq 0$.} That being said, LCPs are solved in practice using specialized algorithms such as pivoting or splitting based methods~\cite{andrews2022contact}, rather than generic QP solvers.

\subsection{Converting N-CCP Conditions to LCP Conditions}
As shown in \cref{fig:lcp}, let $\left\{\hat{t}_k \in \mathbb{R}^3\right\}_{k=1}^K$ be $K$ unit vectors that lie on the tangent plane of contact, shooting outwards from the contact point with equal angles $\frac{2\pi}{K}$ between them.\footnote{The first visualization in \cref{fig:lcp} is adapted from~\cite{andrews2022contact}. We could not have created a better, more informative visualization.} Further let $B_K \in \mathbb{R}^{3 \times K}$ be the matrix that contains $\left\{\hat{t}_k \in \mathbb{R}^3\right\}_{k=1}^K$ in its columns. The main idea behind the LCP conditions is to replace all dependencies on vector norms $\| x\|$ in the N-CCP conditions with linear expressions that depend on the magnitudes of projections on $\left\{\hat{t}_k \in \mathbb{R}^3\right\}_{k=1}$, which can be obtained from a linear matrix product $B_K^\trsp \ x$.

Let us express a contact force not with three decision variables $\left\{f_x, f_y, f_z\right\}$ as $f = [f_x \ f_y \ f_z]^\trsp \in \mathbb{R}^3$, but with $K+1$ decision variables $\left\{\lambda_n, \lambda_1, \cdots,\lambda_K \right\}$ as $f = \lambda_n \hat{n} + B_K \ \lambda_t$, with $\lambda_t = [\lambda_1 \ \cdots \ \lambda_K]^\trsp$ (i.e., the tangential component $f_t$ is expressed via a redundant parameterization $B_K \ \lambda$). Let us also define the notation $\mathbf{1}_K \in \mathbb{R}^K$ for a vector of all ones. One can then linearly approximate the Coulomb law $\|f_{t} \| \leq \mu f_{n}$ in \eqref{eq:nccp2} as $\mathbf{1}_K^\trsp \lambda_t \leq \mu\lambda_n$. Similarly, one can express \cref{eq:mcp_4} that dictates the maximal dissipation direction as a linear complementarity condition of the following form:
\begin{align}
    \text{Find} \:\: & \beta \in \mathbb{R}, \lambda_t \in \mathbb{R}^K: \nonumber \\ 
    & 0 \leq \beta \mathbf{1}_K + B_K^\trsp v_t \perp \lambda_t \geq 0.  
\end{align}
To make sense of this equation, let us first assume $v_t \neq 0$ and identify the tangential component $k^* = \arg\min_k \ \hat{t}_k^\trsp v_t$ that corresponds to the most negative (i.e., smallest) entry $\beta_{min} = \hat{t}_{k^*}^\trsp v$ in $B_K^\trsp v$.
The condition $0 \leq \beta \mathbf{1}_K + B_K^\trsp v$ then implies that $\beta + \beta_{min} \geq 0$ holds. The conditions $(\beta \mathbf{1}_K + B_K^\trsp v) \perp \lambda_t$ and $\lambda_t \geq 0$ together then imply that only the component $\lambda_{t,k^*}$ of $\lambda_t$ can be non-zero (since the entries in $\beta \mathbf{1}_K + B_K^\trsp v$ for $k \neq k^*$ are strictly positive), thus constraining $\lambda_t$ to point exactly towards the maximal dissipation direction $\hat{t}_{k^*}$.

All of the aforementioned considerations can be jointly stated to give the complete list of LCP conditions (for a single contact):
\begin{mysubequations} \label{eq:lcp}
\begin{align}
    \text{Find} \:\: & v^+ \in \mathbb{R}^3, \lambda_{n} \in \mathbb{R}, \lambda_{t} \in \mathbb{R}^K, \beta_i \in \mathbb{R}: \nonumber \\ 
    & 0 \leq (\gamma_n + v_{n}^+) \perp \lambda_{n} \geq 0,  \label{eq:lcp1} \\
    & 0 \leq \beta \mathbf{1}_K + B_K^\trsp v_{t}^+ \perp \lambda_{t} \geq 0, \label{eq:lcp2} \\
    & 0 \leq \left( \mu \lambda_{n} - \mathbf{1}_K^\trsp \lambda_{t}\right) \perp \beta \geq 0. \label{eq:lcp3}
\end{align}
\end{mysubequations}

Looking at equations \eqref{eq:lcp2} and \eqref{eq:lcp3} together, we see that the scalar $\beta$ implements a switching mechanism: if $v_{t}^+ = 0$ and $\lambda_n > 0$ (i.e., sticking contact), then $\beta = 0$ has to hold, since $\beta \neq 0$ in \eqref{eq:lcp3} implies at least one non-zero entry in $\lambda_{t}$, which contradicts the orthogonality condition \eqref{eq:lcp2}. Therefore, when  $v_{t}^+ = 0$, \eqref{eq:lcp2} and \eqref{eq:lcp3} reduce to $\lambda_{t} \geq 0$ and $0 \leq \left( \mu\lambda_{n} - \mathbf{1}_K^\trsp \lambda_{t}\right)$, meaning $\lambda_{t}$ is free to point at whichever direction makes the tangential velocity $v_{t}^+ = 0$ and does not have to saturate the Coulomb law upper limit $\mu\lambda_{n}$. If $v_{t}^+ \neq 0$ and $\lambda_n > 0$ (i.e., sliding contact), then $\beta \neq 0$ since there exists at least one negative entry in the vector $B_K^\trsp v_{t}^+$ and $\beta \mathbf{1}_K + B_K^\trsp v_{t}^+ \geq 0$ has to hold. Therefore, when $v_{t}^+ \neq 0$, $\lambda_{t}$ is constrained by \eqref{eq:lcp2} to point at the maximal dissipation direction, and by \eqref{eq:lcp3} to saturate the Coulomb upper limit $\mu\lambda_{n} = \mathbf{1}_K^\trsp \lambda_{t}$.

To show how the conditions in \crefrange{eq:lcp1}{eq:lcp3} constitute an LCP in the standard form $0 \leq A x + b \perp x \geq 0$, let us carry out the algebra for a system with a single contact point (since extending the algebra to multiple contact points is straight-forward). The complete root finding problem (for a single contact point) is:
\begin{equationbox}{LCP Conditions (Single Contact Point)}
\begin{align}
    \text{Find} \:\: & v^+ \in \mathbb{R}^3, \lambda_{n} \in \mathbb{R}, \lambda_{t} \in \mathbb{R}^K, \beta \in \mathbb{R}: \nonumber \\ 
    &v^+ = v_f + G \left(\hat{n} \lambda_n + B_K \ \lambda_t\right) \Delta t, \nonumber \\
    & 0 \leq (\gamma_n + v_{n}^+) \perp \lambda_{n} \geq 0,  \nonumber \\
    & 0 \leq \beta \mathbf{1}_K + B_K^\trsp v_{t}^+ \perp \lambda_{t} \geq 0, \nonumber \\
    & 0 \leq \left( \mu\lambda_{n} - \mathbf{1}_K^\trsp \lambda_{t}\right) \perp \beta \geq 0. \nonumber
\end{align}
\end{equationbox}
Let us make the following definitions:
\begin{align}
    &s_n = (\gamma_n + v_n^+), \nonumber \\
    &s_\beta = \beta \mathbf{1}_K + B_K^\trsp v_t^+, \nonumber \\
    &s_t = \left( \mu\lambda_n - 1_K^\trsp \lambda_t\right). \nonumber 
\end{align}
Substituting $v^+ = v_f + G \left(\hat{n} \lambda_n + B_K \ \lambda_t\right) \Delta t$ into $s_n$ and $s_\beta$, and extracting the normal component $v_n = \hat{n}^\trsp v^+$ and tangential component $v_t = \left(\Id - \hat{n} \hat{n}^\trsp\right) v^+$, the entire root finding problem can be compactly stated as:
\begin{align}
&\begin{bmatrix}
    \hat{n}^\trsp  G \hat{n} \Delta t & \hat{n}^\trsp G B_K  \Delta t & 0 \\
    B_K^\trsp [\Id - \hat{n}\hat{n}^\trsp]G \hat{n} \Delta t & B_K^\trsp [\Id - \hat{n}\hat{n}^\trsp] G B_K \Delta t & \mathbf{1}_K \\
    \mu & - \mathbf{1}_K^\trsp & 0
\end{bmatrix} 
\begin{bmatrix}
    \lambda_n \\
    \lambda_t \\
    \beta
\end{bmatrix}  \nonumber \\ + 
& \begin{bmatrix}
    \hat{n}^\trsp  v_f + \gamma_n \\
    B_K^\trsp [\Id - \hat{n}\hat{n}^\trsp] v_f \\
    0
\end{bmatrix} = 
\begin{bmatrix}
    s_n \\
    s_\beta \\
    s_t
\end{bmatrix}, \nonumber  \\
\nonumber \\
& 0 \leq s_n \perp \lambda_n \geq 0, \quad 0 \leq s_\beta \perp \lambda_t \geq 0\nonumber, \quad 0 \leq s_t \perp \beta \geq 0. \nonumber 
\end{align}
Writing the first matrix equation as \mbox{$A [\lambda_n \ \lambda_t \ \beta]^\trsp + b = [s_n \ s_\beta \ s_t]^\trsp \nonumber$}, one can express the LCP conditions in standard form:
\begin{equation}
0 \leq \left( A
\begin{bmatrix}
    \lambda_n \\
    \lambda_t \\
    \beta
\end{bmatrix} + b \right) \perp 
\begin{bmatrix}
    \lambda_n \\
    \lambda_t \\
    \beta
\end{bmatrix} \geq 0. \nonumber  
\end{equation}
Extending the derivation to $N$ contacts amounts to duplicating the computations above $N$ times in block matrix form.

\section{A (Not so) Brief History of Cone Complementarity Based Contact Models}\label{sec:history}
The main tutorial presented the transcriptions and solution methods out of chronological order (\cref{sec:outline}). This appendix instead traces their historical emergence for readers who wish to place the material in context or consult the original sources.

\subsection{Early Beginnings and the Continuous-Time Formulation}

\smallbreak
\textbf{The empirical friction law.} The ingredients of complementarity-based contact models date back more than three centuries. Amontons (1699)~\cite{Amontons1699, Hutchings2021} and Coulomb (1781, published 1785)~\cite{Coulomb1785} established the empirical law of dry friction, which in its modern idealized form states that the tangential contact force directly opposes the slipping velocity, and that its magnitude is bounded by a coefficient times the normal force.

\smallbreak
\textbf{Unilateral constraints and multipliers.} Fourier (1798)~\cite{fourier1798statique, eom_virtual_displacements} extended the principle of virtual work to systems in static equilibrium under one-sided constraints, such as a body resting against an obstacle it is not attached to, which is one of the first appearances of what are now called unilateral (i.e., one-sided, inequality) constraints. Ostrogradsky (1838)~\cite{ostrogradsky1838momens, eom_virtual_displacements} then turned Fourier's equilibrium conditions (i.e., tests that must hold for every admissible virtual displacement) into equations that can be solved for the constraint forces, with one multiplier per constraint, which must be nonnegative when the constraint is unilateral. In the same memoir, he extended this to motion, where the multiplier of a unilateral constraint becomes zero as soon as the system leaves the constraint, one of the first appearances of concepts such as active and inactive constraints and complementarity conditions.

\smallbreak
\textbf{Non-uniqueness and non-existence.} Jellett (1872)~\cite{jellett1872treatise} examined the single-contact problem of a rigid body under Coulomb friction by working through its individual contact modes (i.e., contact or separation, and sticking or sliding), and reported configurations in which the solution is not unique. Painlev\'e (1895)~\cite{Painleve1895} further reported configurations where a solution can also fail to exist.

\smallbreak
\textbf{Acceleration-level complementarity.} Mayer (1899)~\cite{mayer1899aufstellung} then derived the acceleration-level complementarity conditions for frictionless unilateral constraints from Gauss's principle of least constraint.
Therefore, by the 1900s, all ingredients of the continuous-time formulation of the complementarity conditions for frictional contact at the acceleration level, for a single contact point, were known.

\subsection{Branching Due to the Pathologies of the Continuous-Time Formulation}
The non-uniqueness and non-existence results of Jellett and Painlev\'e for the continuous-time complementarity-based formulation of frictional contact can be considered a branching point in the tradition. It is instructive for the purposes of this tutorial to study the subsequent developments under three groups:
\begin{itemize}[leftmargin=*, label={\color{bulletcolor}\textbf{\textbullet}}]
    \item Further developments of the frictionless theory
    \item Event-driven simulation
    \item Emergence of the impulse-based theory
\end{itemize}

\subsubsection{Further developments of the frictionless theory} The pathologies that lead to the non-uniqueness and non-existence results only emerge in the presence of friction; therefore, they do not hinder further developments of the frictionless theory. Delassus (1917)~\cite{Delassus1917} addressed the question of which of several simultaneous unilateral constraints remain active, via a systematic treatment of all possible combinations of contact modes, and showed that exactly one combination is always consistent; the quadratic form he used for this, which couples the constraints through the inverse inertia, survives as the matrix that now bears his name.

\smallbreak
\textbf{Elastic contact and variational inequalities.} While Delassus worked with rigid bodies,
Signorini (1933, 1959)~\cite{Signorini1959} instead worked with elastic bodies and proposed treating the contact region as an unknown subset of the boundary, so that there is no finite list of modes to enumerate. Signorini characterized the solution as the minimizer of the potential energy over the admissible displacements, and Fichera (1963)~\cite{Fichera1963} proved existence and uniqueness of this minimizer in a suitable class of displacements. The first-order condition for this minimization is a variational inequality, and Fichera's solution is generally regarded as the beginning of the theory of variational inequalities. This shift from case-by-case enumeration to a single minimization where the solution carries its own active set,
can be seen as the precursor of the way modern simulators address the problem via optimization algorithms rather than combinatorial search over possible constraint activations.

\smallbreak
\textbf{Primal and dual optimization formulations.} Moreau (1966)~\cite{Moreau1966}, starting from the problem of Delassus, showed with the tools of convex analysis that the accelerations solve a convex quadratic program, namely Gauss's principle restricted to the admissible accelerations as in Mayer and Zermelo (1899)~\cite{mayer1899aufstellung, zermelo1899bewegung}, whose optimality conditions are the complementarity conditions, and that the contact forces solve the dual quadratic program. This is the first emergence of what is now called the primal and dual forms of the complementarity-based contact model. These developments mean that by the 1960s the frictionless contact problem had a rigorous optimization-based formulation, a perspective most of the well-adopted simulation frameworks of today also subscribe to.

\subsubsection{Event-driven simulation} The non-uniqueness and non-existence results of Jellett and Painlev\'e set off a decade of debate in mechanics that peaked between 1905 and 1910~\cite{painleve1905lois, lecornu1905loi, klein1909painleve}. The eventual consensus was that Coulomb's law was inconsistent with the perfectly rigid body idealization in continuous time, and two routes for repair were proposed: i)~admit compliance, or ii)~admit discontinuities in the velocity even in the absence of collisions~\cite{brogliato1996nonsmooth, genot1999new}. The first route leads to penalty-based contact models and soft-body simulation, outside the family of complementarity-based contact models covered in this tutorial.

\smallbreak
\textbf{Tracking contact events.} The event-driven simulation paradigm takes the latter route. It allows velocity discontinuities at isolated, explicitly detected instants and keeps the acceleration-level formulation everywhere else by integrating the dynamics between \say{contact events} as a continuous ODE and monitoring for such events along the way. For example, a signed distance between collision geometries crossing zero is a collision, a normal force reaching zero signals that a contact may open, a slip velocity reaching zero that slip may turn into stick, and a friction force reaching the cone boundary that stick may turn into slip.

\smallbreak
\textbf{Resolving mode transitions and impacts.} When an event function triggers, the integrator stops, locates the time instant of the event by root finding, and re-solves the acceleration-level mode selection with the updated active set to decide which transition actually takes place. If the event is a collision or one of Painlev\'e's inconsistent configurations, it applies an \say{impact routine} relating pre- and post-event contact velocities to introduce a discontinuity in the velocity at this time instant. The paradox is thus handled via a \say{detect-and-patch} approach, rather than a unified theory.

\smallbreak
\textbf{Complementarity formulations in mechanics.} The event-driven branch was developed across two communities, mechanics and graphics. In mechanics, building on L\"otstedt (1982)~\cite{lotstedt1982mechanical}, Glocker and Pfeiffer (1993)~\cite{GlockerPfeiffer1993} formulated the mode selection between events (detachment, stick, and slip) for planar friction as an acceleration level complementarity problem, and the impact routine as an impulse-level complementarity problem (1995)~\cite{GlockerPfeiffer1995}. The book of Pfeiffer and Glocker (1996)~\cite{PfeifferGlocker1996} made these the basis of a systematic event-based program, with acceleration-level complementarity problems selecting the contact modes between events and impulse-level ones resolving the impacts.

\smallbreak
\textbf{Computational complexity and graphics solvers.} In computer graphics, Baraff (1991, 1993)~\cite{baraff1991coping, baraff1993issues} asked how expensive the mode selection is and proved that deciding whether a configuration of rigid bodies with sliding Coulomb friction admits bounded contact forces at the acceleration level, which he called the frictional consistency problem, is NP-complete. He then traced the hardness to the traditional principle that impulsive forces should be introduced if and only if bounded forces cannot maintain the constraints (i.e., a detect-and-patch approach), reformulated the problem as computing either bounded forces or impulses at each solve, and, for sliding friction, gave an algorithm built on Lemke's method with polynomial running time. Baraff (1994)~\cite{Baraff1994} then supplied a fast pivoting solver in the style of Dantzig's algorithm, which is guaranteed to terminate with an exact solution for frictionless contact but comes without a termination guarantee with friction.

\subsubsection{Emergence of the impulse-based theory}
Beyond the practical limitations that prevent the event-driven approach from scaling to complex scenes with many active contacts, it also has inherent theoretical limitations. For example, a bouncing ball that comes to rest after infinitely many bounces in finite time cannot be modeled event by event. Or several contacts closing at once may have no canonical event-by-event resolution, since the outcome depends on the order in which the impacts are processed~\cite{brogliato1996nonsmooth, acary2008numerical}.

\smallbreak
\textbf{Unifying forces and impacts.} The impulse-based perspective resolves these problems via the following principle. There is no reason to split the treatment into bounded forces applied over a time interval and unbounded forces (i.e., impacts) applied at a time instant: both are contributions to the total \say{impulse} (i.e., force accumulated across time) over the simulation time step, which determines the end-of-step velocity of the rigid-body system. Therefore, the complementarity relations should be formulated in the language of timestep-aggregate impulses and end-of-step velocities rather than per-time-instance forces and per-time-instance accelerations.

\smallbreak
\textbf{Capturing events within a time step.} Acary and Brogliato~\cite{acary2008numerical} call the resulting time-stepping schemes \say{event capturing}, in contrast to the \say{event tracking} of the previous subsection: accumulations of impacts (e.g., bouncing ball) and simultaneous impacts are both absorbed into the step, and no event time ever needs to be located. The downside is a reduction in accuracy, since contact events are \say{blurred} across the non-infinitesimal simulation stepsize.

\smallbreak
\textbf{Measure differential inclusions.} Moreau's seminal work on the \say{measure differential inclusion} framework (1983, 1985, 1988)~\cite{moreau1983liaisons, moreau1985standard, Moreau1988} supplied the theoretical underpinnings for the impulse-based branch. It provided a single equation of motion that holds during smooth motion and at impacts alike, written in the language of measures that assign to each time interval the total impulse transmitted during it, rather than of functions that assign a force to each instant.

\smallbreak
\textbf{From theory to time-stepping.} Moreau (1988, 1994)~\cite{Moreau1988, moreau1994numerical} and Jean (1999)~\cite{Jean1999} in turn provided an integration scheme to discretize the measure differential inclusion, the so-called method of \say{nonsmooth contact dynamics}, and a numerical solution scheme based on the Gauss--Seidel method. This made the measure differential inclusion framework practically relevant, setting the stage for the discrete time-stepping routines that constitute the focus of this tutorial and the basis of most well-adopted simulation frameworks of today. Dubois, Acary, and Jean~\cite{DuboisJeanAcary2018} provide a historical and methodological account of this transition from Moreau's nonsmooth mechanics to the practical Contact Dynamics method.

\subsection{Emergence of the Modern Discrete Time-Stepping Methods}

\smallbreak
\textbf{Nonsmooth time-stepping.} The nonsmooth contact dynamics research program of Moreau and Jean can be considered the progenitor of the modern discrete time-stepping approach to contact simulation. It used circular friction cones and the exact Coulomb law, solving each step by repeated Gauss--Seidel sweeps over the contacts, in which the local problem of separation, sticking, or sliding at one contact is solved exactly in two dimensions and with a few generalized Newton iterations in three. Alart and Curnier (1991)~\cite{AlartCurnier1991}, working on quasi-static contact between elastic bodies, wrote the same law as the root of a projection residual, which opened it to Newton's method.

\smallbreak
\textbf{Polyhedral approximations.} Stewart and Trinkle (1996)~\cite{stewart1996implicit} and Anitescu and Potra (1997)~\cite{anitescu1997formulating} instead replaced the circular cone by a polyhedral one, so that the whole step becomes a linear complementarity problem that Lemke's pivoting algorithm can solve in finitely many steps. Stewart and Trinkle, who imposed nonpenetration on a linearized gap, could guarantee a solution for a single contact, or for several only under conditions on the contact normals, while Anitescu and Potra, who imposed it on the normal velocity, proved that a solution exists and is found by Lemke's algorithm for any number of contacts. The downside is anisotropic friction and additional unknowns per contact.

\smallbreak
\textbf{Convergence to continuous-time solutions.} Stewart (1998)~\cite{stewart1998convergence} then proved, at least for a single contact, that as the step size shrinks the solutions of such a scheme converge to solutions of the measure differential inclusion of Moreau. This gave the first proof that Painlev\'e's examples admit solutions once velocity is allowed to jump without collision, which the acceleration-level theory could not express.

\smallbreak
\textbf{Cone complementarity and projection methods.} De Saxc\'e and Feng (1991, 1998)~\cite{DeSaxceFeng1991, DeSaxceFeng1998}, working on quasi-static contact between elastic bodies, showed that nonpenetration and Coulomb friction, two conditions on different objects, are together equivalent to a single condition on the friction cone, which can be written as one complementarity condition between the friction cone and its dual. This formulation shifts the contact velocity along the normal direction by the friction coefficient times the slip speed, introducing what is now called the \say{de Saxc\'e correction}. This new transcription is algorithmic: it turns the exact law into the fixed point of a single projection onto the cone, a form amenable to solution using a wide variety of numerical schemes.

Glocker and Studer (2005--2007)~\cite{GlockerStuder2005, StuderGlocker2006, StuderGlocker2007} combined the midpoint time-stepping scheme of Moreau (1988)~\cite{Moreau1988} with Glocker's set-valued laws and replaced the exact constraint-by-constraint solves of the Gauss--Seidel sweep with a projected step, projecting the normal force onto the nonnegative numbers and the friction force onto its disk separately. This yields the JORprox and SORprox iterations, projected Jacobi and Gauss--Seidel methods whose fixed points satisfy the exact law and whose updates require only a closed-form projection per constraint.

\smallbreak
\textbf{Convex relaxations and scalable solvers.} The exact frictional step is a nonconvex problem, with hardness at the acceleration level proved by Baraff and no polynomial-time algorithm known at the velocity level, whereas a convex program can be solved to any prescribed accuracy in polynomial time and, in practice, by the standard methods of convex optimization. Anitescu (2006)~\cite{anitescu2006optimization} therefore relaxed the law to a convex one, which in retrospect can be called \say{dropping the de Saxc\'e correction}; the relaxed scheme first appeared in his work with Hart (2004)~\cite{anitescu2004constraint}, and the 2006 paper proved that it converges as the step size shrinks.
This relaxation, however, comes at the cost of a physical artifact: sliding contacts separate at a rate proportional to the slip speed.

Anitescu and Tasora (2010)~\cite{AnitescuTasora2010} later supplied projected Gauss--Seidel and Jacobi iterations for the circular cone with a convergence proof, Tasora and Anitescu (2011)~\cite{TasoraAnitescu2011} extended them to joints in a matrix-free form, and accelerated projected gradient methods followed (Mazhar et al., 2015)~\cite{mazhar2015nesterov}. Acary, Cadoux, Lemar\'echal, and Malick (2011)~\cite{acary2011formulation} showed that the exact frictional step is a fixed point over convex second-order cone constrained quadratic programs in which the de Saxc\'e correction term is frozen.

\smallbreak
\textbf{Regularization and primal formulations.} Todorov (2011, 2014)~\cite{Todorov2011, todorov2014convex} arrived at a convex program of the same kind, minimizing kinetic energy in contact space subject to friction-cone constraints, with non-penetration imposed as a constraint in 2011 and through a reference velocity in 2014, and smoothed it so that the dynamics can be inverted. The 2011 model does this with an interior-point barrier and a regularization term; the 2014 model keeps only a regularization term scaled by a user parameter, which MuJoCo later reparameterized as impedance, along with constraint stabilization generated by a spring--damper reference.

Todorov's further work on the MuJoCo simulator, documented in its manual rather than in a paper, first wrote this problem in the accelerations with the constraints imposed on auxiliary variables (2016)~\cite{MuJoCoManual2016}, and then eliminated these variables and the associated constraints analytically (MuJoCo 1.50, 2017)~\cite{MuJoCo150Release, MuJoCoReducedPrimal}. The resulting unconstrained nonlinear least squares problem in the accelerations, in which the cone constraints implicitly enter through the cost, is solved by Newton's method.

\smallbreak
\textbf{Recent solver developments.} Hwangbo, Lee, and Hutter (2018)~\cite{hwangbo2018per} kept hard, unregularized contact and solved it contact by contact inside a Gauss--Seidel-type sweep, finding each contact's impulse by a bisection search. Howell et al.\ (2022)~\cite{howell2022dojo} instead proposed solving the nonlinear complementarity problem with the circular cone by a primal--dual interior-point method, and obtained smooth approximate gradients with respect to the state, the control inputs, and the model parameters by implicitly differentiating the solver's relaxed optimality conditions.

Castro, Permenter, and Han (2023)~\cite{castro_sim} arrived at a closely related unconstrained convex problem in the velocities, which they derived as the primal counterpart of earlier impulse-based formulations, took the regularization from physical stiffness and dissipation, and proved that their Newton-type solver converges globally.

Carpentier, Le Lidec, and Montaut (2024)~\cite{carpentier2024compliant} adopted the cascaded variant of the fixed-point structure of Acary et al.\ discussed by Acary, Br\'emond, and Huber (2018)~\cite{acary2018solving}, running ADMM on the cone program with the de Saxc\'e correction term refreshed at every iteration rather than after a converged inner solve; they added a proximal term that keeps the linear solves well posed when the Delassus matrix is singular and vanishes at a fixed point, and made the regularization an optional compliance parameter, so that one solver covers rigid and compliant contact.

\end{document}